\documentclass[]{article}
\usepackage{proceed2e}
\usepackage{url}
\usepackage{amsmath}
\usepackage{amsfonts}
\usepackage{graphicx}
\usepackage{subfigure}

\newtheorem{lemma}{Lemma}

\title{Accurate Estimators for Improving Minwise Hashing and b-Bit Minwise Hashing}

\author{ {\bf Ping Li} \\
Department of Statistical Science \\
Cornell University\\
Ithaca, NY 14853 \\
\textbf{pingli@cornell.edu}
\And
{\bf Christian K\"onig}  \\
Microsoft Research          \\
Microsoft Cooporation\\
Redmond, WA 98052 \\
\textbf{chrisko@microsoft.com}
}

\begin{document}

\maketitle

\begin{abstract}\vspace{-0.2in}
Minwise\footnote{First draft in  March, slightly modified in June, 2011.} hashing is the standard technique in the context of search and databases for efficiently estimating set (e.g., high-dimensional 0/1 vector) similarities. Recently, b-bit minwise hashing was proposed which significantly improves upon the original minwise hashing in practice by storing only the lowest b bits of each hashed value, as opposed to using 64 bits. b-bit hashing is particularly effective in applications which mainly concern sets of high similarities (e.g., the {\em resemblance} $>0.5$). However, there are  other important applications in which not just pairs of high similarities matter. For example, many learning algorithms require all pairwise similarities and it is expected that only a small fraction of the pairs are similar. Furthermore, many applications care more about {\em containment} (e.g., how much one object is contained by another object)  than the {\em resemblance}. In this paper, we show that the estimators for minwise hashing and b-bit minwise hashing used in the current practice can be systematically improved and the improvements are most significant for set pairs of low resemblance and high containment.
\end{abstract}\vspace{-0.2in}

\section{Introduction}\vspace{-0.1in}

Computing the size of set intersections is a fundamental problem in information retrieval, databases, and  machine learning.  For example, binary document vectors represented using $w$-shingles can be viewed either as  vectors of very high dimensionality  or as sets. The seminal work of {\em minwise hashing}~\cite{Proc:Broder,Proc:Broder_WWW97} is a standard tool for efficiently computing  {\em resemblances} (Jaccard similarity) among extremely high-dimensional (e.g., $2^{64}$) binary vectors, which may be documents represented by $w$-shingles ($w$-grams, $w$ contiguous words) with $w=5$ or 7~\cite{Proc:Broder,Proc:Broder_WWW97}. Minwise hashing    has been successfully applied to a very wide range of real-world problems especially in the context of search; a partial list  includes~\cite{Proc:Broder,Proc:Broder_WWW97,Proc:Bendersky_WSDM09,Article:Forman09,Proc:Cherkasova_KDD09,Proc:Pandey_WWW09,Proc:Buehrer_WSDM08,Article:Urvoy08,Proc:Nitin_WSDM08,
Article:Dourisboure09,Proc:Chierichetti_KDD09,Proc:Gollapudi_WWW09,Article:Kalpakis08,Proc:Najork_WSDM09}.

The {\em resemblance}, $R$, is a widely used measure of similarity between two sets. Consider two sets
$S_1, \  S_2 \subseteq \Omega = \{0, 1, 2, ..., D-1\}$, where $D$, the size of the dictionary, is often set to be $D=2^{64}$ in industry practice. Denote $a=|S_1\cap S_2|$.  $R$ is defined as
\begin{align}\notag
R = \frac{|S_1 \cap S_2|}{|S_1 \cup S_2|} = \frac{a}{f_1 + f_2 - a}, \hspace{0.15in}  \  f_1 = |S_1|, \  f_2 = |S_2|.
\end{align}

Minwise hashing applies a random permutation $\pi: \Omega\rightarrow\Omega$ on $S_1$ and $S_2$.  Based on an elementary probability result:
\begin{align}
\mathbf{Pr}\left(\text{min}({\pi}(S_1)) = \text{min}({\pi}(S_2)) \right) = \frac{|S_1
  \cap S_2|}{|S_1 \cup S_2|}=R,
\end{align}
one can store the smallest elements under $\pi$, i.e.,  ${\min}({\pi}(S_1))$ and ${\min}({\pi}(S_2))$, and then repeat the permutation $k$ times to estimate $R$. After $k$ minwise independent permutations, $\pi_1$,
$\pi_2$, ..., $\pi_k$,  one can  estimate $R$ without bias,  as:
\begin{align}\label{eqn_RM}
&\hat{R}_{M} = \frac{1}{k}\sum_{j=1}^{k}1\{{\min}({\pi_j}(S_1)) =
  {\min}({\pi_j}(S_2))\}, \\\label{eqn_Var_M}
&\text{Var}\left(\hat{R}_{M}\right) = \frac{1}{k}R(1-R).
\end{align}

The common practice is to store each hashed value, e.g., ${\min}({\pi}(S_1))$ and ${\min}({\pi}(S_2))$, using 64 bits~\cite{Proc:Fetterly_WWW03}. The storage cost (and consequently the computational cost) will be prohibitive in  large-scale applications~\cite{Proc:Manku_WWW07}.

It is well-understood in practice that one can reliably replace a permutation with a reasonable hashing function; see the original minwise hashing paper~\cite{Proc:Broder} and the followup theoretical work~\cite{Proc:Broder_Stoc98}. In other words, there is no need to store these $k$ permutations. 

In this paper, we first observe the standard practice of minwise hashing, i.e., using (\ref{eqn_RM}), can be substantially improved for important scenarios. In fact, we will show that (\ref{eqn_RM}) is  optimal only when the sets are of the same size, i.e., $f_1=f_2$, which is not too common in practice.

\begin{figure}[h]
\begin{center}
\includegraphics[width = 2.5in]{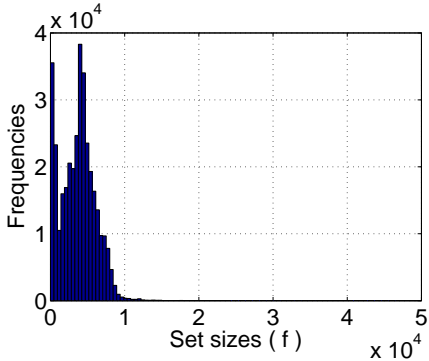}
\end{center}
\vspace{-0.2in}
\caption{Histograms of $f_i$ (number of non-zeros in the $i$-th data vector) in {\em webspam} dataset.  }\label{fig_FSpam}
\end{figure}

Figure~\ref{fig_FSpam} presents an example based on the {\em webspam} dataset (available from the LibSVM site), which contains 350000 documents represented using binary vectors of $D=16$ million dimensions. Compared to the Web scale datasets with billions of documents in $2^{64}$ dimensions, {\em webspam} is relatively small and only uses 3-grams. Nevertheless, this example demonstrates that the set sizes (numbers of non-zeros),  $f_i = |S_i|$, distribute in a wide range. Therefore, when we compare two sets, say $S_1$ and $S_2$, we expect the ratio $f_1/f_2$ will often significantly deviate from 1.

Indeed, we computed the ratios $f_1/f_2$ for all pairs in {\em webspam}. Without loss of generality, we always assume $f_1\geq f_2$. There are altogether 61 billion pairs with the mean $f_1/f_2$ = 5.5 and the standard deviation (std) = 9.5. Thus, we expect that $f_2/f_1 = 0.2\sim0.5$ is common and $f_2/f_1<0.1$ is also fairly frequent.

\vspace{-0.05in}
\subsection{The 3-Cell Multinomial Problem}\vspace{-0.05in}

The standard estimator (\ref{eqn_RM}) is based on a binomial distribution. However, the problem really  follows a 3-cell multinomial distribution. Define $z_1 = {\min}({\pi}(S_1))$ and $z_2 = {\min}({\pi}(S_2))$. The three probabilities are:
\begin{align}\label{eqn_P=}
&P_= = \mathbf{Pr}\left(z_1 = z_2 \right) = \frac{a}{f_1+f_2-a} = R,\\\label{eqn_P<}
&P_< = \mathbf{Pr}\left(z_1 < z_2\right) =\frac{f_1-a}{f_1+f_2-a},\\\label{eqn_P>}
&P_> = \mathbf{Pr}\left(z_1 > z_2\right) =\frac{f_2-a}{f_1+f_2-a}
\end{align}
These probabilities are easy to understand. For example, for the event $\{z_1 < z_2\}$,  the size of sample space is $|S_1\cup S_2| = f_1+f_2-a$ and the size of event space is $|S_1 - S_1\cap S_2| =f_1 -a $, and hence $P_<  =\frac{f_1-a}{f_1+f_2-a}$.

We will show that the estimator solely based on $P_=$ (\ref{eqn_P=}) is optimal only when $f_1=f_2$. Assuming $f_1\geq f_2$, then (\ref{eqn_P<}) should not be used for the estimation task. The estimator based on $P_>$ (\ref{eqn_P>}) is superior to $P_=$ (\ref{eqn_P>}) when $f_1\geq f_2 \approx a$. However, since we do not know $a$ in advance, we must combine all three probabilities to ensure  accurate estimates.

\vspace{-0.05in}
\subsection{The Measure of Containment}\vspace{-0.05in}

 The ratio  $T = a/f_2$ (assuming $f_1\geq f_2$) is known as the {\em containment}.  It is possible that the resemblance $R$ is small but the containment $T$ is large. Note that $R = \frac{a}{f_1+f_2-a}\leq a/f_1\leq f_2/f_1$. Thus, if, for example, $f_2/f_1\leq 0.2$, then R has to be small, even when $a\approx f_2$ (which corresponds to $T\approx 1$).

 While the literature on minwise hashing has mainly focused on the estimation of set resemblance, accurate estimation of set containment is also crucial to a number of different applications. For example, \cite{Proc:Dasu_SIGMOD02} uses both resemblance and containment estimates of the $w$-grams contained in text columns to characterize the similarity of database table contents in a tool that allows users to quickly understand database content. In a similar context, \cite{Zhang:2010:MFK:1920841.1920944} tests the (estimated) level of containment between the distinct values contained in different  (sets of) database columns to automatically detect foreign key constraints. \cite{Ouyang:2002:CDC:645962.674230} describes the use of (estimated) shingle containment in the context of cluster-based compression schemes.
In the context of overlay networks, \cite{Byers:2002:ICD:633025.633031} uses the estimated containment (and resemblance) of the working sets of peers to coordinate between them, in turn reducing communication cost and complexity; because only small messages should be passed for coordination, this estimation has to be based on small synopses. The use of containment estimates in the context of peer-to-peer networking is discussed in ~\cite{Gupta02approximaterange}.

\vspace{-0.05in}
\subsection{b-Bit Minwise Hashing}\vspace{-0.05in}

The recent development of {\em b-bit minwise hashing}~\cite{Proc:Li_Konig_WWW10,Proc:Li_Konig_NIPS10,Article:Li_Konig_CACM11,Report:HashLearning11} provides a solution to the (storage and computational) problem of {\em minwise hashing} by storing only the lowest b bits (instead of 64 bits) of each hashed value for a small b. \cite{Proc:Li_Konig_WWW10} proved that using only $b=1$ bit per hashed value can achieve at least a 21.3-fold improvement (in terms of storage) compared to using $b=64$ bits if the target resemblance $R>0.5$. This is a very encouraging result which may lead to substantial improvement in applications like (near)duplicate detection of Web pages~\cite{Proc:Broder}.

On the other hand, when $R$ is small, as shown in \cite{Proc:Li_Konig_WWW10,Report:HashLearning11}, one might have to increase $b$ in order to achieve an adequate accuracy without substantially increasing $k$, the number of permutations.

In fact, machine learning algorithms like SVM require (essentially) all pairwise similarities and it is expected that most pairs are not too similar. Our concurrent work~\cite{Report:HashLearning11} attempts to combine linear SVM~\cite{Proc:Joachims_KDD06,Proc:Shalev-Shwartz_ICML07,Article:Fan_JMLR08,Proc:Hsieh_ICML08,Proc:Yu_KDD10} with b-bit hashing; and our initial experiments suggest that $b\geq4$ (especially $b=8$) is needed to achieve good  performance.

In this paper, we will provide estimators for both the standard minwise hashing and b-bit minwise hashing.

\vspace{-0.1in}
\section{Estimators for Minwise Hashing}\vspace{-0.1in}

 Consider  two sets $S_1, S_2\in\Omega=\{0, 1, 2, ..., D-1\}$. $f_1 = |S_1|$, $f_2 = |S_2|$. $a = |S_1\cap S_2|$. We apply $k$ random permutations $\pi_j: \Omega \rightarrow \Omega$, and record the minimums $z_{1,j} = \min(\pi_j(S_1)$, $z_{2,j} = \min(\pi_j(S_2)$, $j=1$ to $k$. We will utilize the sizes of three disjoint sets:
 \begin{align}
 k_=  &= |\{z_{1,j} = z_{2,j}, j = 1, 2,..., k\}|\\
 k_<  &= |\{z_{1,j} < z_{2,j}, j = 1, 2,..., k\}|\\
 k_>  &= |\{z_{1,j} > z_{2,j}, j = 1, 2,..., k\}|
 \end{align}
Note  that $E(k_=) = kP_=$, $E(k_<) = kP_<$, $E(k_>) = kP_>$,  $Var(k_=) = kP_=(1-P_=)$, etc. Thus, $\frac{k_=}{k}$, $\frac{k_<}{k}$, and $\frac{k_>}{k}$ are unbiased estimators of $P_=$ (\ref{eqn_P=}), $P_<$ (\ref{eqn_P<}), and $P_>$ (\ref{eqn_P>}), respectively.   For the convenience of presentation, we estimate the intersection $a = (f_1+f_2)\frac{R}{1+R}$:
\begin{align}
\hat{a}_= &= (f_1+f_2) \frac{k_=/k}{1+k_=/k} = \frac{(f_1+f_2)k_=}{k+k_=}\\
\hat{a}_< &=f_1 - f_2 \frac{k_<}{k-k_<}\\
\hat{a}_> &=f_2 - f_1 \frac{k_>}{k-k_>},
\end{align}
which are  asymptotically (for large $k$) unbiased estimators of $a$. The variances are provided by Lemma~\ref{lem_est=<>}.
\begin{lemma}\label{lem_est=<>}
\begin{align}
&Var(\hat{a}_=) = \frac{1}{k}\frac{(f_1+f_2-a)^2a(f_1+f_2-2a)}{(f_1+f_2)^2} + O\left(\frac{1}{k^2}\right)\\
&Var(\hat{a}_<) = \frac{1}{k}\frac{(f_1+f_2-a)^2(f_1-a)}{f_2} + O\left(\frac{1}{k^2}\right)\\
&Var(\hat{a}_>) = \frac{1}{k}\frac{(f_1+f_2-a)^2(f_2-a)}{f_1} + O\left(\frac{1}{k^2}\right)
\end{align}
\noindent \textbf{Proof}: The asymptotic variances can be computed by the ``delta method'' $Var(g(x)) \approx Var(x)\left[g^\prime(E(x))\right]^2$ in a straightforward fashion. We skip the details.$\Box$
\end{lemma}

\vspace{-0.05in}
\subsection{The Maximum Likelihood Estimator}\vspace{-0.05in}

Lemma~\ref{lem_est=<>} suggests that the current standard estimator $\hat{a}_=$ may be severely less optimal when $f_2/f_1$ deviates from 1. In fact, if we know $f_1>f_2 \approx a$ (i.e., when the resemblance is small but the containment is large), we will obtain good results by using $\hat{a}_>$. The problem is that we do not know $a$ in advance and hence we should resort to the maximum likelihood estimator (MLE).

\begin{lemma}\label{lem_MLE}
The MLE, denoted by $\hat{a}_{MLE}$, is the solution to the following  equation:
\begin{align}\label{eqn_MLE}
k_=\frac{f_1+f_2}{a}- k_< \frac{f_2}{f_1-a} - k_>\frac{f_1}{f_2-a}=0
\end{align}
which is asymptotically unbiased with the variance
\begin{align}\label{eqn_MLE_var}
Var\left(\hat{a}_{MLE}\right) =& \frac{1}{k}\frac{(f_1+f_2-a)^2}{\frac{f_1+f_2}{a} +
\frac{f_2}{f_1-a}+ \frac{f_1}{f_2-a}} + O\left(\frac{1}{k^2}\right)
\end{align}
\textbf{Proof:}\ \ The result follows from classical multinomial estimation theory. See Section~\ref{sec_MLE}.$\Box$
\end{lemma}
\vspace{-0.05in}
\subsection{Comparing MLE with Other Estimators}\vspace{-0.05in}

\begin{figure}[h]
\begin{center}
\mbox{
\includegraphics[width = 1.8in]{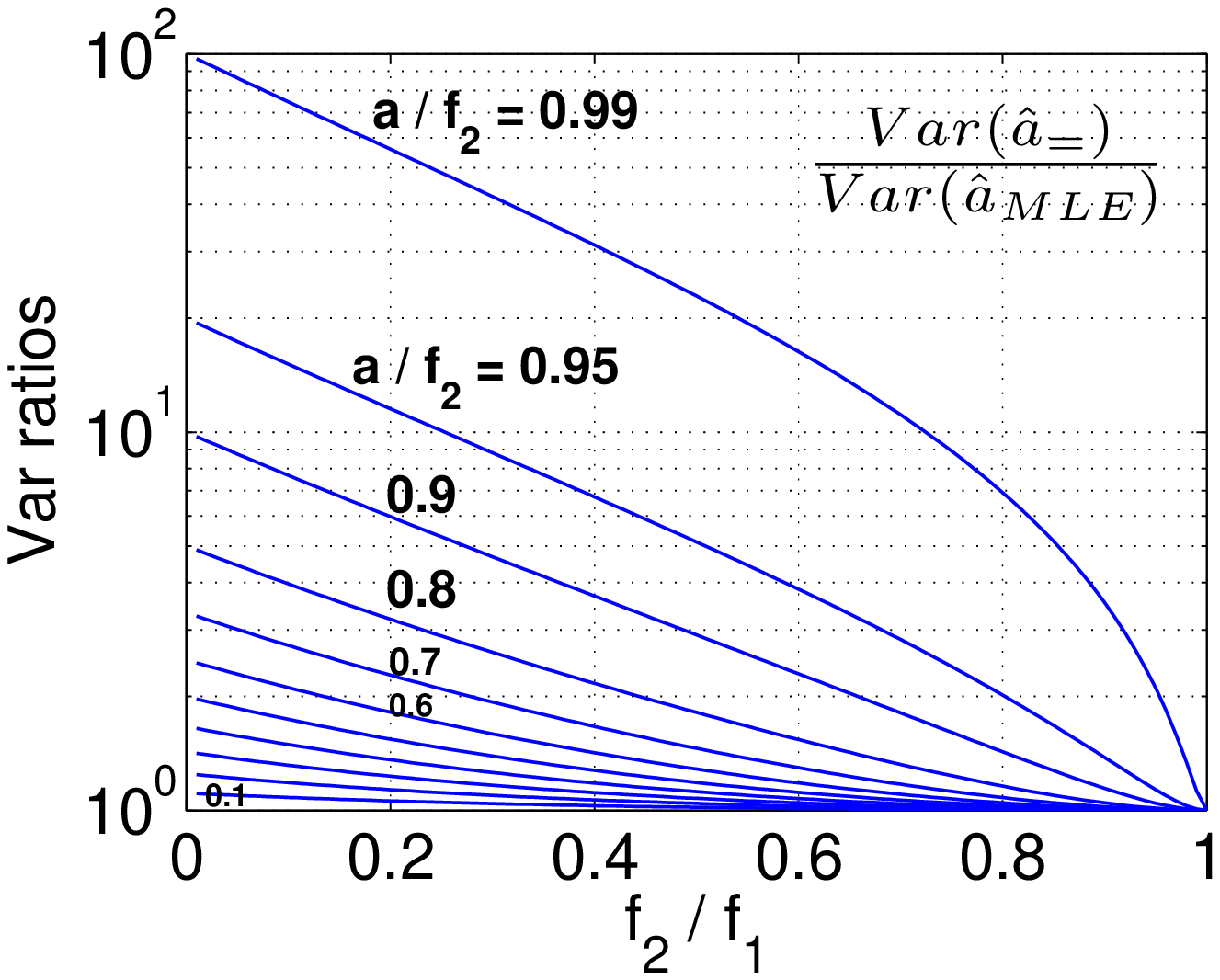}\hspace{-0.2in}
\includegraphics[width = 1.8in]{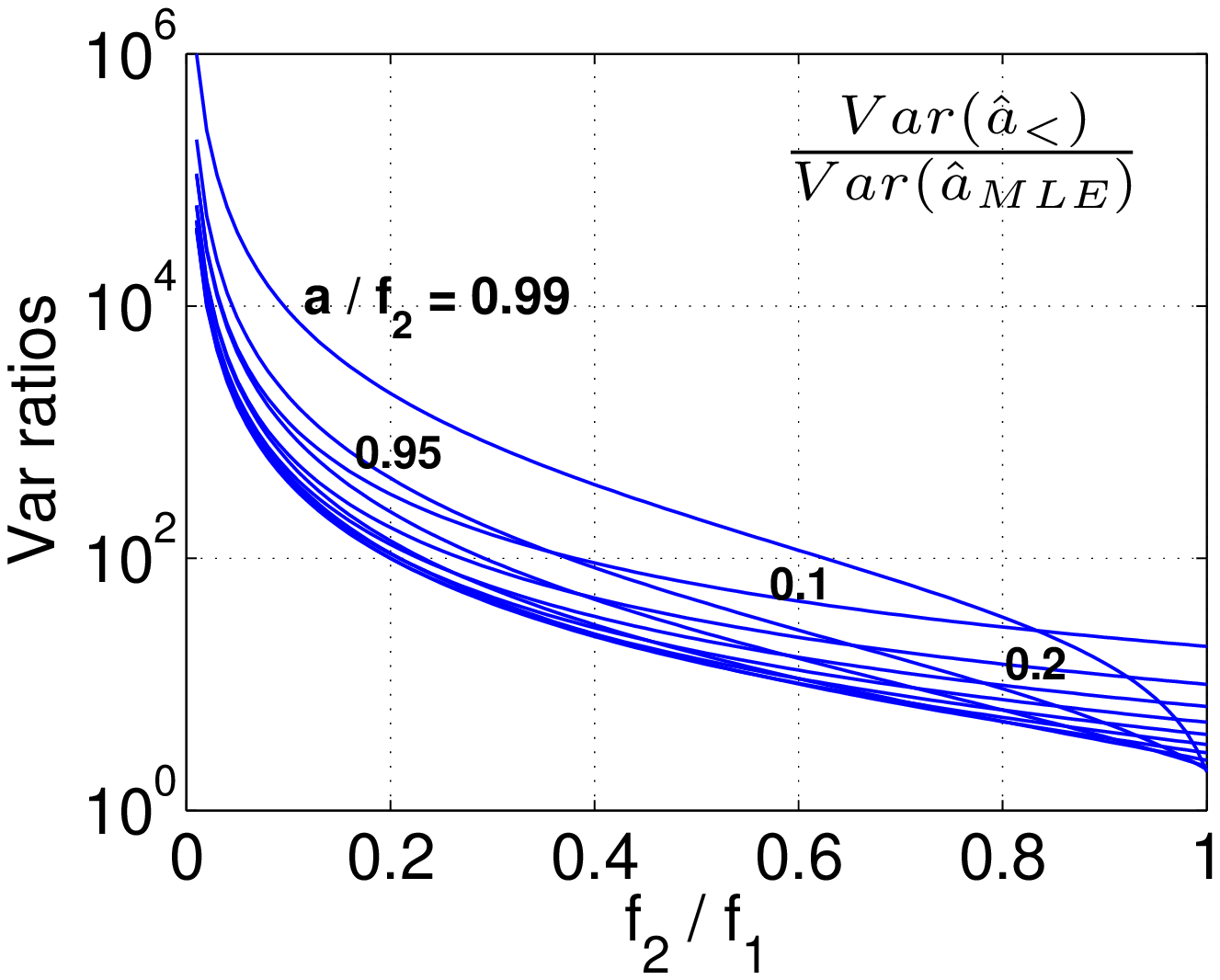}}
\mbox{
\includegraphics[width = 1.8in]{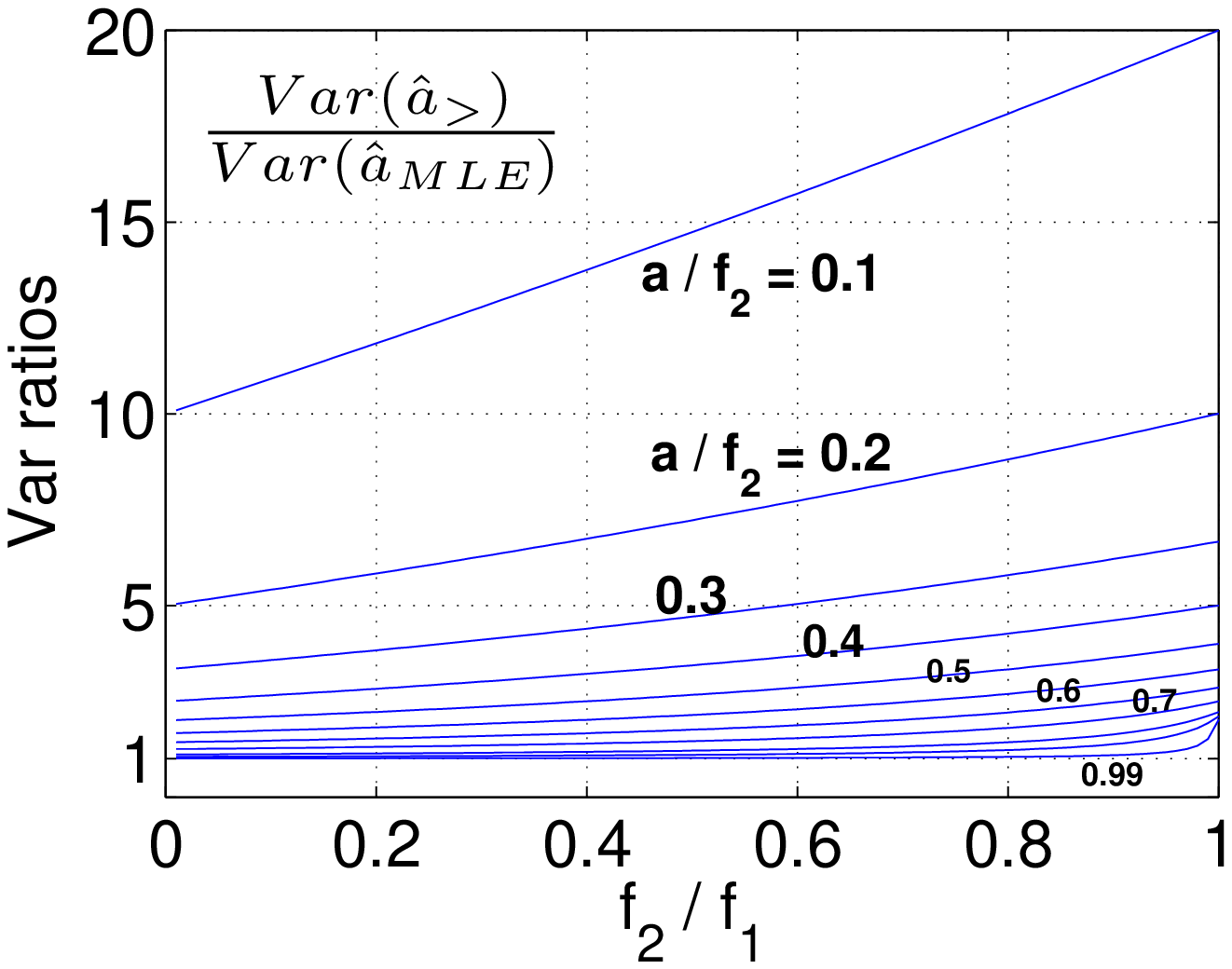}\hspace{-0.2in}
\includegraphics[width = 1.8in]{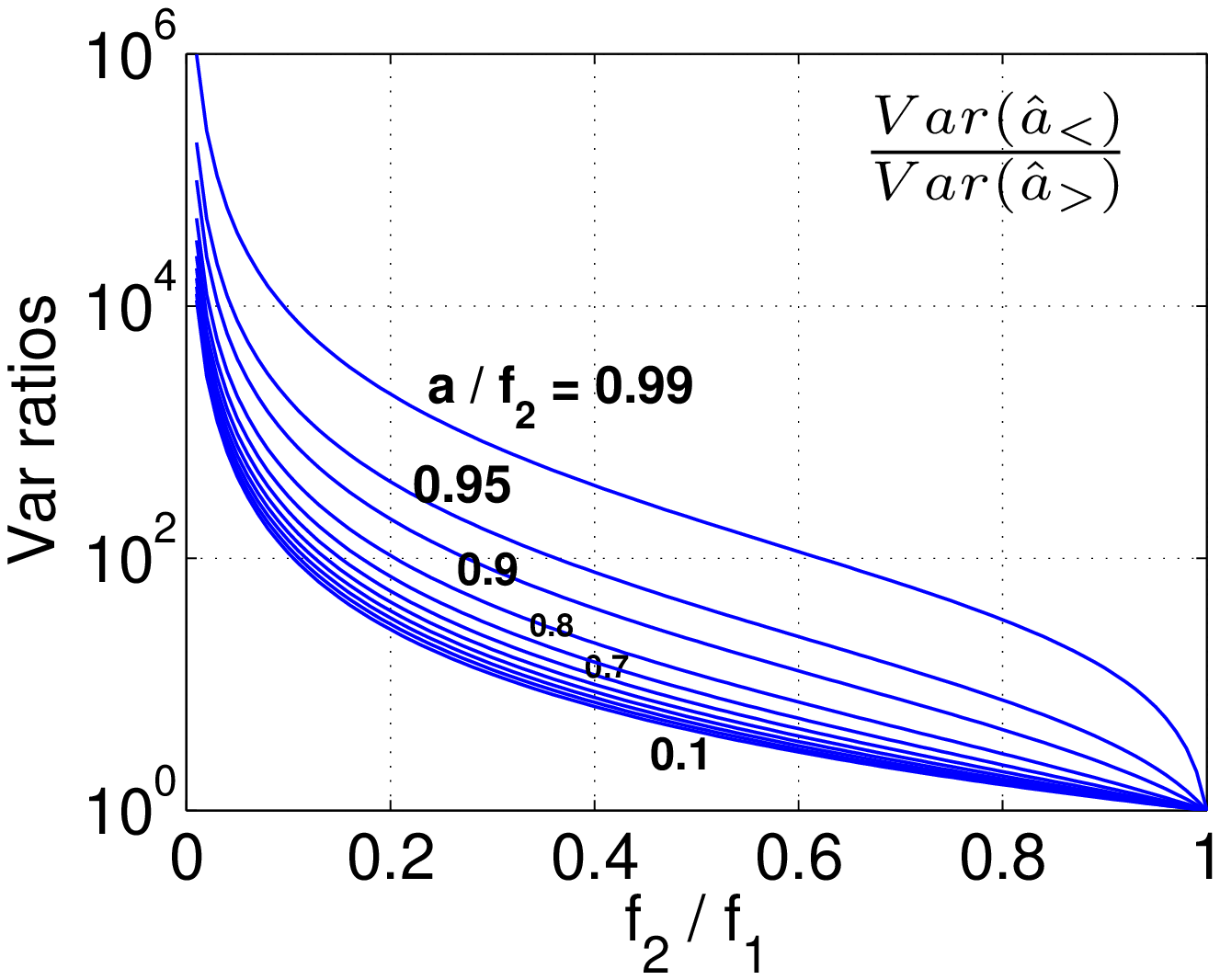}}
\vspace{-.3in}
\caption{Variance ratios (the lower the better). Except the bottom-right panel, we compare the other three estimators, $\hat{a}_=$, $\hat{a}_<$ and $\hat{a}_>$, with the MLE $\hat{a}_{MLE}$.
}\label{fig_var_a_MLE}
\end{center}

\end{figure}

Figure~\ref{fig_var_a_MLE} compares the ratios of the variances of  estimators of $a$ (only using the $O\left(\frac{1}{k}\right)$ term of the variance). The top-left panel illustrates that when $f_2/f_1<0.5$ (which is common), the MLE  $\hat{a}_{MLE}$ can reduce the variance of the standard estimator $\hat{a}_{=}$ by a large factor. When the target containment $T=\frac{a}{f_2}$ approaches 1, the improvement can be as large as  100-fold.

The top-right panel of Figure~\ref{fig_var_a_MLE} suggests that, if $f_2\leq f_1$, then we should not use $\hat{a}_<$, because its variance  can be  magnitudes larger than the variance of the MLE. The bottom-left panel confirms that if we know the containment is very large (close to 1), then we will do  well by using $\hat{a}_>$ which is simpler than the MLE. The problem is of course that we do not know $a$ in advance and hence we may still have to use the MLE. The bottom-right panel  verifies that $\hat{a}_{>}$ is significantly better  $\hat{a}_{<}$.

%

\vspace{-0.05in}
\subsection{Experiment}\vspace{-0.05in}

For the purpose of verifying the theoretical improvements, we use two pairs of sets corresponding to the occurrences of four common words (``A -- TEST'' and  ``THIS -- PERSON'') in a chunk of real world Web crawl data. Each (word) set is a set of document (Web page) IDs which contained that word at least once. For ``A -- THE'', the resemblance = 0.0524 and containment = 0.9043. For `THIS -- PERSON'', the resemblance = 0.0903 and containment = 0.8440.

\begin{figure}[h]
\begin{center}
\mbox{
\includegraphics[width = 1.8in]{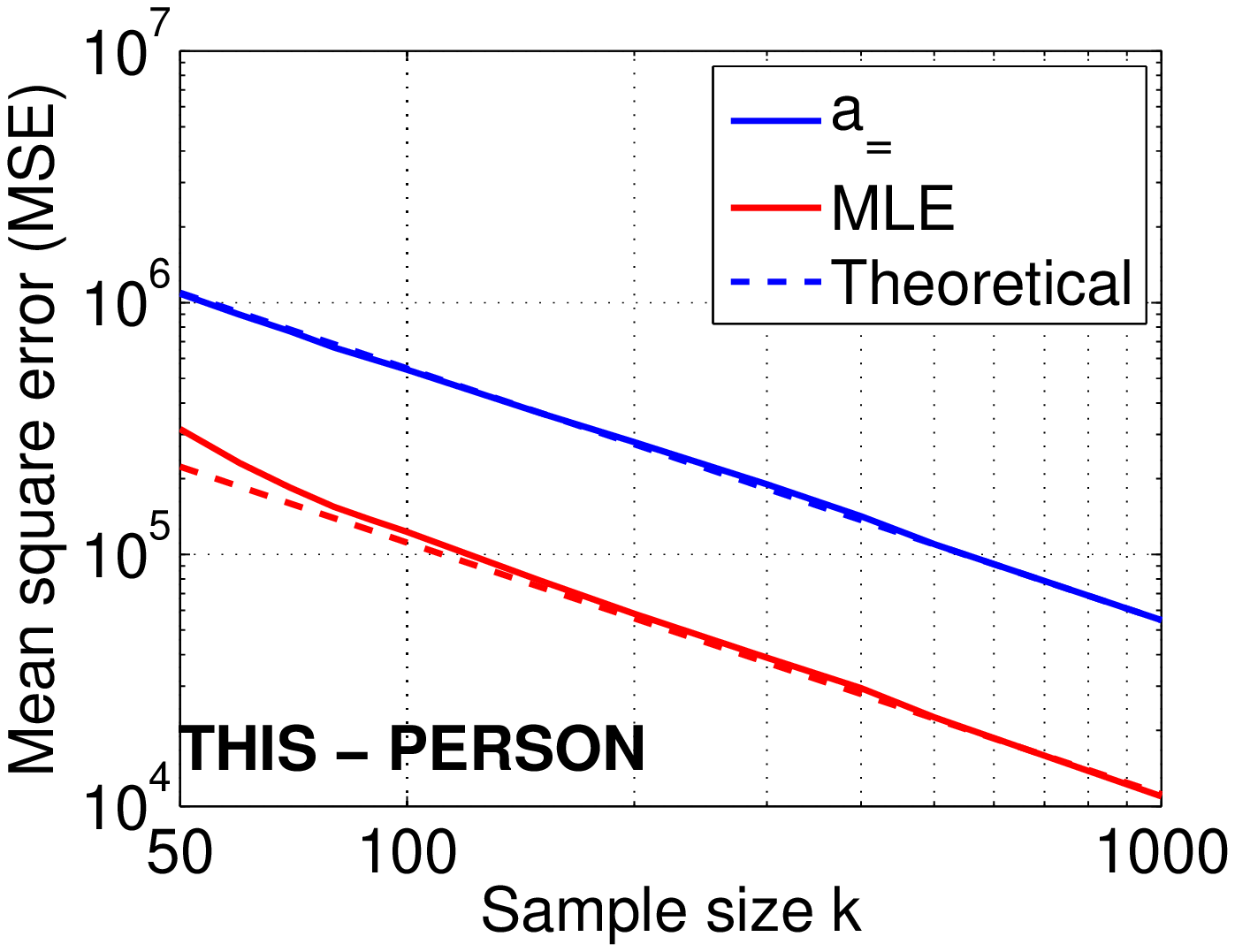}\hspace{-0.2in}
\includegraphics[width = 1.8in]{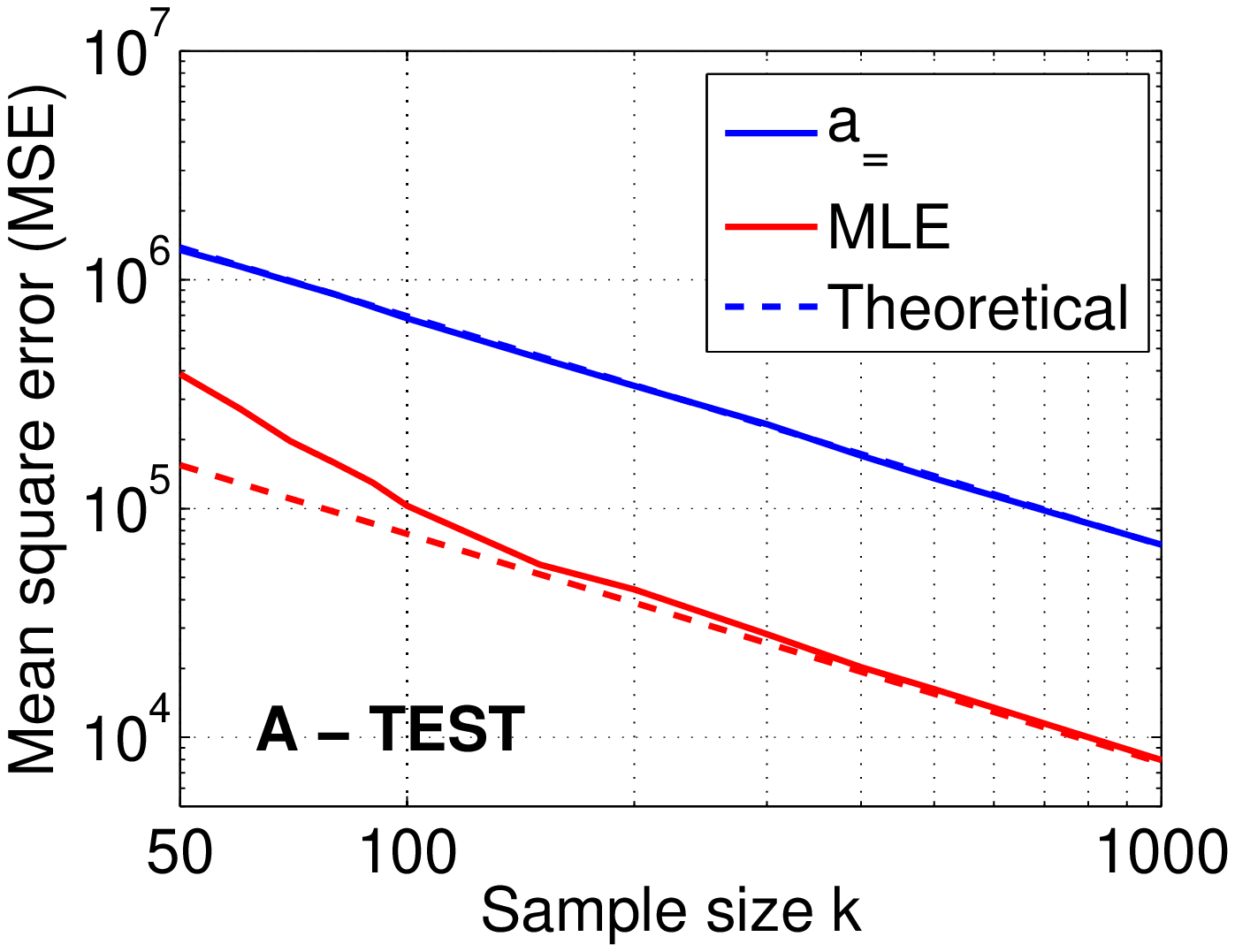}}
\end{center}
\vspace{-0.2in}
\caption{A simulation study using two pairs of real-world  vectors (of low resemblance and high containment) to verify (i) $\hat{a}_{MLE}$ is significantly better than $\hat{a}_=$; and (ii) the theoretical variances match the simulations and the  bias of the MLE vanishes as $k$ increases.
}\label{fig_mse_minwise}

\end{figure}

Figure~\ref{fig_mse_minwise} presents the mean square errors (MSE) of the estimates using $\hat{a}_=$ and $\hat{a}_{MLE}$. The results verify our theoretical predictions:
\begin{itemize}
\item For pairs of low resemblance and high containment, the MLE $\hat{a}_{MLE}$ provides significantly better (in these two cases, about an order of magnitude better) results than the standard estimator $\hat{a}_=$.
\item The MLE is asymptotically unbiased. The small bias at small $k$ (which is common for MLE in general) vanishes as $k$ increases.
\item The theoretical variances match the simulations.
\end{itemize}

\vspace{-0.15in}
\section{$b$-Bit Minwise Hashing}\vspace{-0.1in}

b-Bit minwise hashing~\cite{Proc:Li_Konig_WWW10} stores each hashed value, e.g., $z_1 = \min(\pi(S_1))$, $z_2 = \min(\pi(S_2))$, using the lowest $b$ bits instead of 64 bits. In this section, we will show that because the original b-bit minwise hashing only used part of the available information, it can be substantially improved.

We first define:

$u_{1,b}$ = the number formed by the lowest $b$ bits of $z_1$

$u_{2,b}$ = the number formed by the lowest $b$ bits of $z_2$.

\cite{Proc:Li_Konig_WWW10} derived the probability formula $\mathbf{Pr}\left(u_{1,b} = u_{2,b}\right)$ by assuming $D=|\Omega|$ is large (which is virtually always satisfied in practice). We will also need to derive
\begin{align}\notag
P_{b, (t,d)} = \mathbf{Pr}\left(u_{1,b} = t, u_{2,b} = d\right),\ \ \ t, d = 0, 1, 2, ..., 2^b-1
\end{align}

We follow the convention in \cite{Proc:Li_Konig_WWW10} by defining
\begin{align}
r_1 = \frac{f_1}{D},\ \ r_2 = \frac{f_2}{D},\ \ s = \frac{a}{D}
\end{align}

Instead of estimating $a$, we equivalently estimate $s$ in the context of b-bit hashing.  Lemma \ref{lem_Ptd} provides the probability formulas as the basic tool.
\begin{lemma}\label{lem_Ptd}
Assume $D$ is very large.
\begin{align}
&\mathbf{Pr}\left(u_{1,b} = t, u_{2,b} = d, \  t<d\right)\\\notag
=&P_< \frac{r_2\left[1-r_2\right]^{d-t-1}}{1-\left[1-r_2\right]^{2^b}}
\frac{\left[1-(r_1+r_2-s)\right]^{t}(r_1+r_2-s)}{1-\left[1-(r_1+r_2-s)\right]^{2^b}}\\\notag
+&P_> \frac{r_1\left[1-r_1\right]^{t+2^b-d-1}}{1-\left[1-r_1\right]^{2^b}}
\frac{\left[1-(r_1+r_2-s)\right]^{d}(r_1+r_2-s)}{1-\left[1-(r_1+r_2-s)\right]^{2^b}}\\\notag
\end{align}
\begin{align}
&\mathbf{Pr}\left(u_{1,b} = t, u_{2,b} = d, \  t>d\right)\\\notag
=&P_> \frac{r_1\left[1-r_1\right]^{t-d-1}}{1-\left[1-r_1\right]^{2^b}}
\frac{\left[1-(r_1+r_2-s)\right]^{d}(r_1+r_2-s)}{1-\left[1-(r_1+r_2-s)\right]^{2^b}}\\\notag
+&P_< \frac{r_2\left[1-r_2\right]^{d+2^b-t-1}}{1-\left[1-r_2\right]^{2^b}}
\frac{\left[1-(r_1+r_2-s)\right]^{t}(r_1+r_2-s)}{1-\left[1-(r_1+r_2-s)\right]^{2^b}}\\\notag
\end{align}
\begin{align}
&\mathbf{Pr}\left(u_{1,b} = t, u_{2,b} = t\right) \\\notag
=&\left(R+P_<\frac{r_2\left[1-r_2\right]^{2^b-1}}{1-\left[1-r_2\right]^{2^b}}+P_>\frac{r_1\left[1-r_1\right]^{2^b-1}}{1-\left[1-r_1\right]^{2^b}}\right)\times\\\notag
+&\frac{\left[1-(r_1+r_2-s)\right]^{t}(r_1+r_2-s)}{1- \left[1-(r_1+r_2-s)\right]^{2^b}}
\end{align}
\noindent \textbf{Proof:}\ \ See Appendix \ref{proof_lem_Ptd}.$\Box$
\end{lemma}

Therefore, we encounter a multinomial probability estimation problem with each cell probability being a function of $s$. Note that the total number of cells, i.e., $2^b\times 2^b$, is large especially when $b$ is not small.

In addition to $P_{b, (t,d)}$, we also define the following three probability summaries analogous to $P_=$, $P_<$, and $P_>$.
\begin{align}\notag
&P_{b,=} = \mathbf{Pr}\left(u_{1,b} = u_{2,b}\right) = \sum_{t=0}^{2^b-1}P_{b,(t,t)}\\\notag
&P_{b,<,} = \mathbf{Pr}\left(u_{1,b} < u_{2,b}\right) =\sum_{t<d}^{2^b-1}P_{b,(t,d)}\\\notag
&P_{b,>} = \mathbf{Pr}\left(u_{1,b} > u_{2,b}\right)=\sum_{t>d}^{2^b-1}P_{b,(t,d)}
\end{align}

Suppose we conduct $k$ permutations. We  define the observed counts, $k_{b,(t,d)}$, $k_{b,=}$, $k_{b,<}$, and $k_{b,>}$, which correspond to $P_{b,(t,d)}$,  $P_{b,=}$, $P_{b,<}$, and $P_{b,>}$, respectively. Note that $k = \sum_{t,d} k_{b,(t,d)}$.

\cite{Proc:Li_Konig_WWW10} only used $P_{b,=}$ to estimate $R$ (and hence also $s$). We expect to achieve substantial improvement if we can take advantage of the matrix of probabilities $P_{b,(t,d)}$. Here, we first review some basic statistical procedure for multinomial estimation and the classical (asymptotic) variance analysis.

\vspace{-0.05in}
\subsection{Review Classical Multinomial Estimation}\label{sec_MLE}
\vspace{-0.05in}

Consider a  table with $m$ cells, each of which is associated with a probability $q_i(\theta)$, $i= 1, 2, ..., m$. Here we assume the probability $q_i$ is parameterized by $\theta$ (for example, the $s$ in our problem), and the task is to estimate $\theta$. Suppose we draw $k$ i.i.d. samples and the number of observations from the $i$-th cell is $k_i$, $\sum_{i=1}^m k_i = k$. The joint log-likelihood is proportional to
\begin{align}\label{eqn_MLE}
l(\theta) = \sum_{i}^m k_i \log q_i(\theta).
\end{align}
The maximum likelihood estimator (MLE), which is optimal or asymptotically (for large $k$) optimal in terms of the variance, is the solution $\hat{\theta}_{MLE}$ to the MLE equation $l^\prime(\theta) = 0$, i.e.,
\begin{align}\label{eqn_d1lik}
l^\prime(\theta) = \sum_{i=1}^m k_i \frac{q_i^\prime(\theta)}{q_i(\theta)} = 0,
\end{align}
solving which often requires a numerical procedure. For one-dimension problems as in our case, the numerical procedure is straightforward.

The estimation variance of $\hat{\theta}_{MLE}$ is related to the Fisher Information $I(\theta) = -E(l^{\prime\prime}(\theta))$:
\begin{align}\notag
&I(\theta) = -E(l^{\prime\prime}(\theta)) = -\sum_{i=1}^m E(k_i) \frac{q_i^{\prime\prime}q_i - \left[q_i^\prime\right]^2}{q_i^2} \\
&=-\sum_{i=1}^m kq_i \frac{q_i^{\prime\prime}q_i - \left[q_i^\prime\right]^2}{q_i^2}
=-k\sum_{i=1}^m  \frac{q_i^{\prime\prime}q_i - \left[q_i^\prime\right]^2}{q_i}\\\notag
&Var\left(\hat{\theta}_{MLE}\right) = \frac{1}{I(\theta)} + O\left(\frac{1}{k^2}\right)\\
&=\frac{1}{k} \frac{1}{\sum_{i=1}^m \frac{\left[q_i^\prime\right]^2}{q_i} - q^{\prime\prime}_i} + O\left(\frac{1}{k^2}\right)
\end{align}

For $b$-bit hashing, since we have $2^b\times 2^b$ cells with probabilities $P_{b,(t,d)}$, we can either use the full (entire) probably matrix or various reduced forms by grouping (collapsing) cells (e.g., $P_{b,=}$, $P_{b,<}$, and $P_{b,>}$) to ease the burden of numerically solving the MLE equation (\ref{eqn_d1lik}).

\subsection{Five Levels of Estimators for $s$}

We first introduce the notation for the following five estimators of $s$:
\begin{enumerate}
\item $\hat{s}_{b,f}$ denotes the full MLE solution by using all $m=2^b\times 2^b$ cell probabilities $P_{b,(t,d)}$, $t,d=0, 1, 2, ..., 2^b-1$. This estimator will be most accurate and computationally most intensive.
\item $\hat{s}_{b,do}$ denotes the MLE solution by using $m=2^b + 2$ cells which include the $2^b$ diagonal probabilities $P_{b,(t,t)}, t = 0, 1, ..., 2^b-1$ and two  summaries of the off-diagonals: $P_{b,<} = \sum_{t<d}P_{b,(t,t)}$ and $P_{b,>} = \sum_{t>d}P_{b,(t,t)}$.

\item $\hat{s}_{b,d}$ denotes the MLE solution by using $m=2^b + 1$ cells which include the $2^b$ diagonal probabilities $P_{b,(t,t)}, t = 0, 1, ..., 2^b-1$ and the sum of the rest, i.e., $P_{b,<}+P_{b,>}$.
\item $\hat{s}_{b,3}$ denotes the MLE solution by using $m=3$ cells which include the sum of the diagonals and two sums
of the off-diagonals, i.e., $P_{b,=} = \sum_{t=0}^{2^b-1} P_{b,(t,t)}$,  $P_{b,<}$, and $P_{b,>}$.

\item $\hat{s}_{b,=}$ denotes the MLE solution by using only $m=2$ cells, i.e., $P_{b,=}$ and $1-P_{b,=}$. This estimator requires no numerical solutions and is the one used in the original b-bit minwise hashing paper~\cite{Proc:Li_Konig_WWW10}.
\end{enumerate}

\begin{figure}[h!]
\begin{center}
\mbox{
\includegraphics[width = 1.8in]{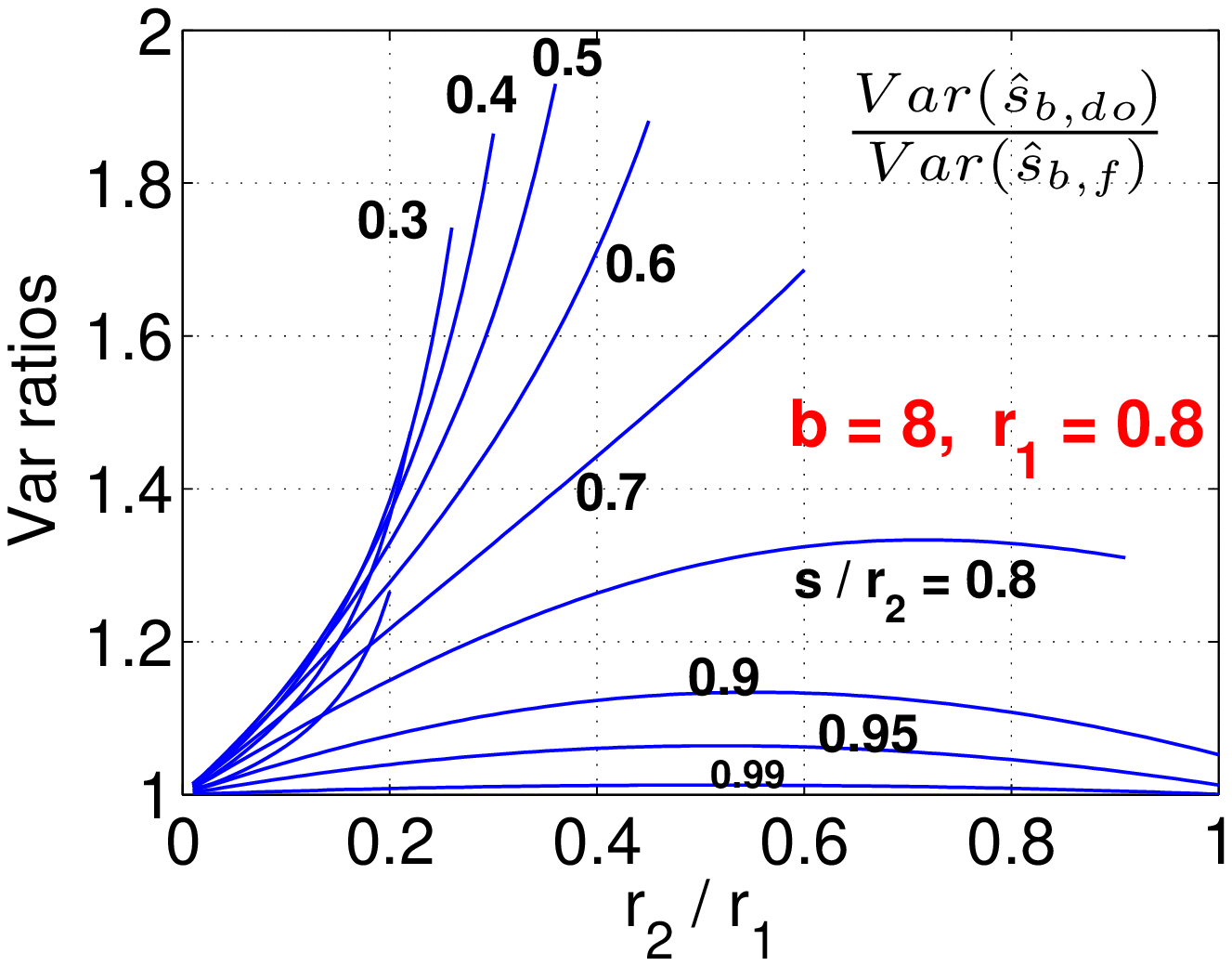}\hspace{-0.2in}
\includegraphics[width = 1.8in]{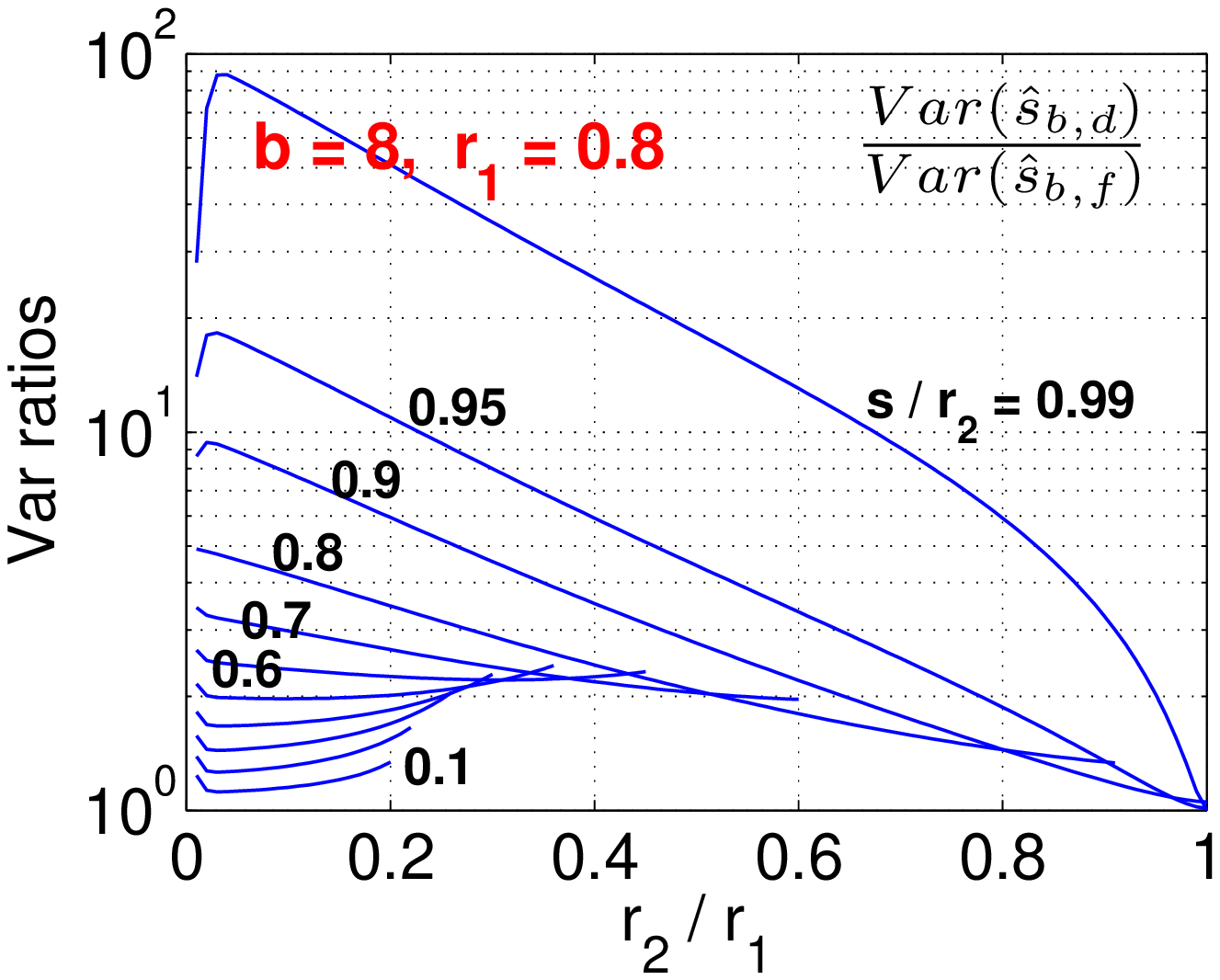}}
\mbox{
\includegraphics[width = 1.8in]{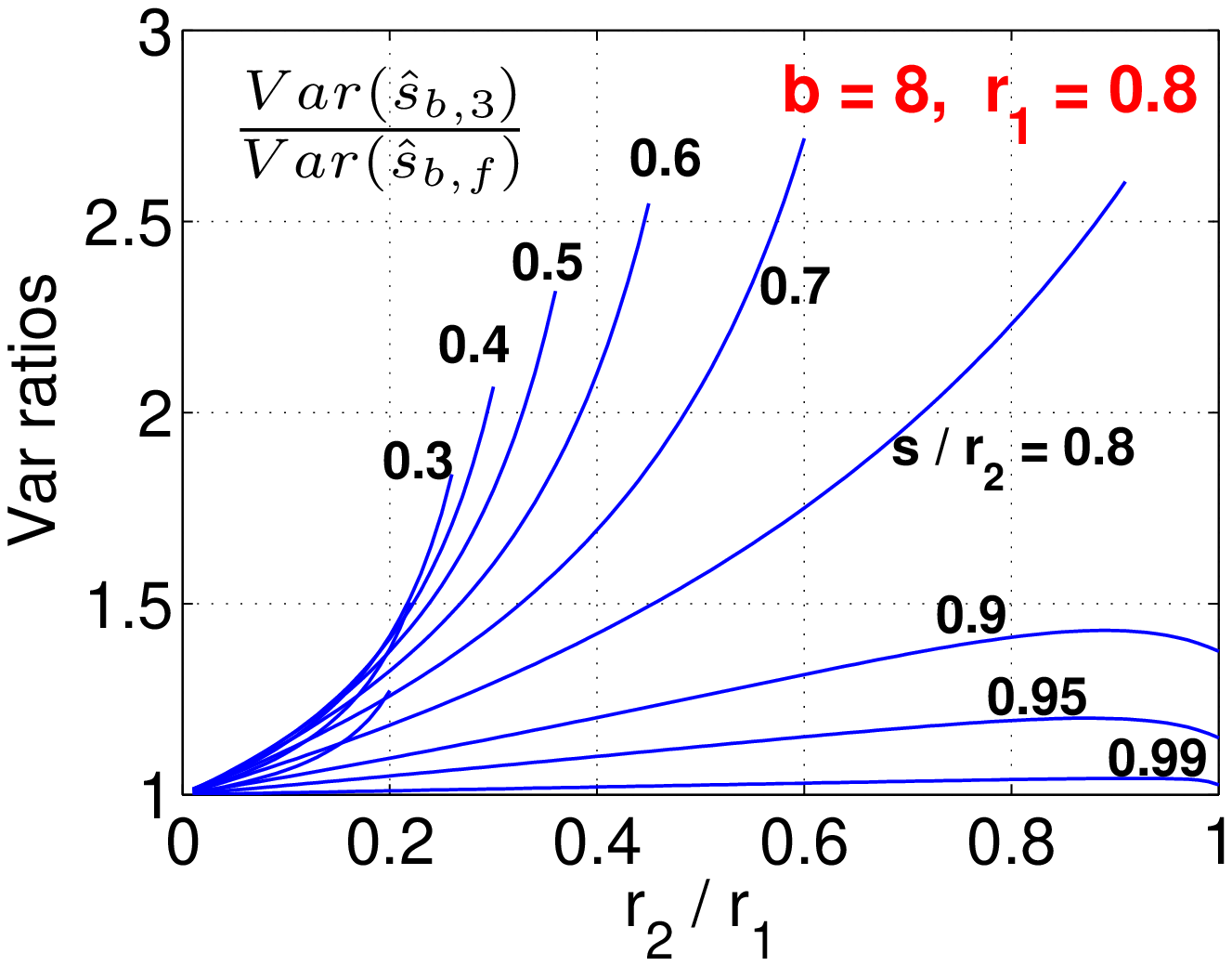}\hspace{-0.2in}
\includegraphics[width = 1.8in]{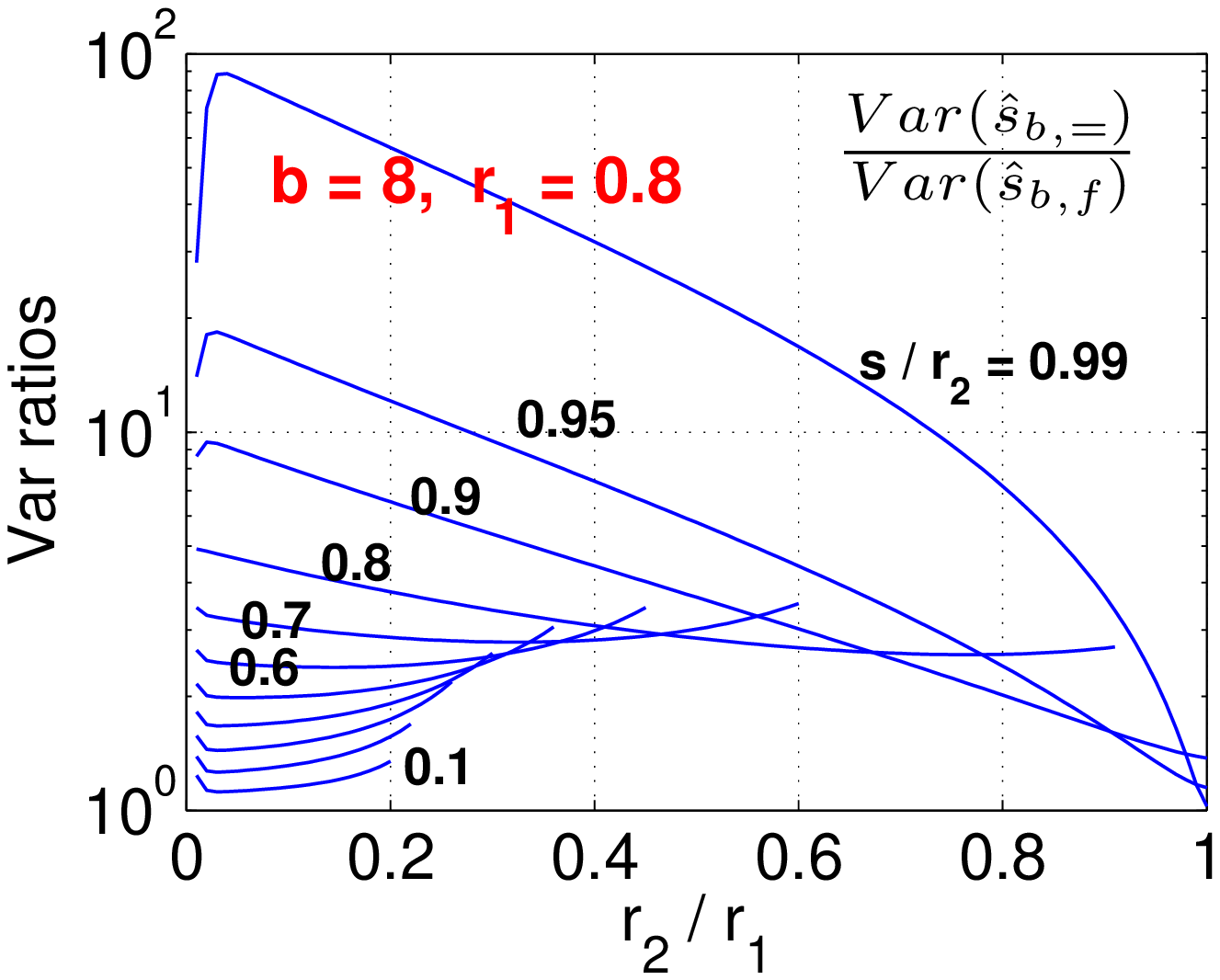}}
\end{center}\vspace{-0.2in}
\caption{Given a contingency table of size $2^b\times 2^b$, we have defined five different estimators. We compare the theoretical variances of the other four estimators, $\hat{s}_{b,do}$, $\hat{s}_{b,d}$, $\hat{s}_{b,3}$, and $\hat{s}_{b,=}$,  with the variance of the full MLE $\hat{s}_{b,f}$. The bottom-right panel measures the real improvement of this paper compared to the previous standard practice (i.e., $\hat{s}_{b,=}$). In this case, the variance ratios of about 10 to 100 are very substantial. The other three panels are for testing whether simpler estimators can still achieve substantial improvements. For example, both $\hat{s}_{b,do}$ and $\hat{s}_{b,3}$ (the left two panels) only magnify the variance of the full MLE by small factors compared to $\hat{s}_{b,=}$, and hence they might be good estimators in lieu of the quite sophisticated full MLE solution. In this figure, we consider $b=8$ and $r_1$ = 0.8. Note that $s / r_2$ is the containment. The resemblance is upper bounded by $r_2 / r_1$.
}\label{fig_var_b8_r08}.
\end{figure}

\begin{figure}[h!]
\begin{center}
\mbox{
\includegraphics[width = 1.8in]{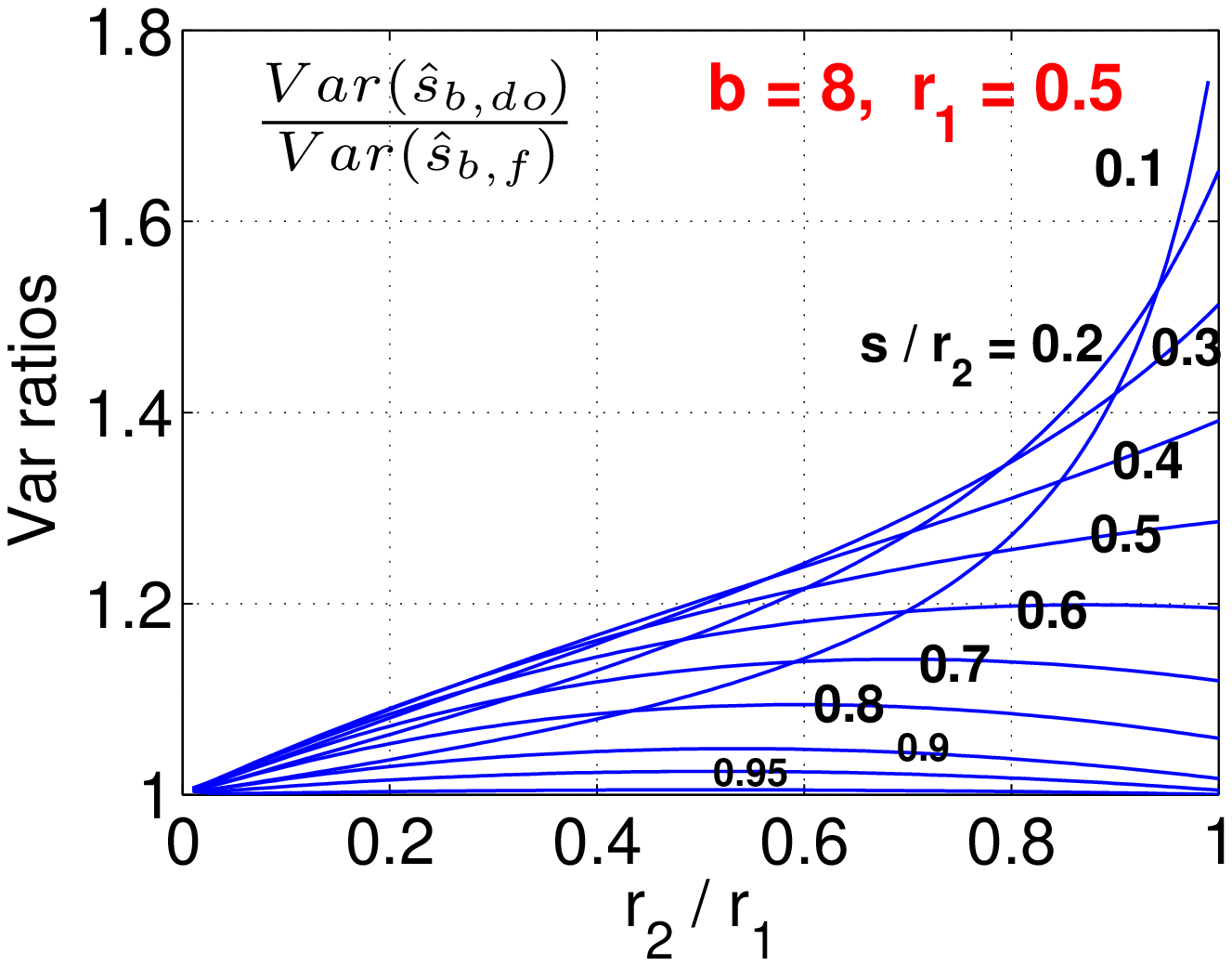}\hspace{-0.2in}
\includegraphics[width = 1.8in]{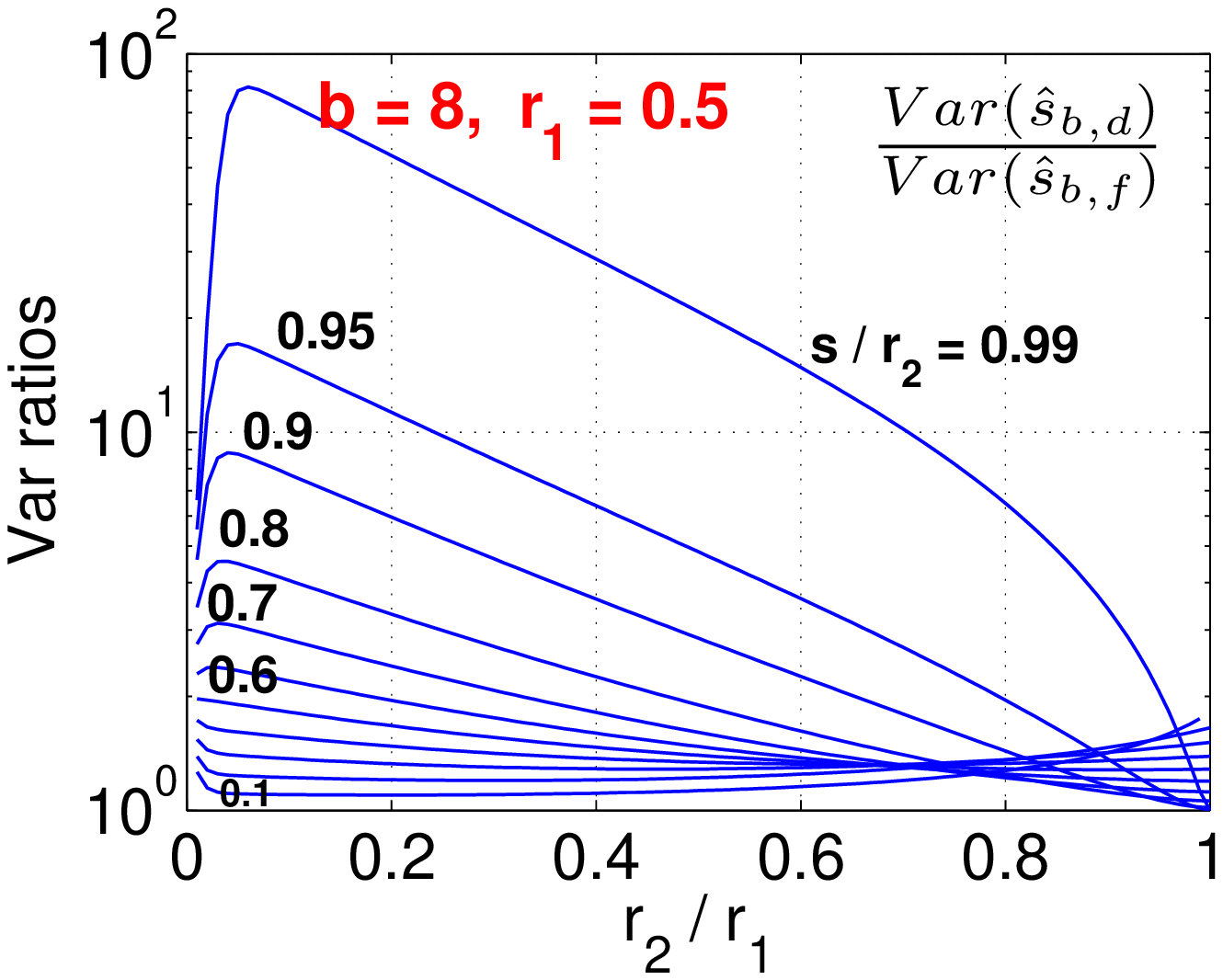}}
\mbox{
\includegraphics[width = 1.8in]{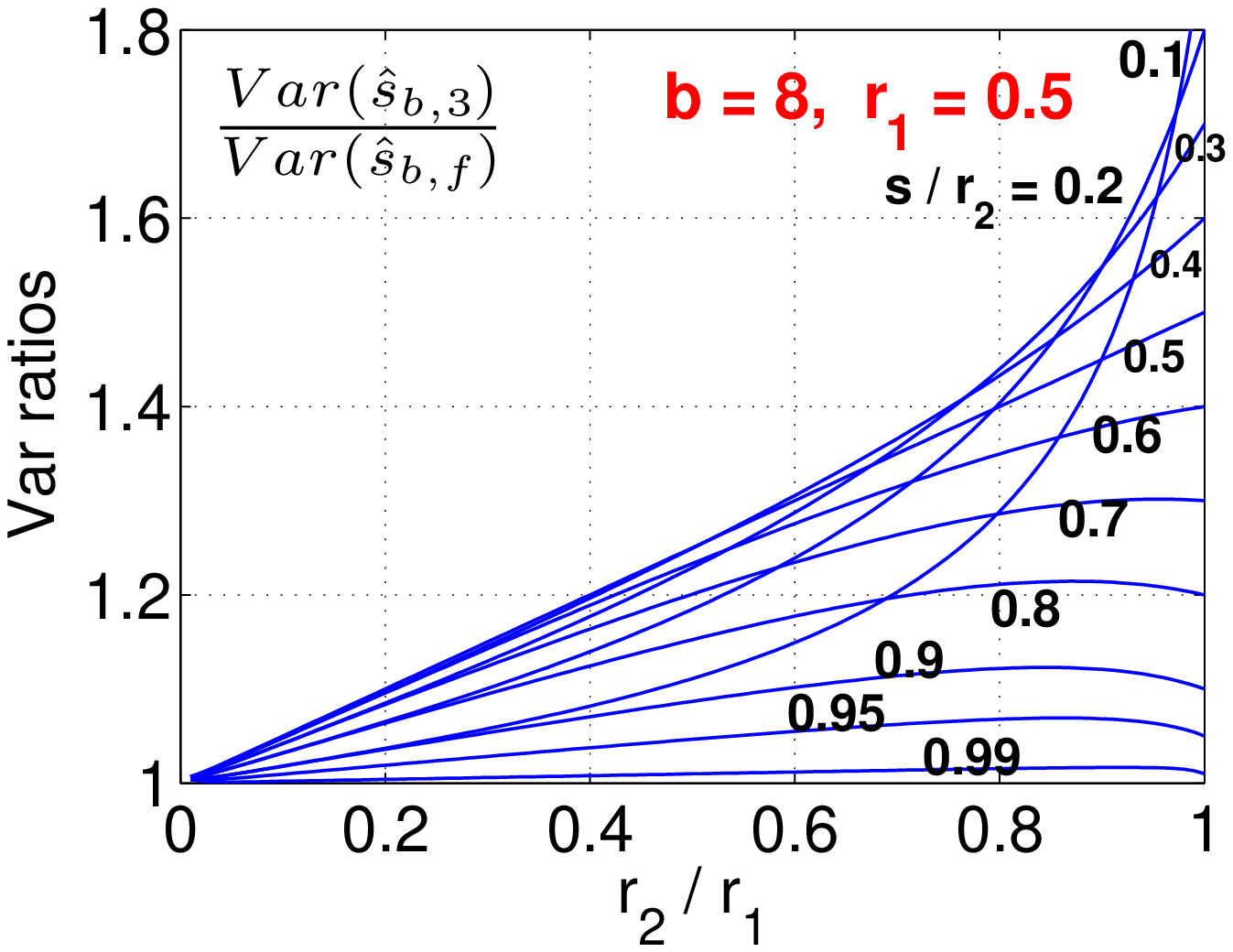}\hspace{-0.2in}
\includegraphics[width = 1.8in]{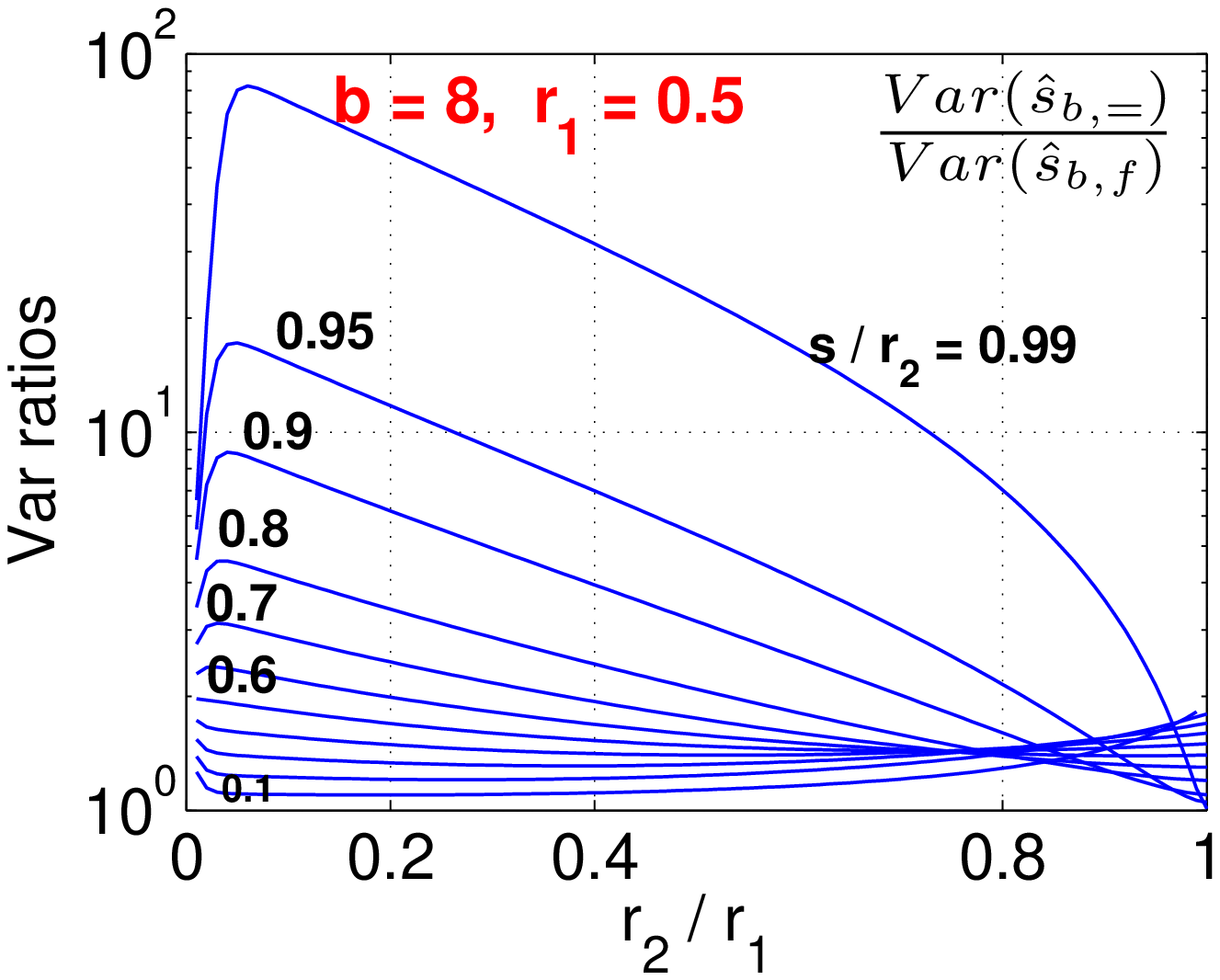}}
\end{center}\vspace{-0.3in}
\caption{Variance ratios for $b=8$ and $r_1=0.5$. See the caption of Figure~\ref{fig_var_b8_r08} for more details. }\label{fig_var_b8_r05}
\end{figure}

\begin{figure}[h!]
\begin{center}
\mbox{
\includegraphics[width = 1.8in]{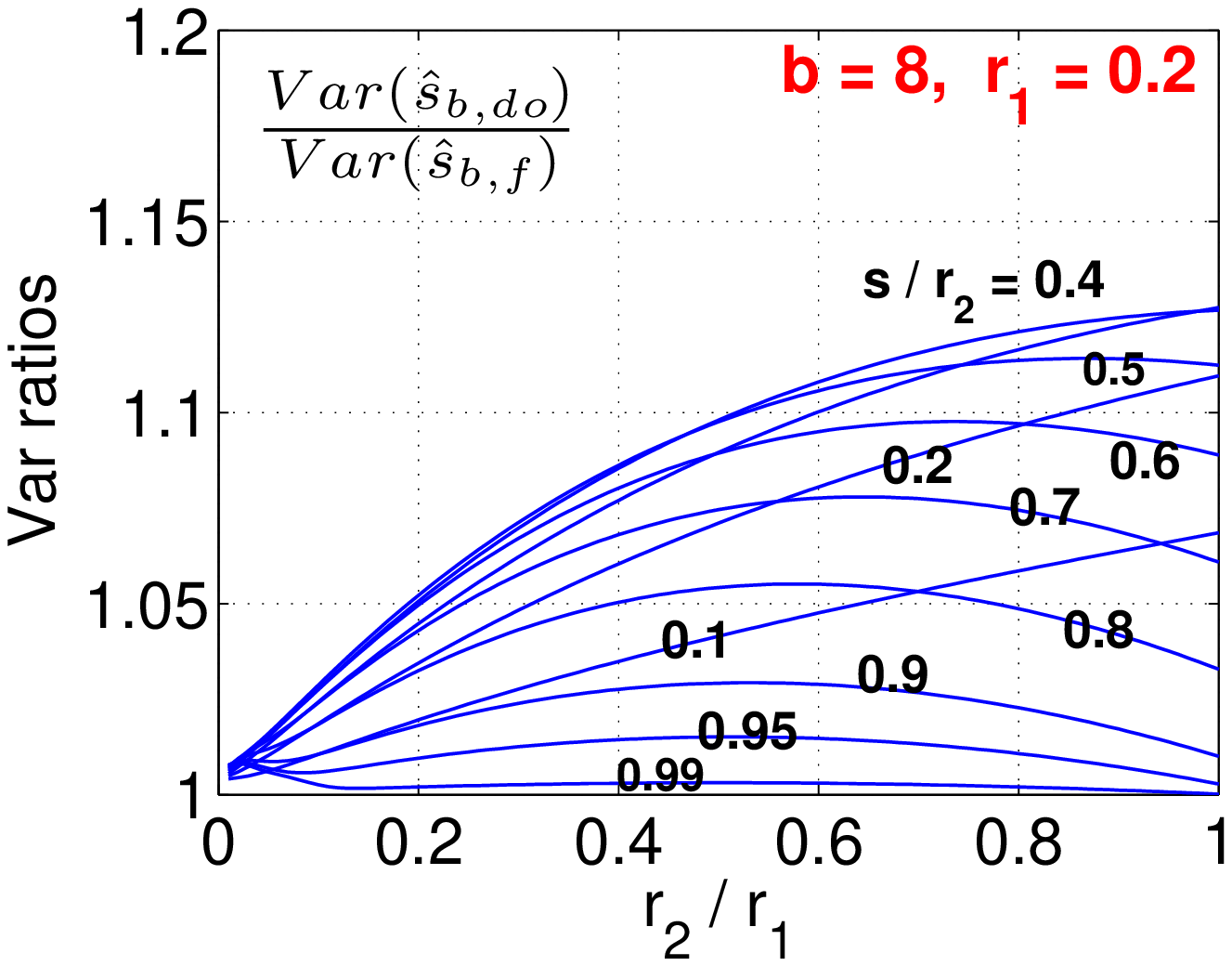}\hspace{-0.2in}
\includegraphics[width = 1.8in]{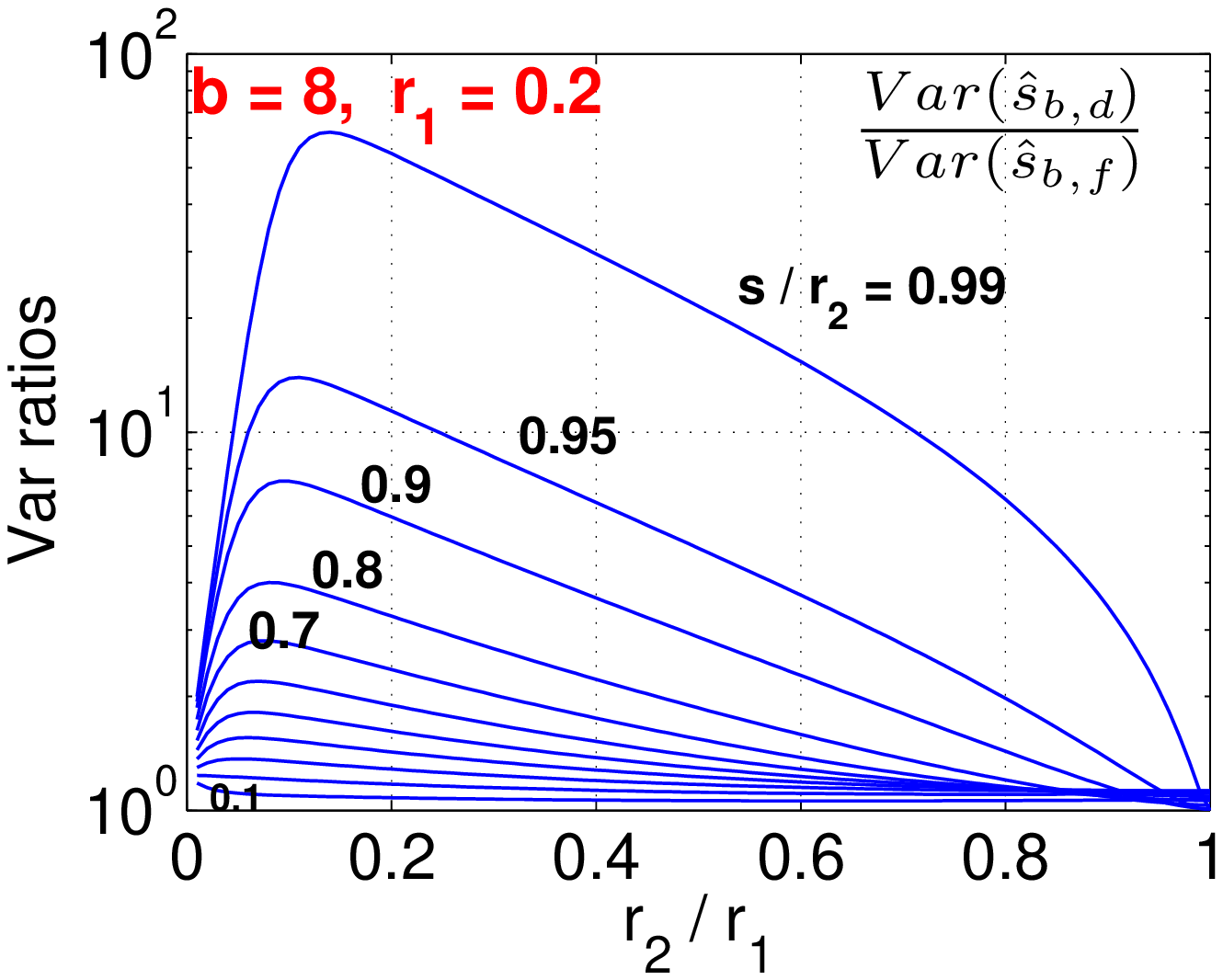}}
\mbox{
\includegraphics[width = 1.8in]{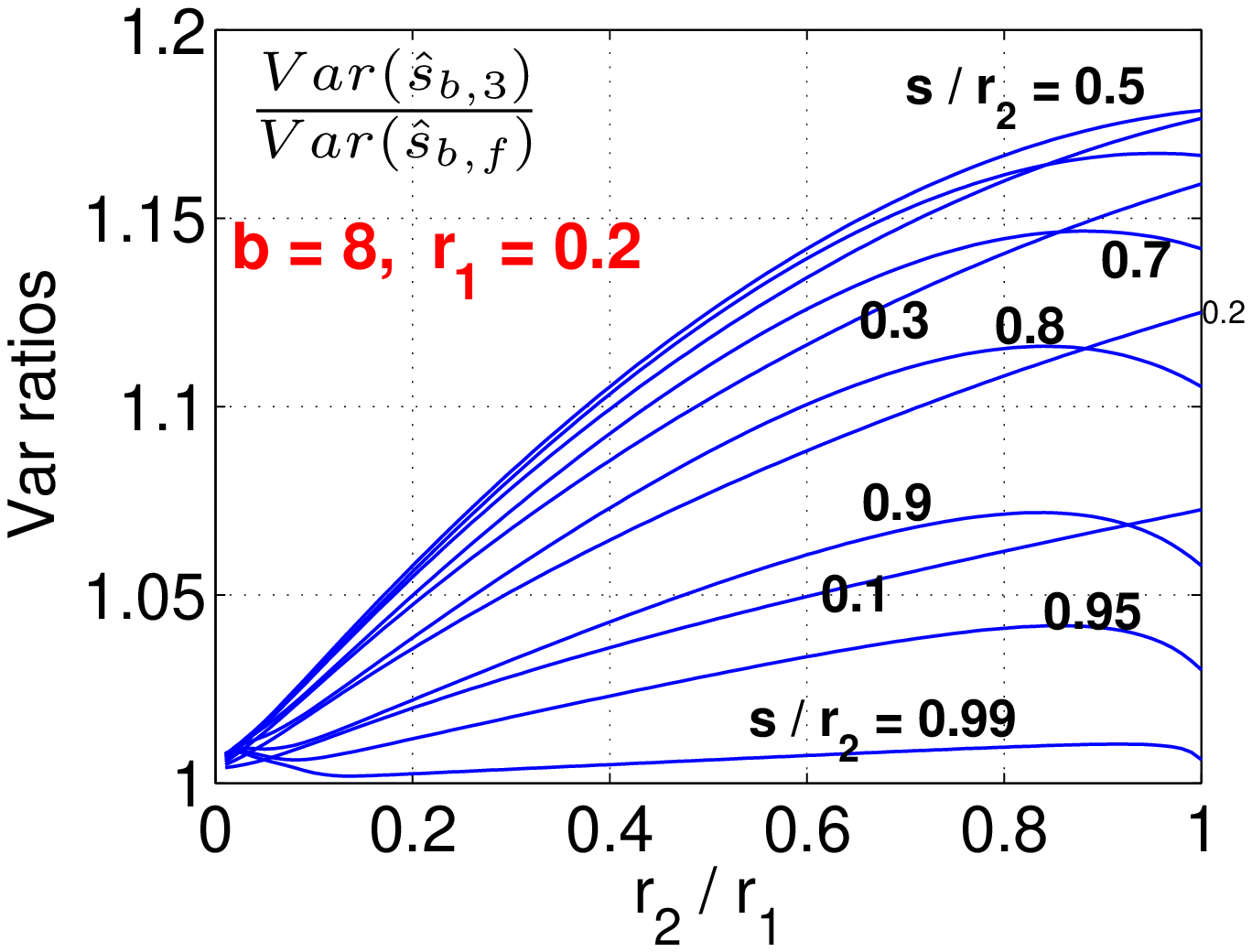}\hspace{-0.2in}
\includegraphics[width = 1.8in]{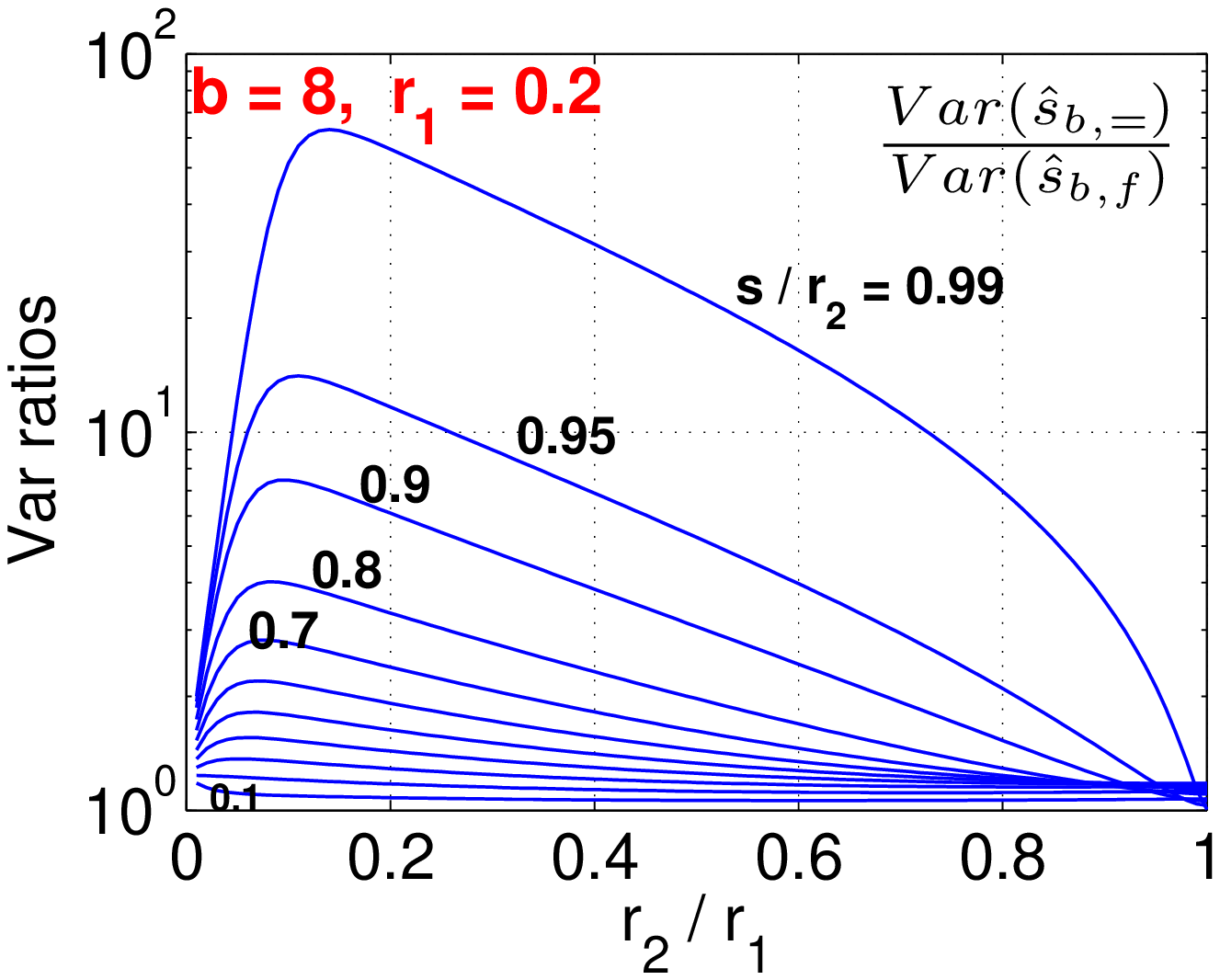}}
\end{center}\vspace{-0.3in}
\caption{Variance ratios for $b=8$ and $r_1=0.2$.}\label{fig_var_b8_r02}
\end{figure}

\begin{figure}[h!]
\begin{center}
\mbox{
\includegraphics[width = 1.8in]{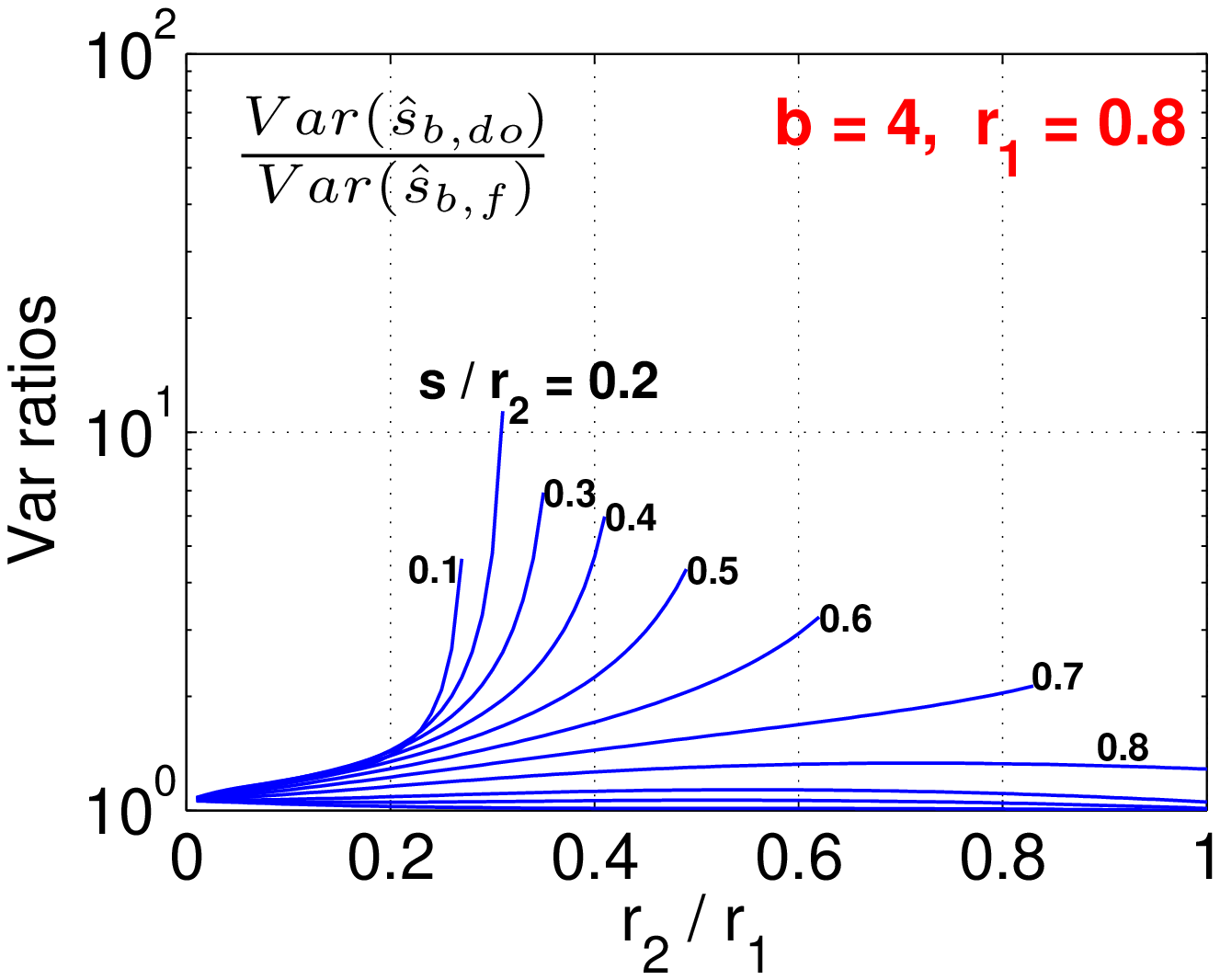}\hspace{-0.2in}
\includegraphics[width = 1.8in]{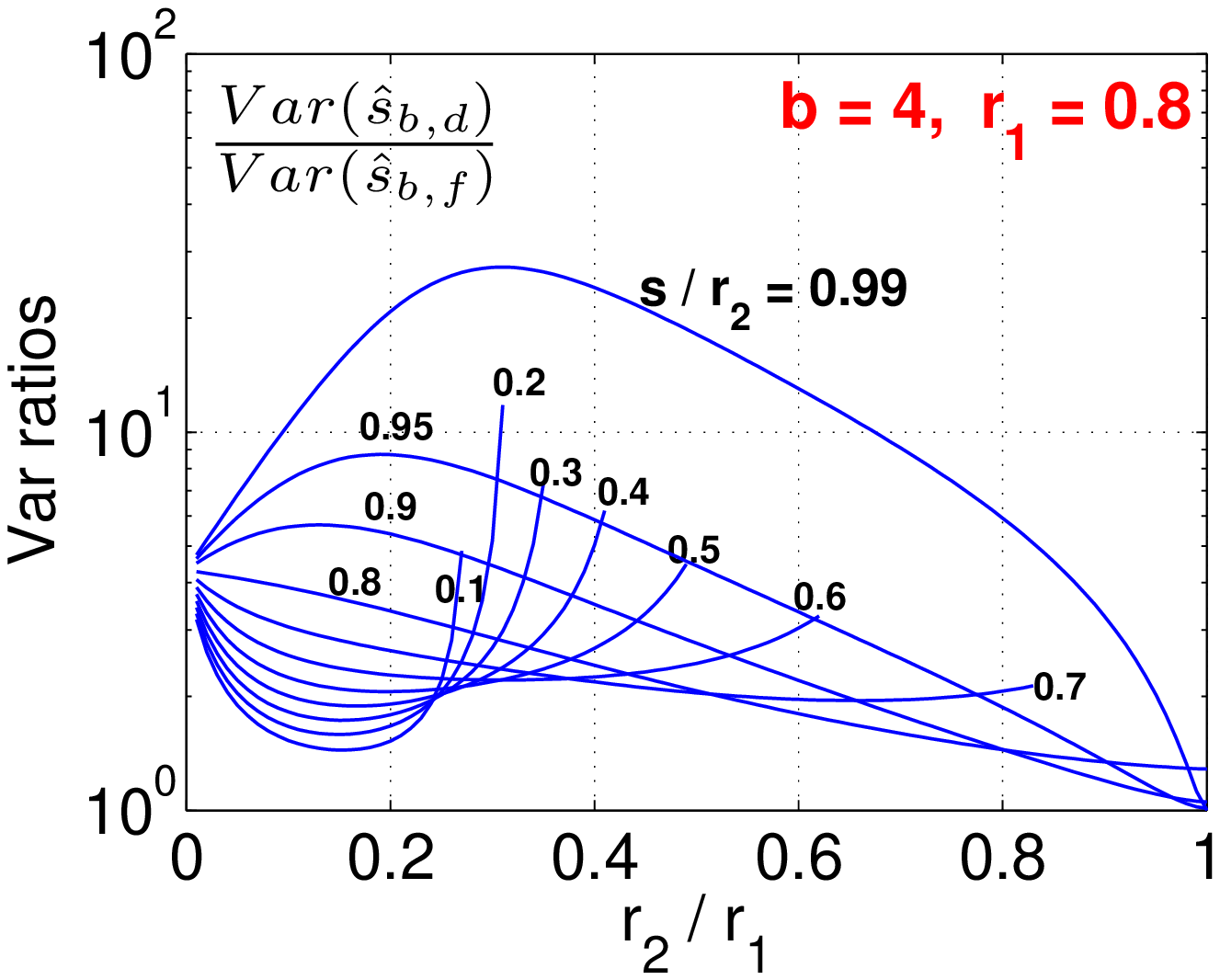}}
\mbox{
\includegraphics[width = 1.8in]{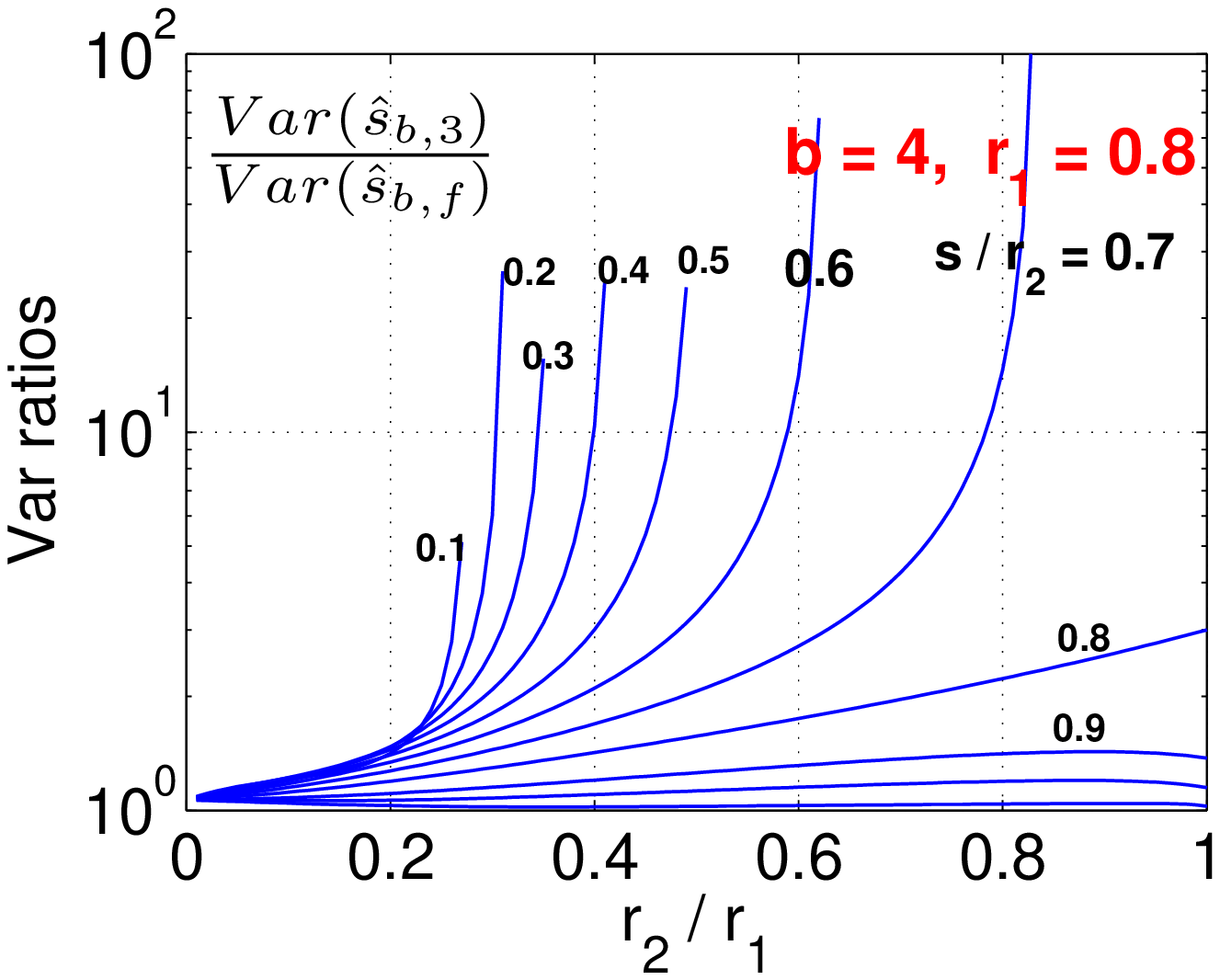}\hspace{-0.2in}
\includegraphics[width = 1.8in]{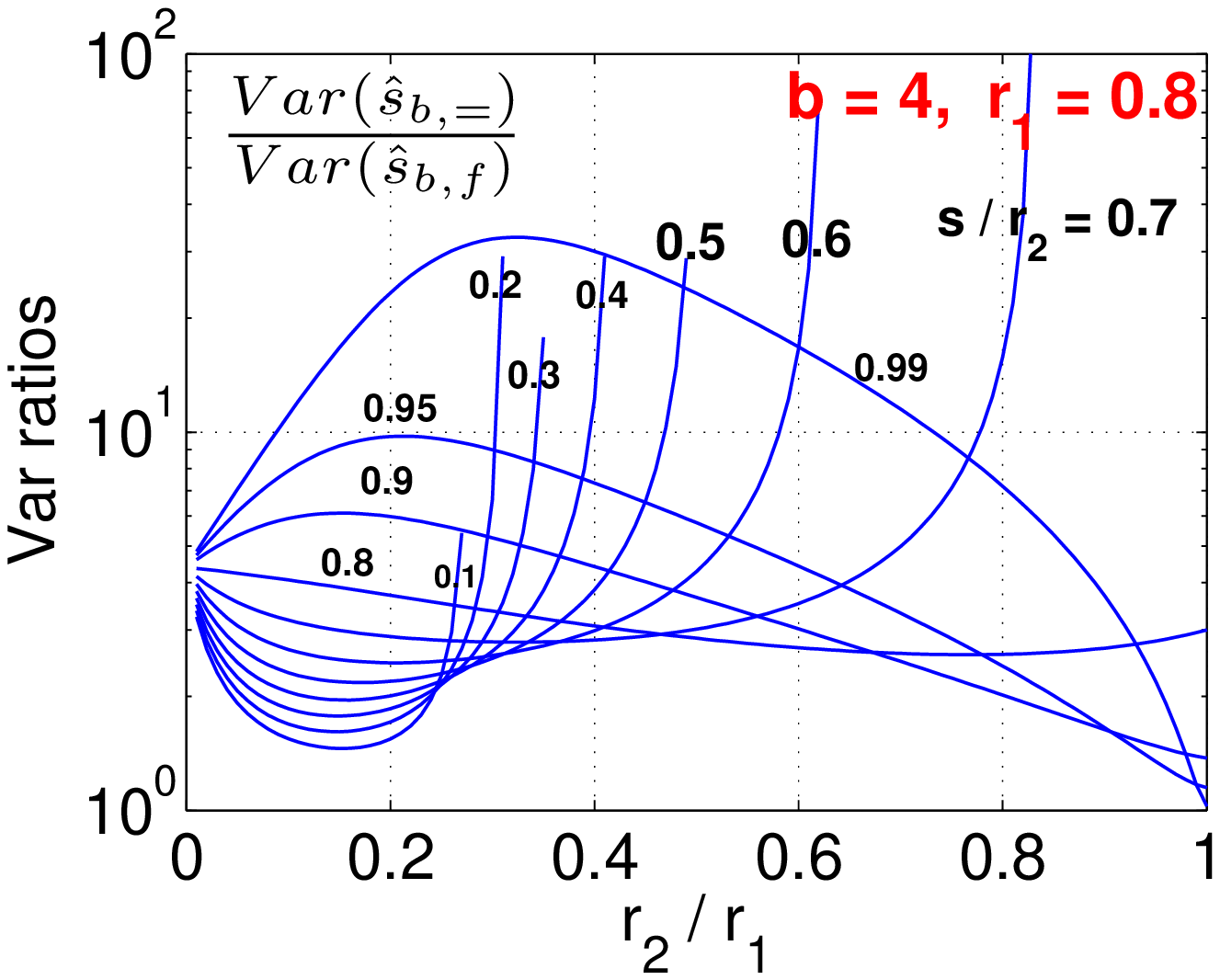}}
\end{center}\vspace{-0.3in}
\caption{Variance ratios for $b=4$ and $r_1=0.8$.}\label{fig_var_b4_r08}
\end{figure}

\begin{figure}[h!]
\begin{center}
\mbox{
\includegraphics[width = 1.8in]{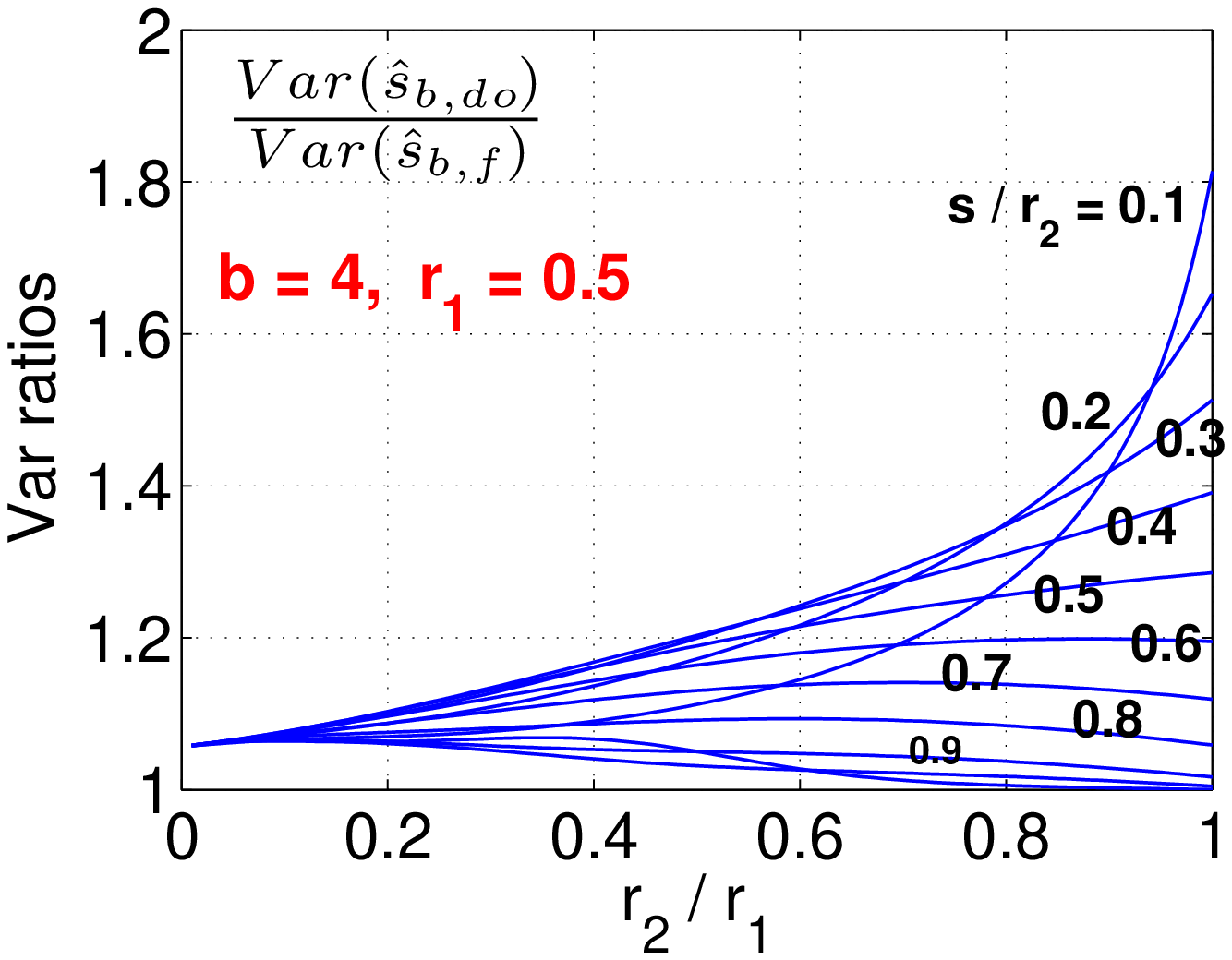}\hspace{-0.2in}
\includegraphics[width = 1.8in]{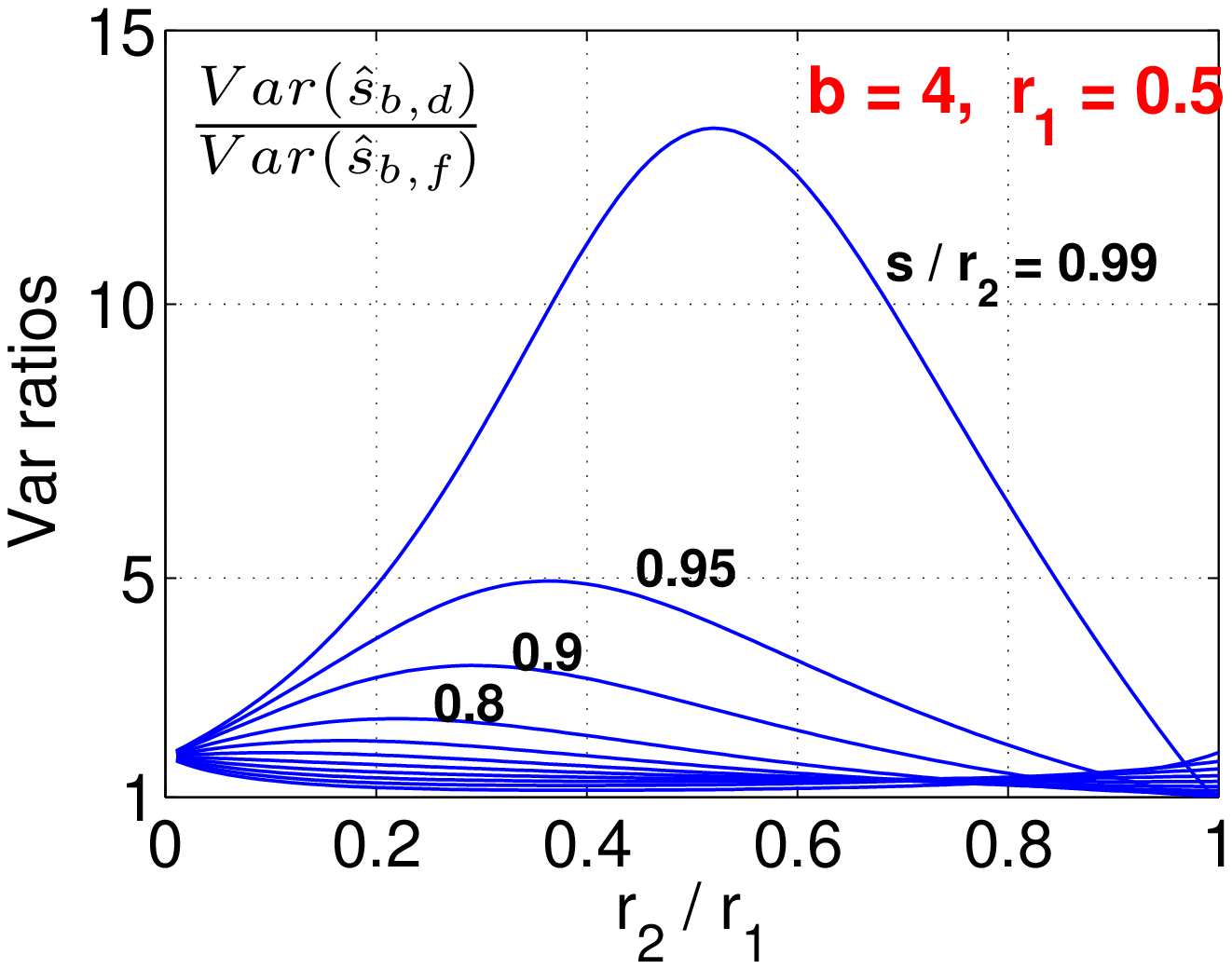}}
\mbox{
\includegraphics[width = 1.8in]{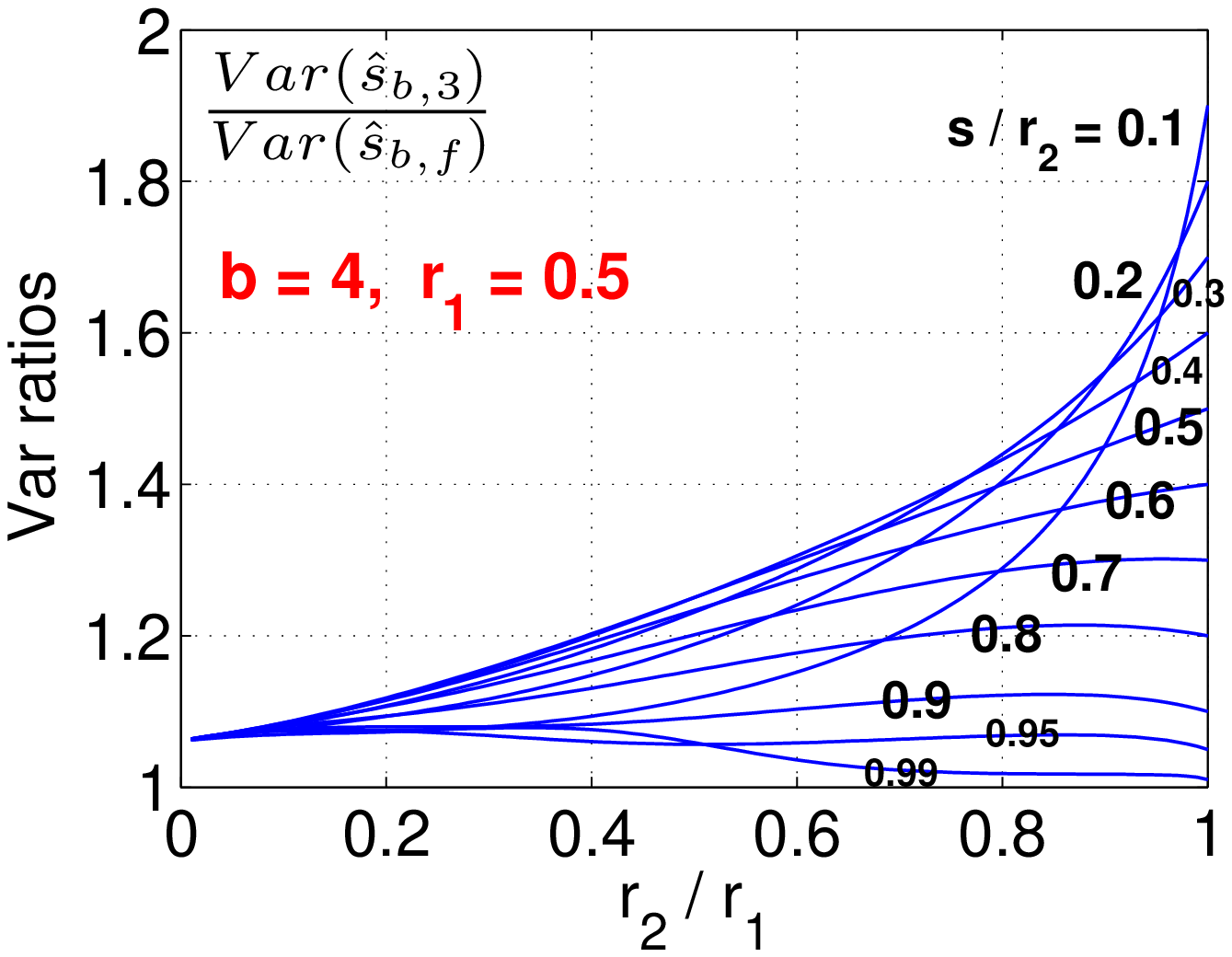}\hspace{-0.2in}
\includegraphics[width = 1.8in]{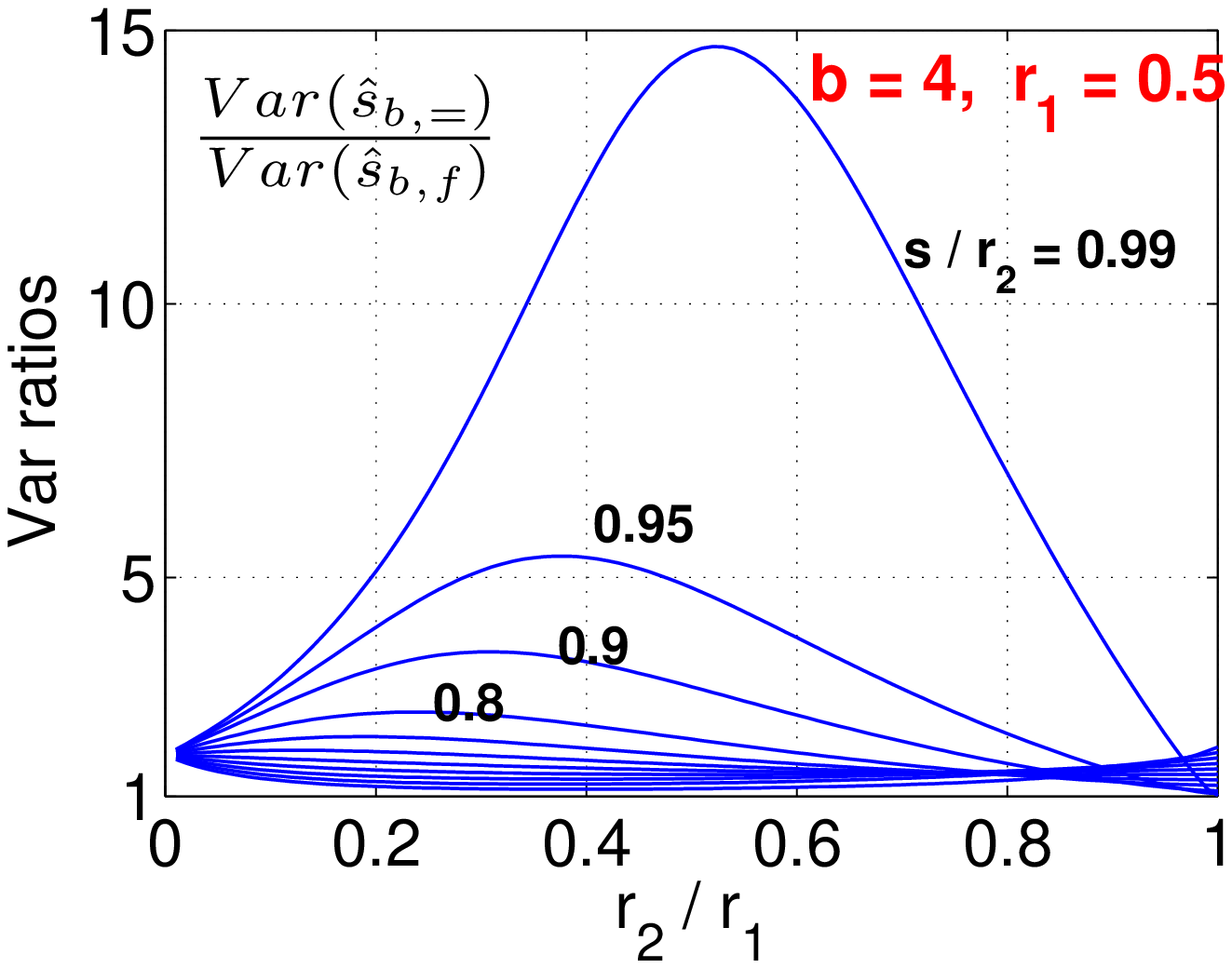}}
\end{center}\vspace{-0.3in}
\caption{Variance ratios for $b=4$ and $r_1=0.5$.}\label{fig_var_b4_r05}
\end{figure}

\begin{figure}[h!]
\begin{center}
\mbox{
\includegraphics[width = 1.8in]{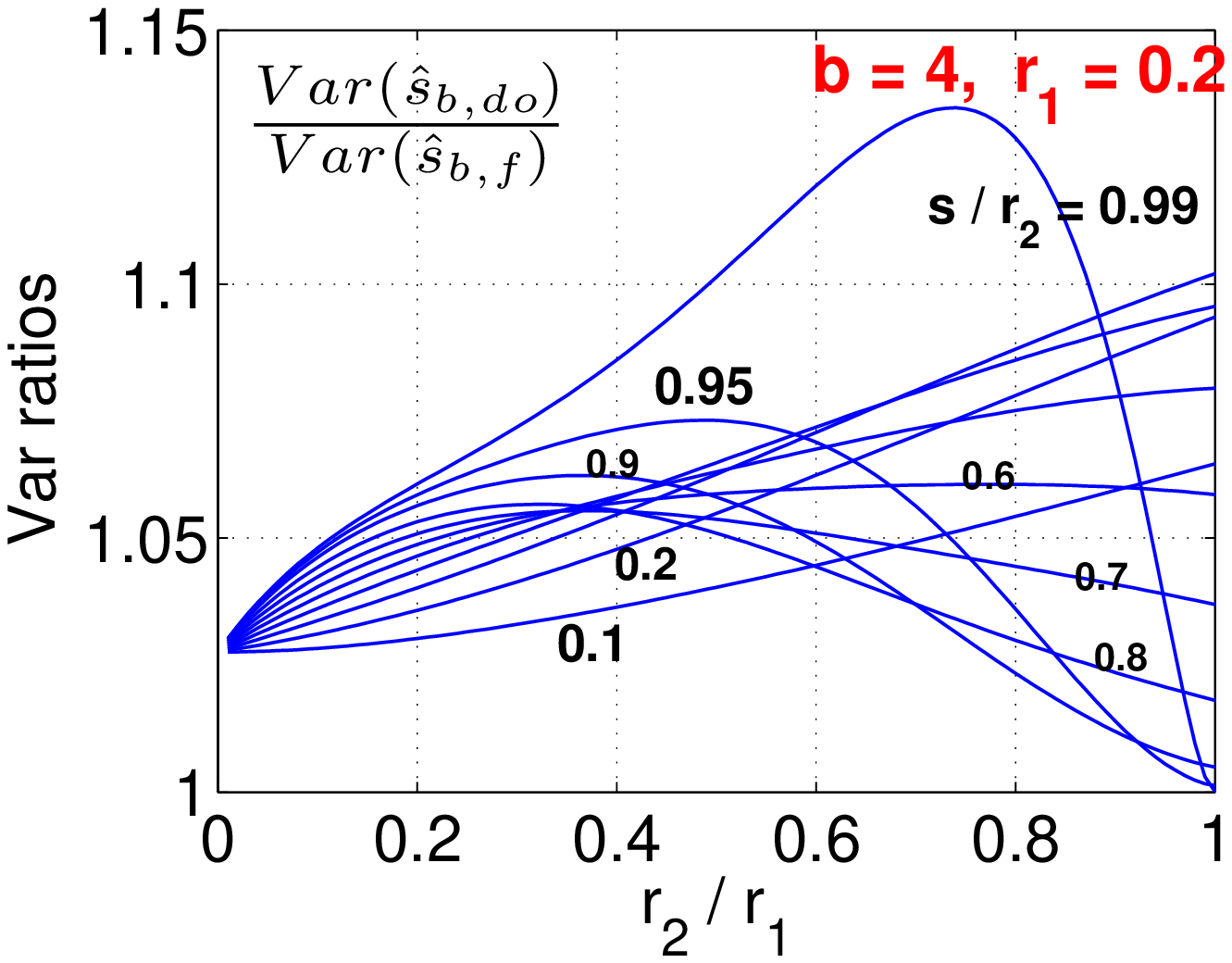}\hspace{-0.2in}
\includegraphics[width = 1.8in]{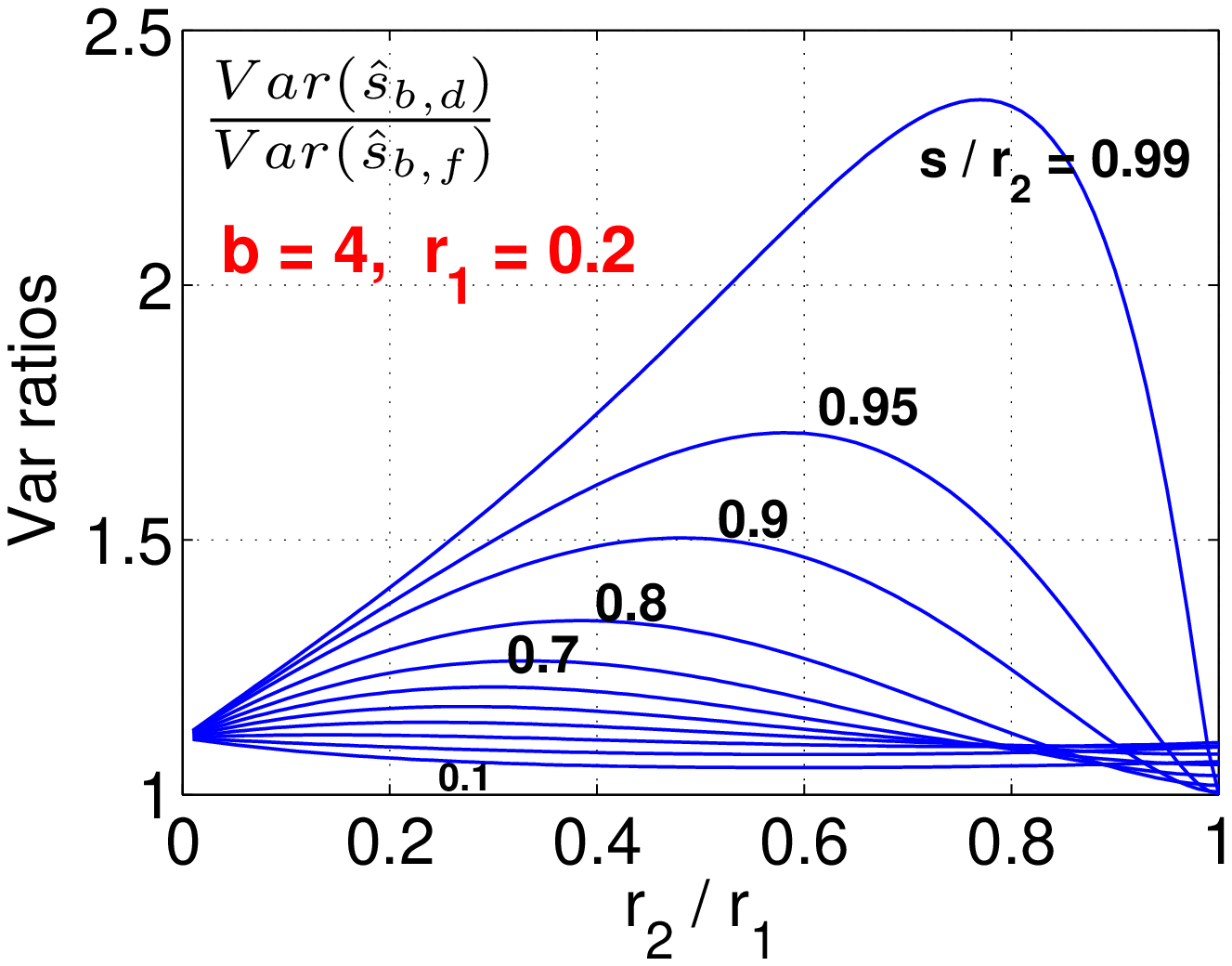}}
\mbox{
\includegraphics[width = 1.8in]{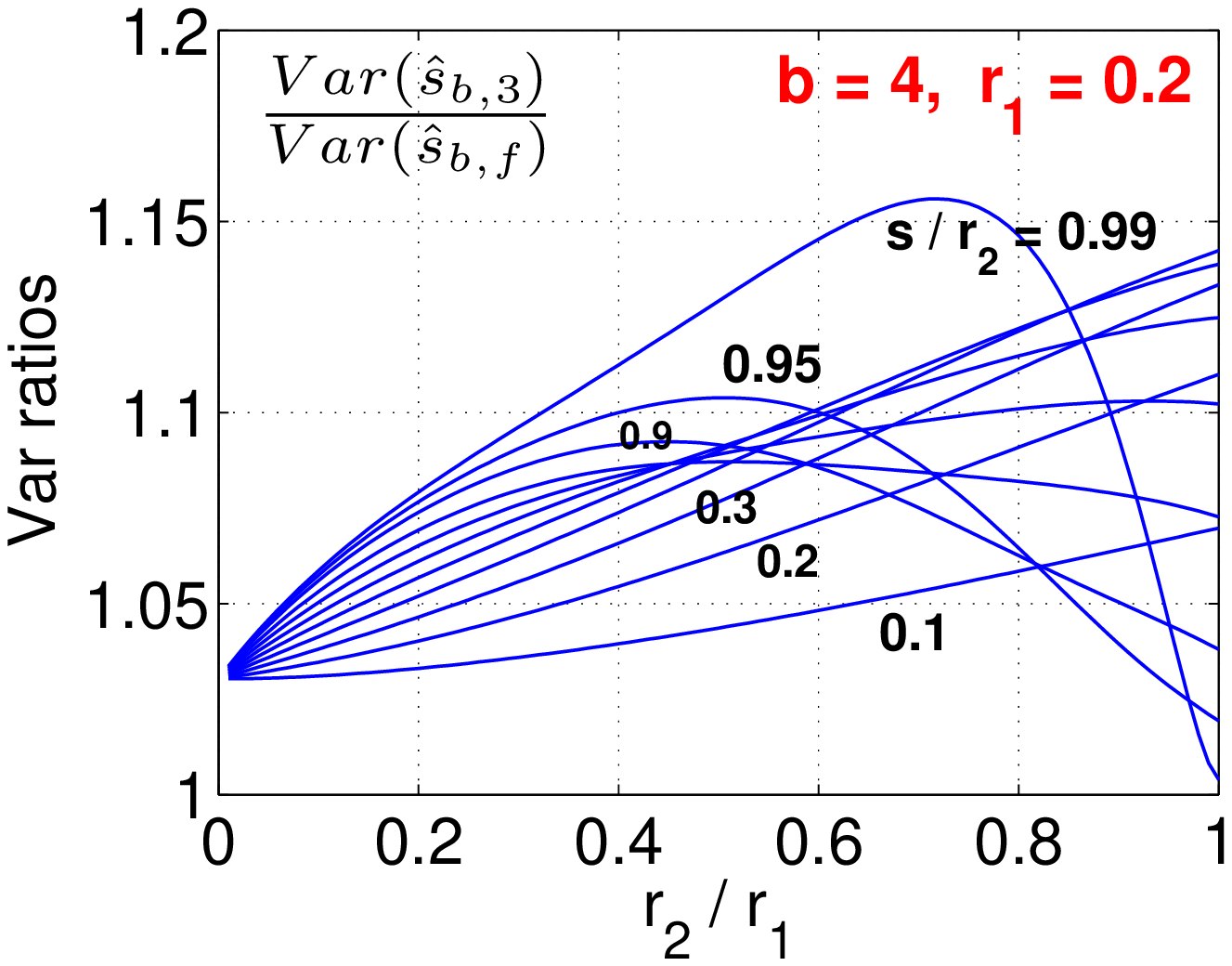}\hspace{-0.2in}
\includegraphics[width = 1.8in]{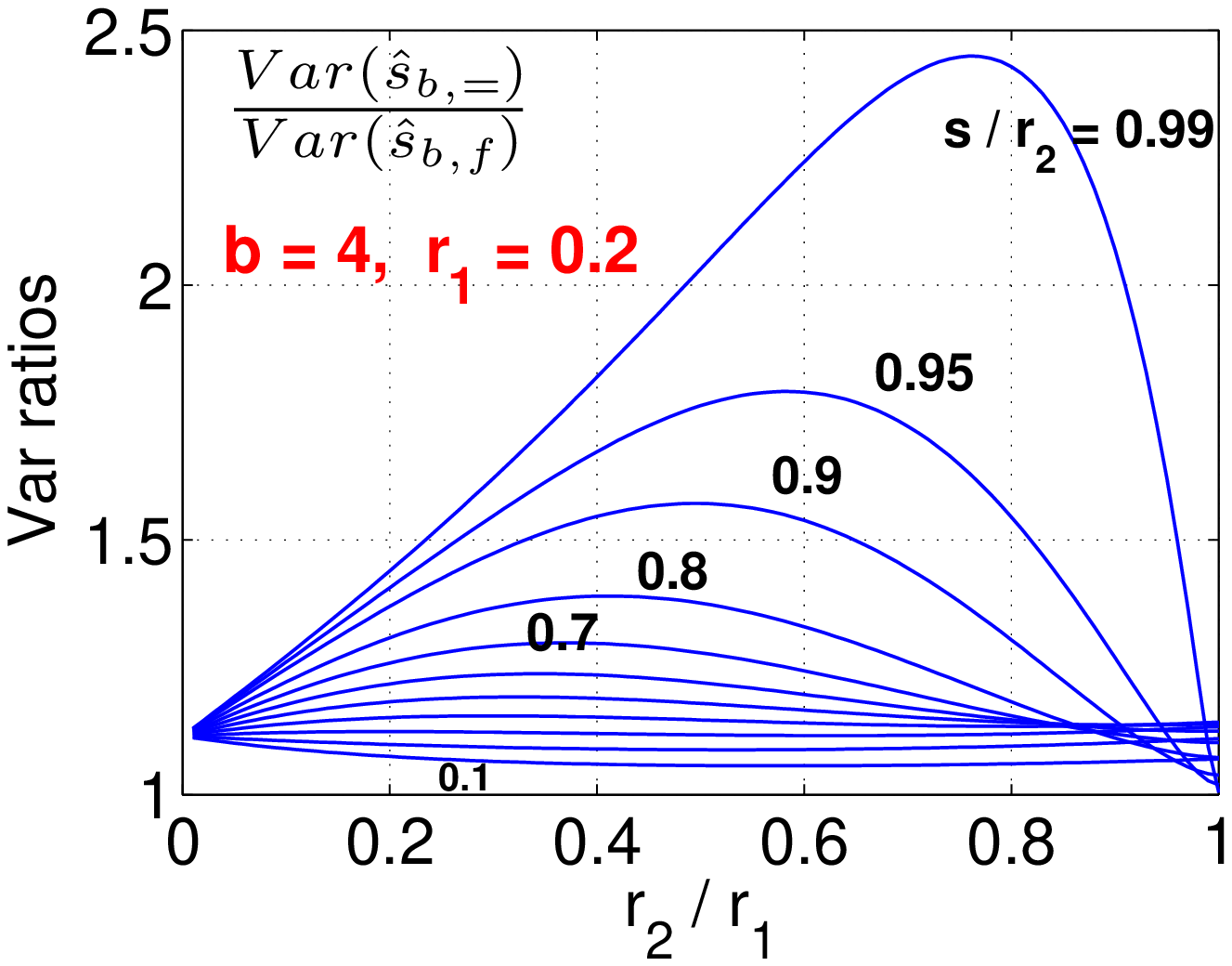}}
\end{center}\vspace{-0.3in}
\caption{Variance ratios for $b=4$ and $r_1=0.2$.}\label{fig_var_b4_r02}
\end{figure}

\begin{figure}[h!]
\begin{center}
\mbox{
\includegraphics[width = 1.8in]{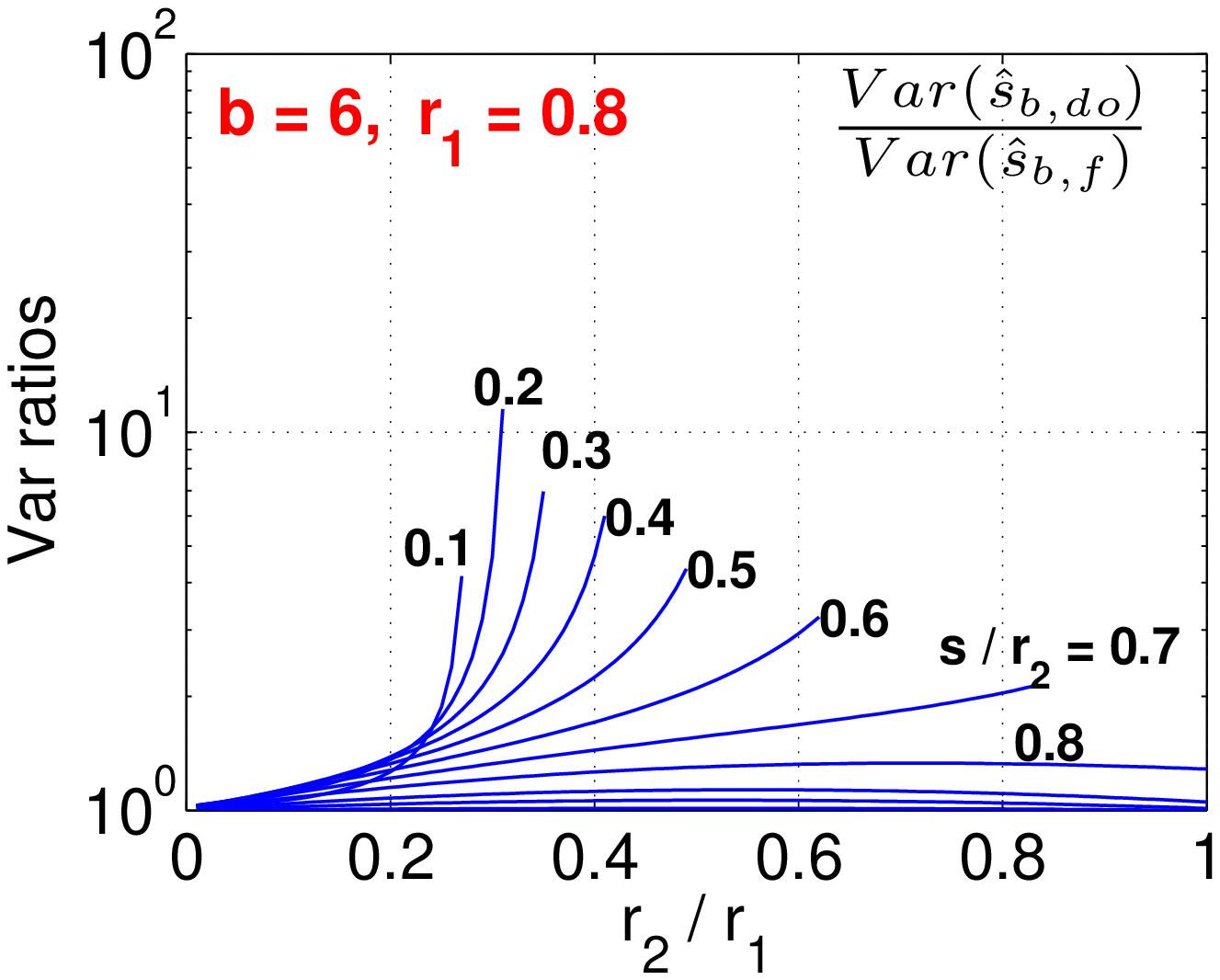}\hspace{-0.2in}
\includegraphics[width = 1.8in]{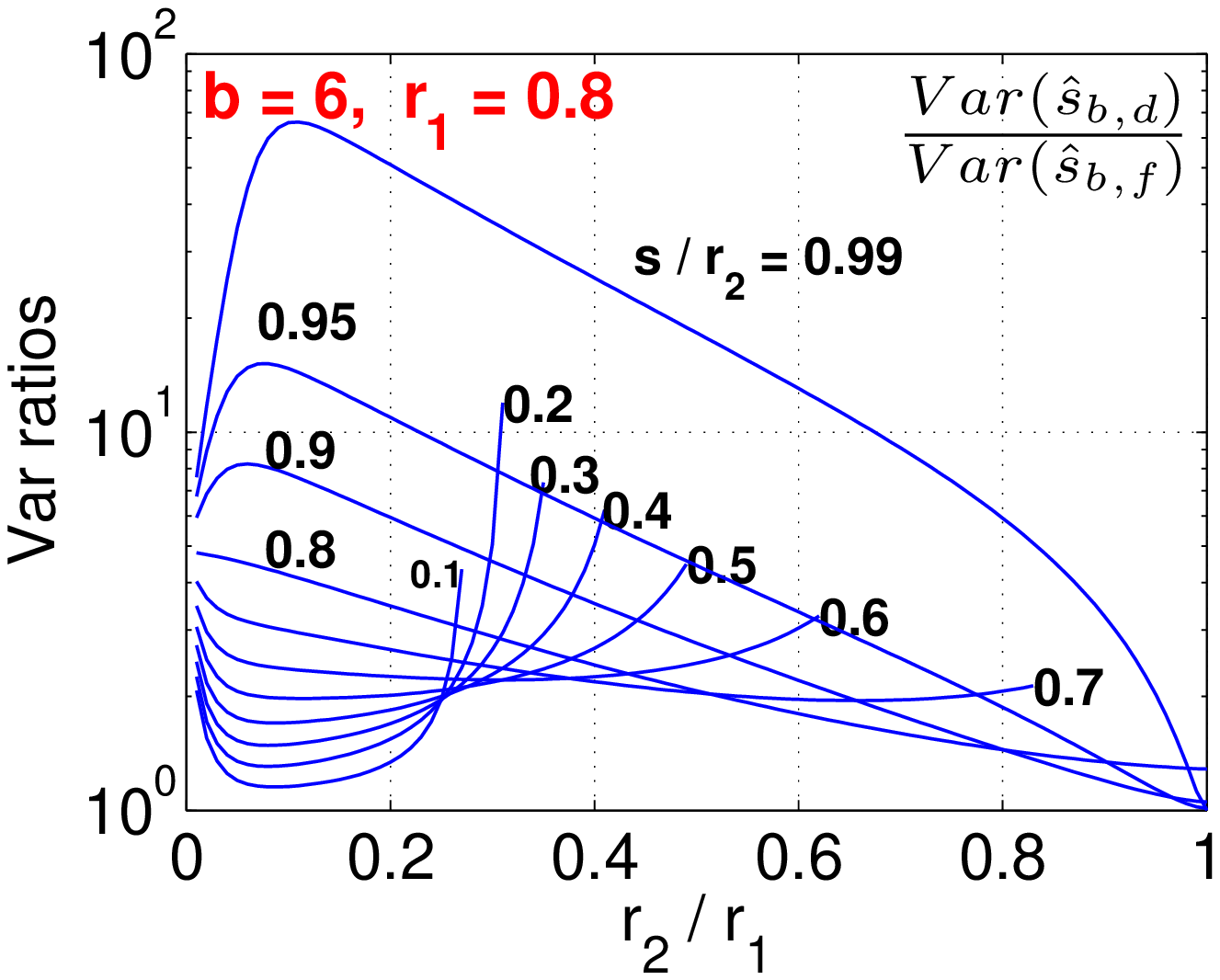}}
\mbox{
\includegraphics[width = 1.8in]{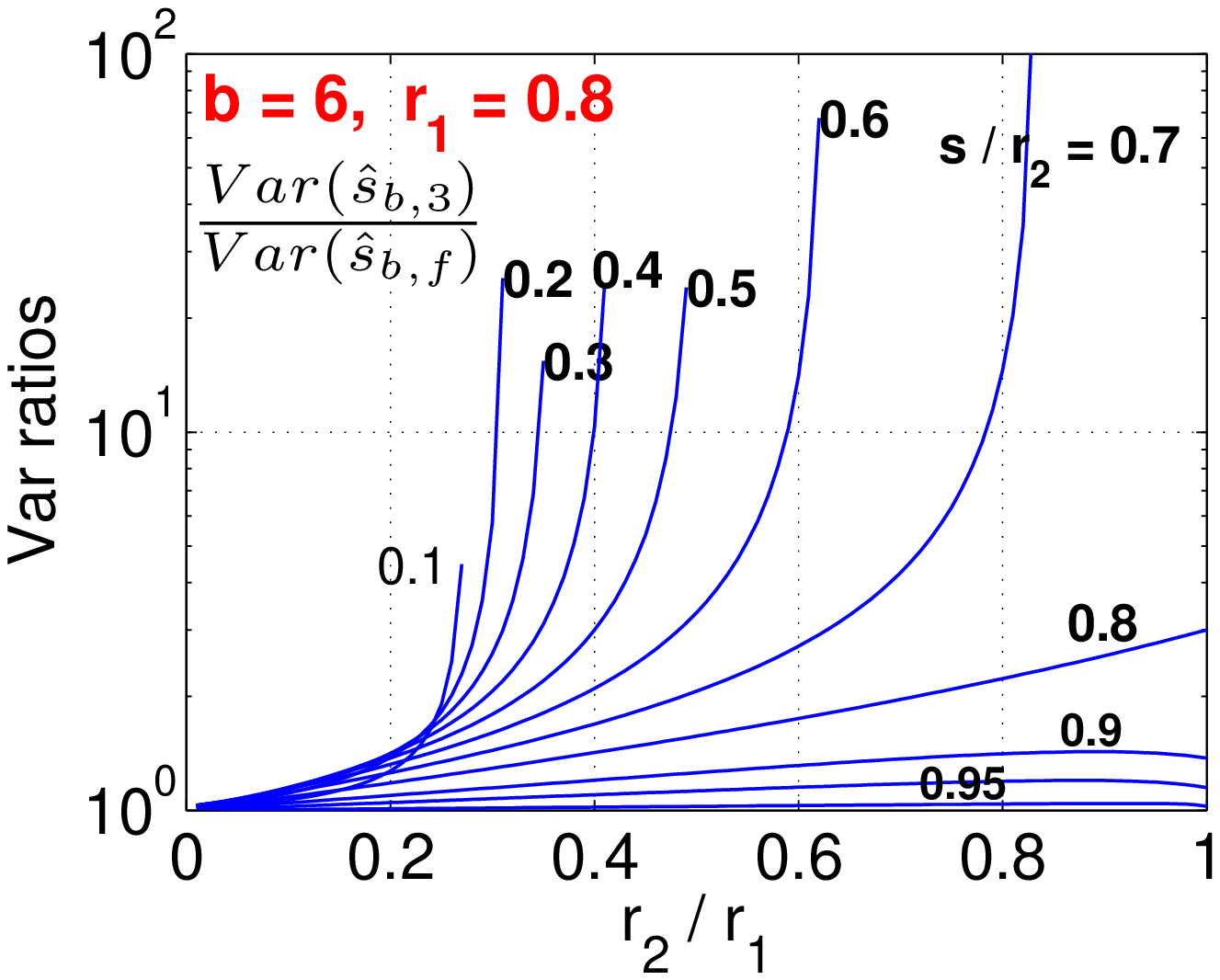}\hspace{-0.2in}
\includegraphics[width = 1.8in]{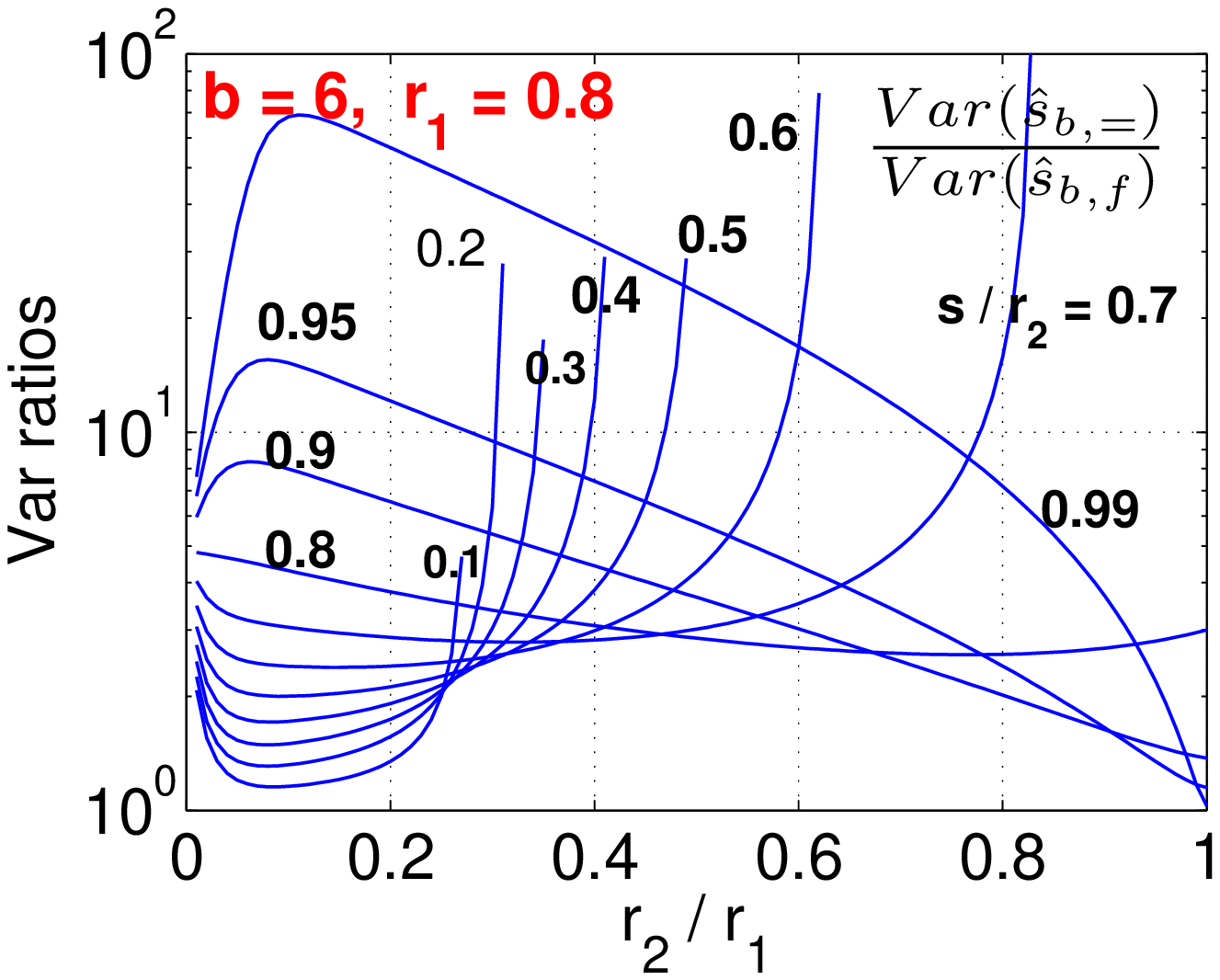}}
\end{center}
\vspace{-0.3in}
\caption{Variance ratios for $b=6$ and $r_1=0.8$.}\label{fig_var_b6_r08}
\end{figure}

\begin{figure}[h!]
\begin{center}
\mbox{
\includegraphics[width = 1.8in]{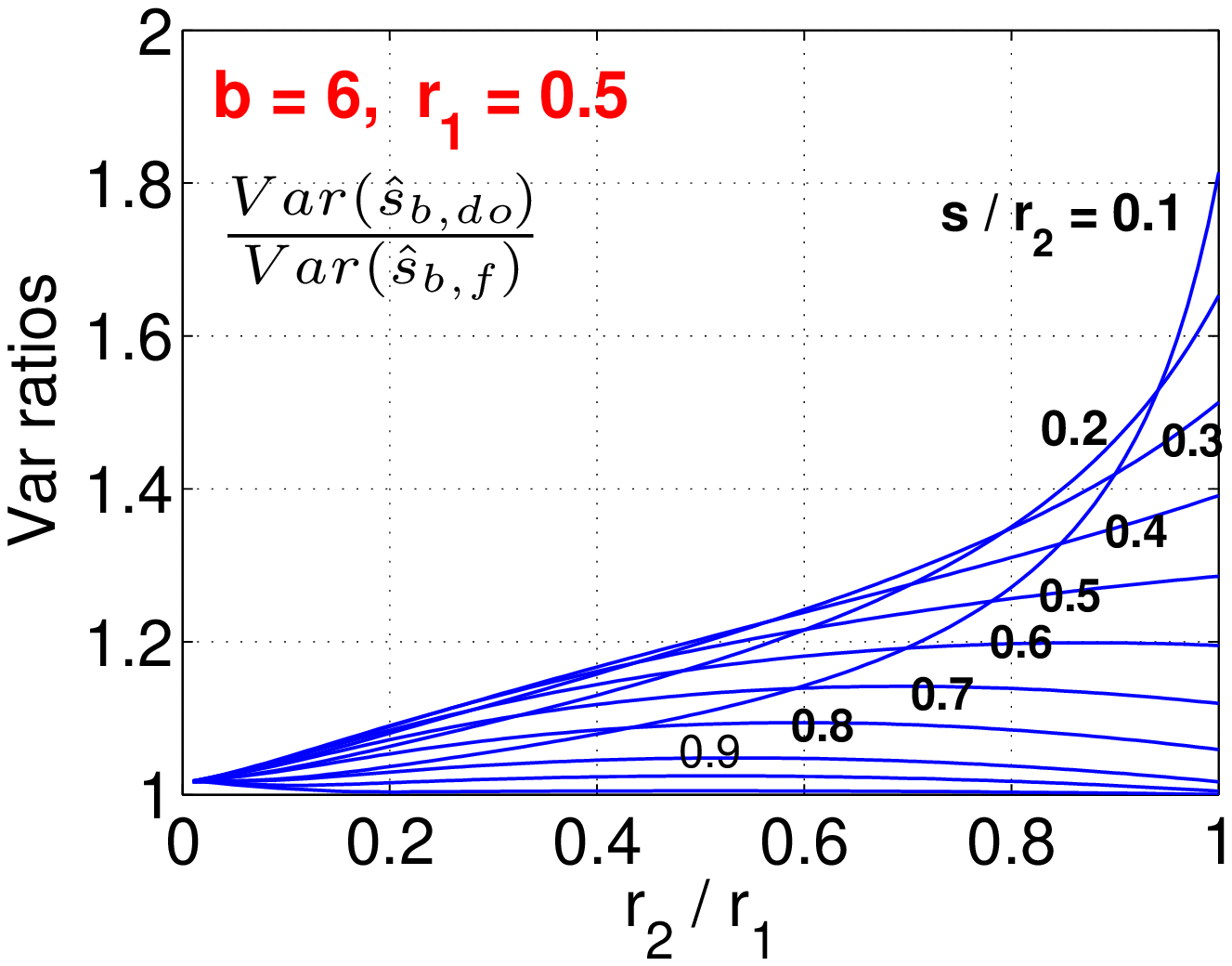}\hspace{-0.2in}
\includegraphics[width = 1.8in]{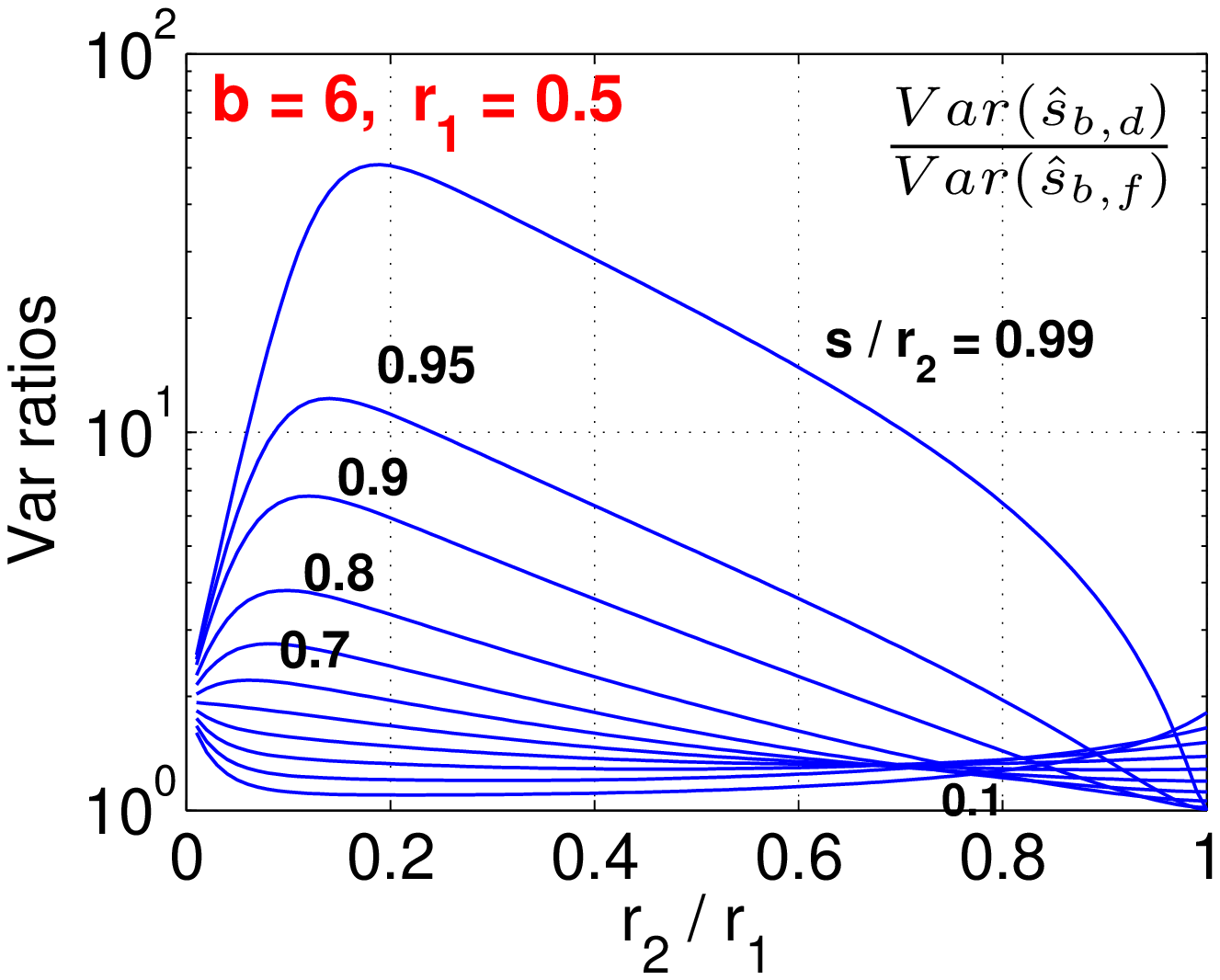}}
\mbox{
\includegraphics[width = 1.8in]{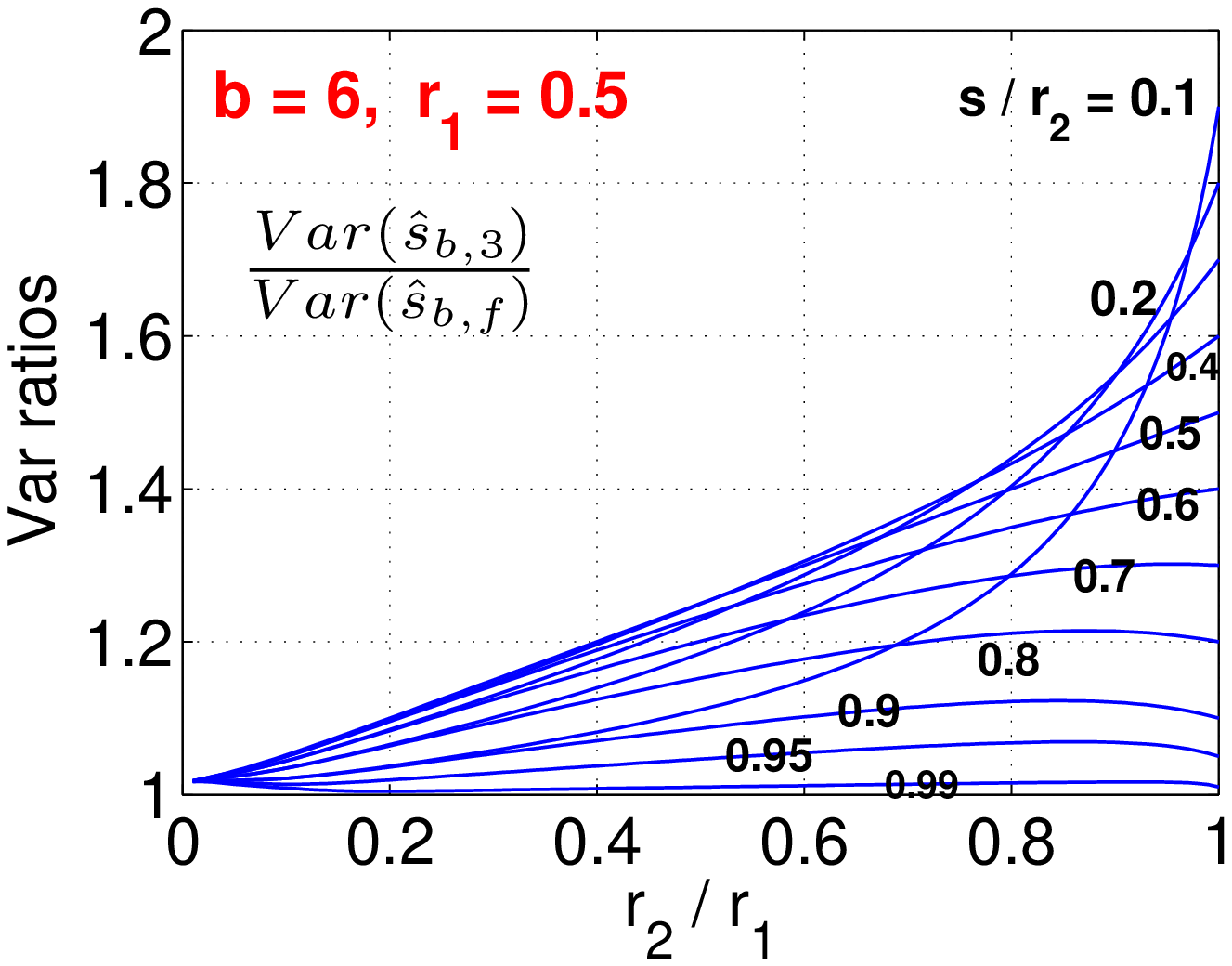}\hspace{-0.2in}
\includegraphics[width = 1.8in]{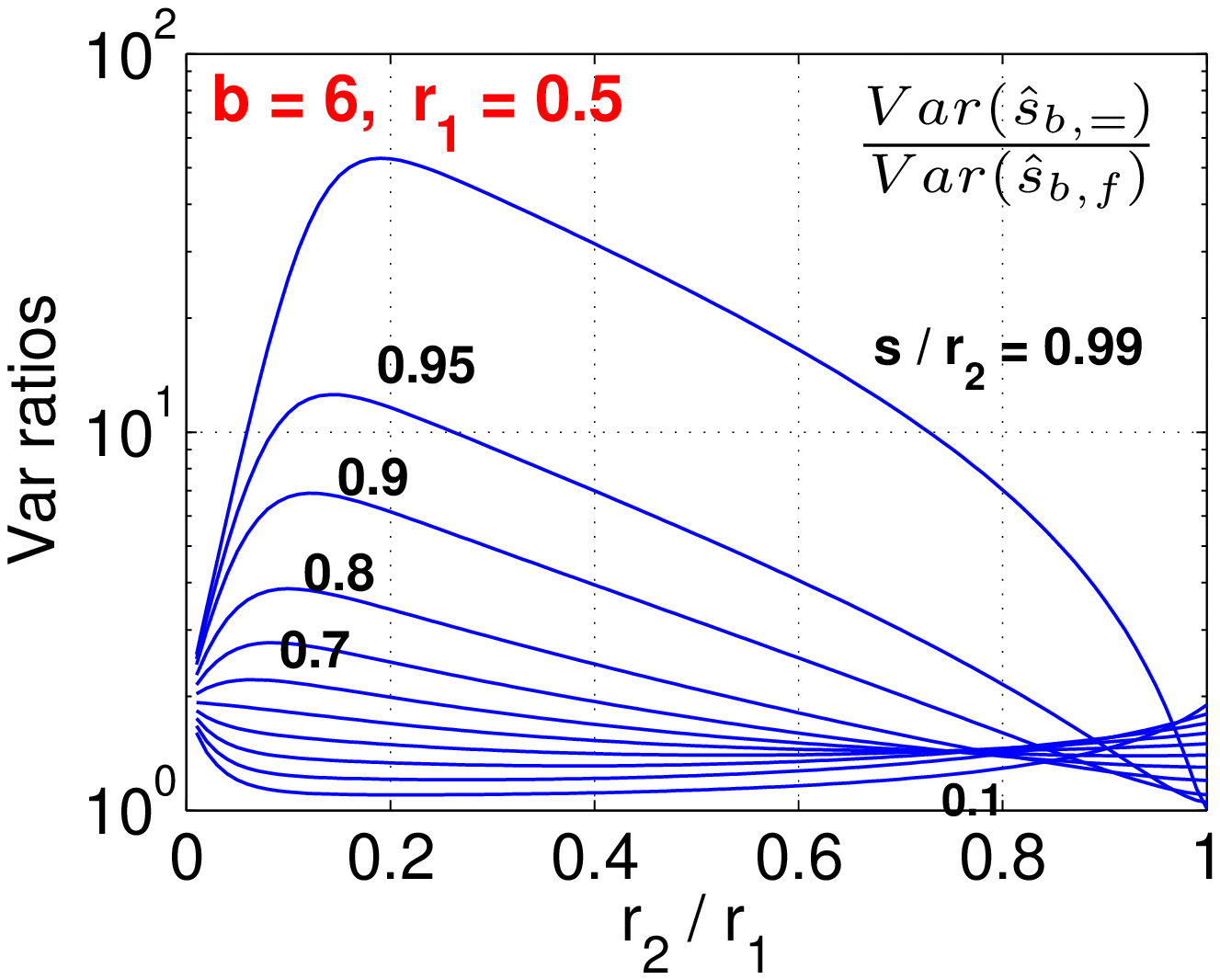}}
\end{center}
\vspace{-0.3in}
\caption{Variance ratios for $b=6$ and $r_1=0.5$.}\label{fig_var_b6_r05}
\end{figure}

\begin{figure}[h!]
\begin{center}
\mbox{
\includegraphics[width = 1.8in]{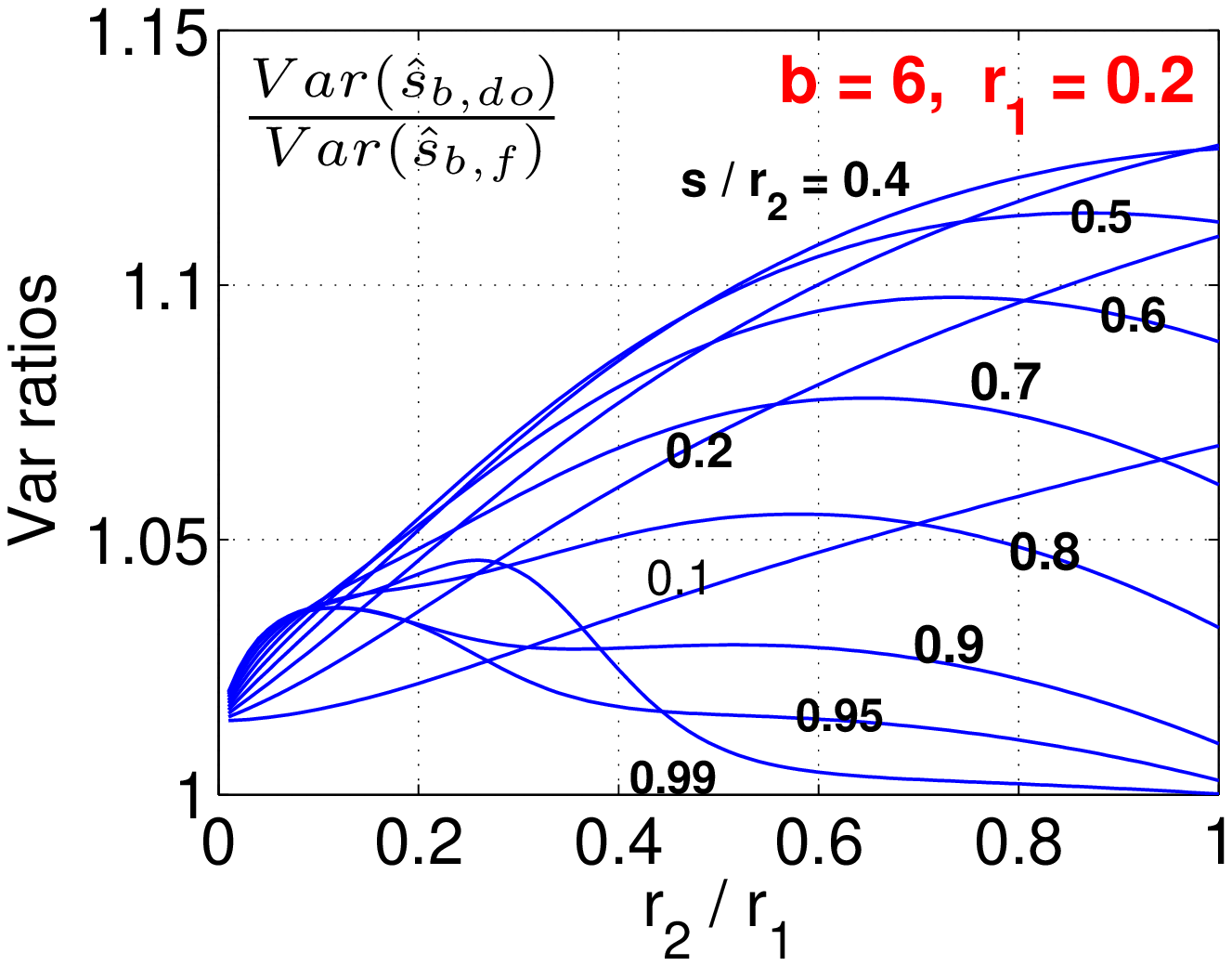}\hspace{-0.2in}
\includegraphics[width = 1.8in]{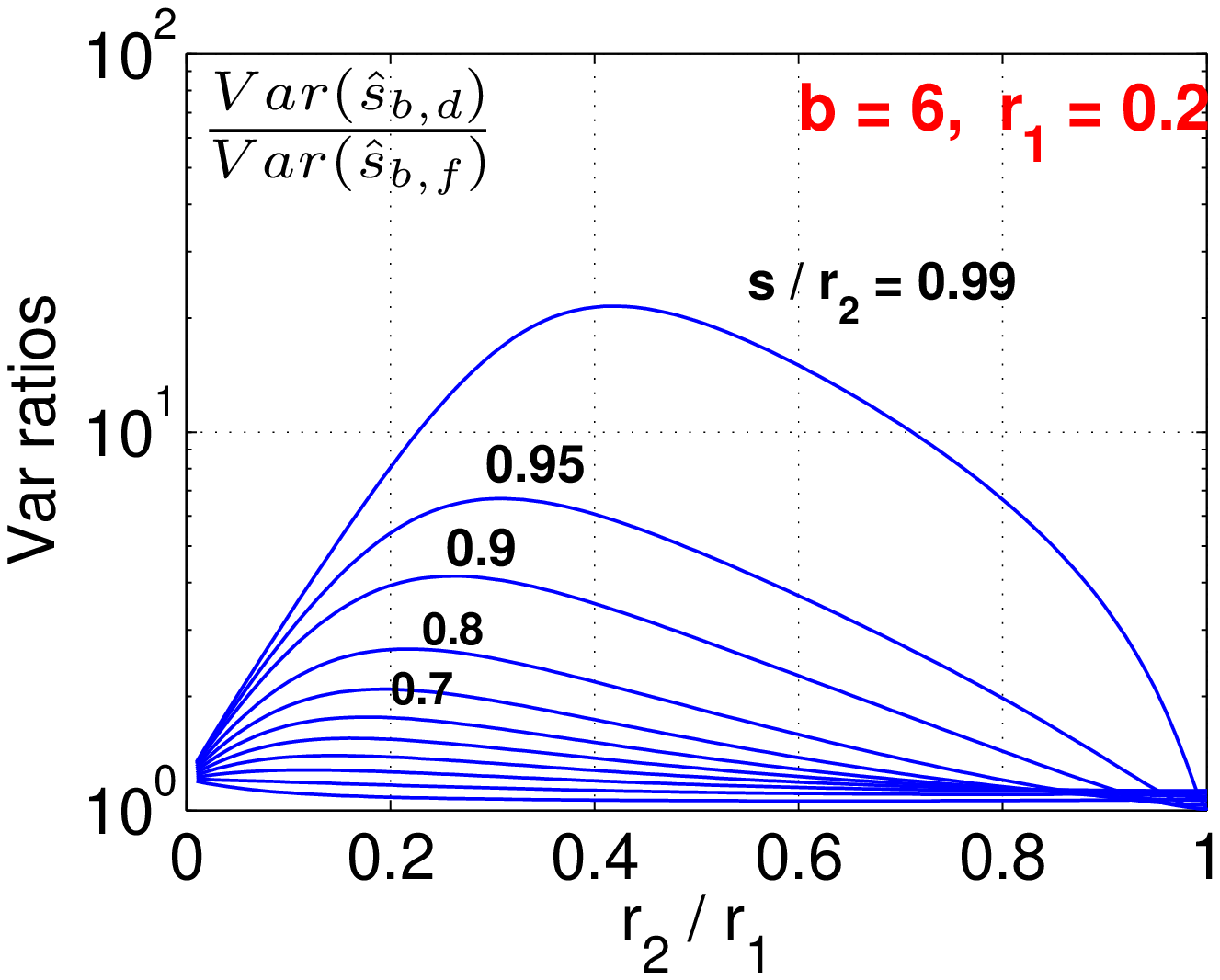}}
\mbox{
\includegraphics[width = 1.8in]{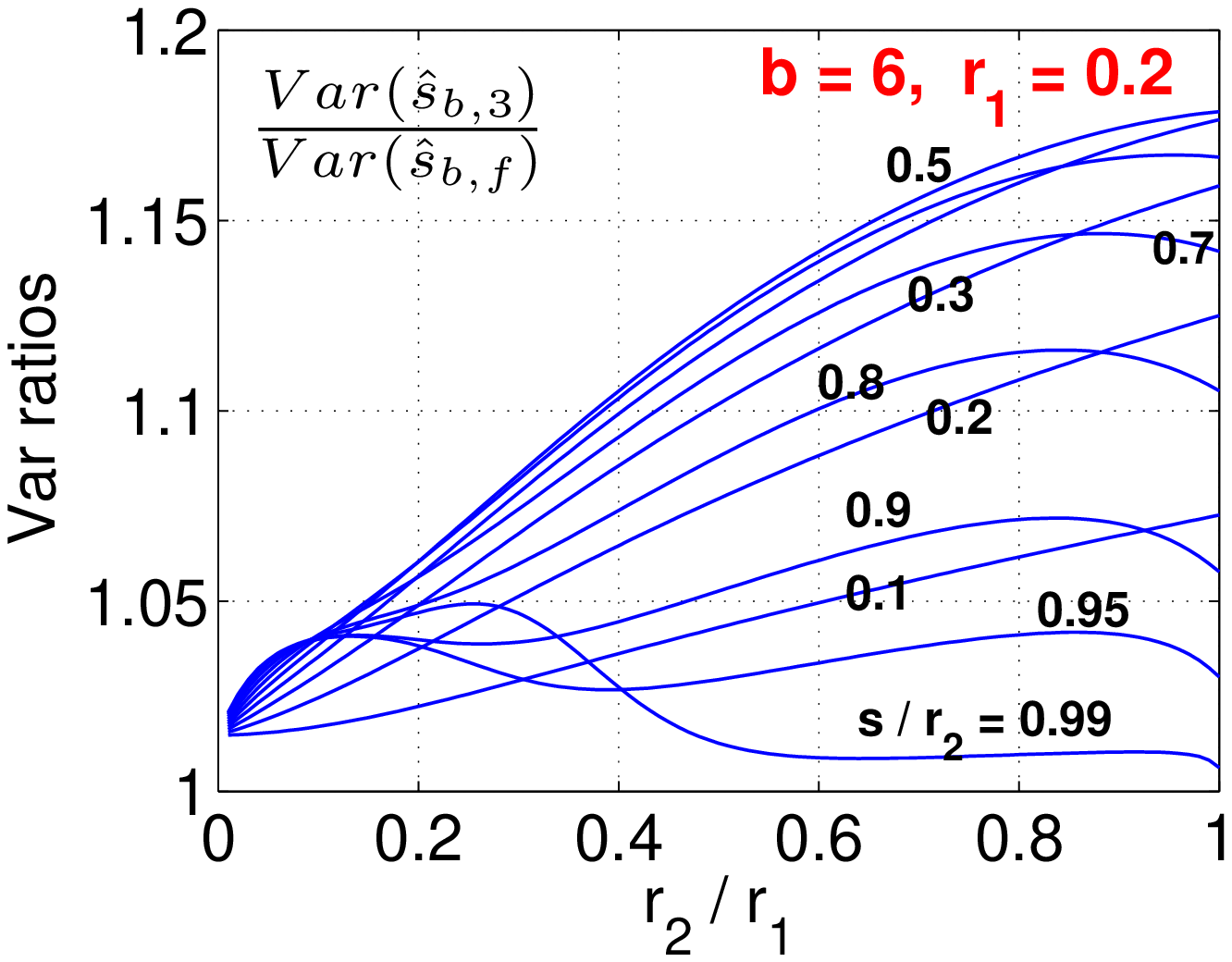}\hspace{-0.2in}
\includegraphics[width = 1.8in]{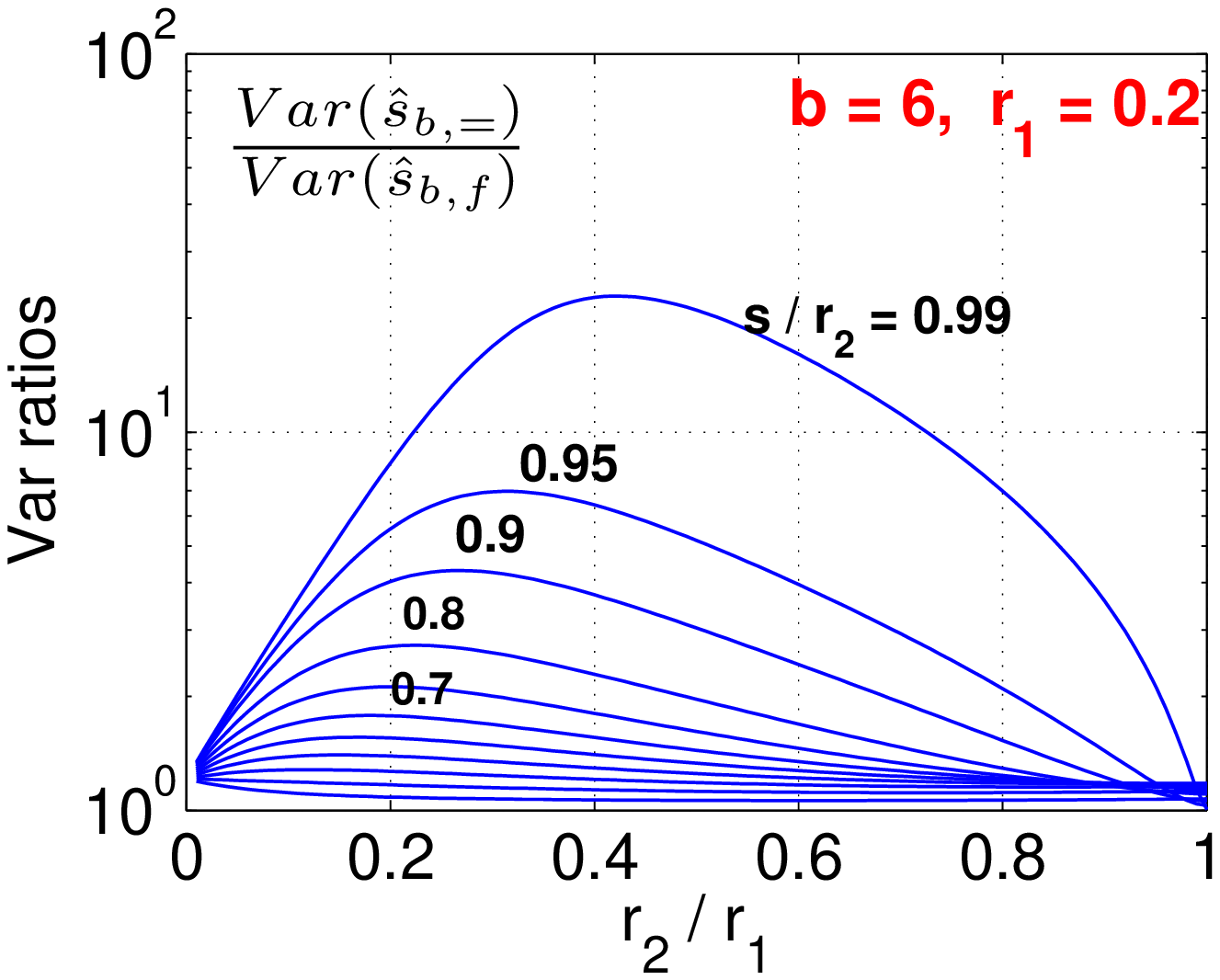}}
\end{center}
\vspace{-0.3in}
\caption{Variance ratios for $b=6$ and $r_1=0.2$.}\label{fig_var_b6_r02}
\end{figure}

We compare the asymptotic variances of the other four estimators, $\hat{s}_{b,do}$, $\hat{s}_{b,d}$, $\hat{s}_{b,3}$, and $\hat{s}_{b,=}$,  with the variance of the full MLE $\hat{s}_{b,f}$ in Figures~\ref{fig_var_b8_r08} to~\ref{fig_var_b6_r02}. We consider $b=8, 4, 6$, $r_1 = 0.8, 0.5, 0.2$ and the full ranges of $\frac{r_2}{r_1}$ and $\frac{s}{r_2}$ (which is the containment). Note that the improvement of this paper compared to the previous standard practice (i.e., $\hat{s}_{b,=}$) is only reflected in the bottom-right panel of each figure. We present other estimators in the hope of finding one which is much simpler than the full MLE $\hat{s}_{b,f}$ and still retains much of the improvement. Our observations are:
\begin{itemize}
\item The full MLE $\hat{s}_{b,f}$, which uses a matrix of $2^b\times 2^b$ probabilities, can achieve substantial improvements (for example, 5- to 100-fold) compared to the standard practice $\hat{s}_{b,=}$, especially for cases of low resemblance and high containment.
\item Two other estimators, $\hat{s}_{b,do}$ and $\hat{s}_{b,3}$ usually perform very well compared to the full MLE. $\hat{s}_{b,do}$  uses $2^b+2$ cells and $\hat{s}_{b,3}$ uses merely 3 cells: the sum of the diagonals and the two sums of the off-diagonals. Therefore, we consider $\hat{s}_{b,3}$ is likely to be particularly useful in practice.
\end{itemize}

\clearpage

Our analysis has demonstrated that the much simpler estimator $\hat{s}_{b,3}$, which only uses 3 cells, is often remarkably accurate. We expect it will be used in practice. $\hat{s}_{b,3}$ involves three summary probabilities: $P_{b,=}$, $P_{b,<}$, and $P_{b,>}$. For efficient estimation, we will need to use more compact presentations instead of the double summation forms. Since we already know $P_{b,=}$ as derived in~\cite{Proc:Li_Konig_WWW10}, we only need to derive $P_{b,<}$ and then $P_{b,>}$ follows by symmetry.  After some  algebra, we obtain
{\small\begin{align}\notag
&P_{b,<} = \mathbf{Pr}\left(u_{1,b} < u_{2,b}\right)
=\sum_{t<d}^{2^b-1}\mathbf{Pr}\left(u_{1,b}=t, \ u_{2,b}=d\right)\\
=&\frac{1}{1-\left[1-r_2\right]^{2^b}}
\frac{(r_1-s)}{1-\left[1-(r_1+r_2-s)\right]^{2^b}}A\\\notag
&+\frac{\left[1-r_1\right]^{2^b-1}}
{1-\left[1-r_1\right]^{2^b}}\frac{(r_2-s)}{1-\left[1-(r_1+r_2-s)\right]^{2^b}}B
\end{align}}
where
{\tiny\begin{align}\notag
A=&\frac{1-\left[1-(r_1+r_2-s)\right]^{2^b-1}}{r_1+r_2-s}
-\frac{\left[1-r_2\right]^{2^b}-\left[1-(r_1+r_2-s)\right]^{2^b-1}\left[1-r_2\right]}{r_1-s}\\\notag
B=&\frac{\left[1-r_1\right]^{2^b}-\left[1-(r_1+r_2-s)\right]^{2^b}}{r_2-s}\left[1-r_1\right]^{1-2^b}\\\notag
&-\frac{\left[1-(r_1+r_2-s)\right] - \left[1-(r_1+r_2-s)\right]^{2^b}}{r_1+r_2-s}
\end{align}}

\vspace{-0.2in}
\section{Conclusion}\vspace{-0.1in}

Computing set or vector similarity is a routine task in numerous applications in machine learning, information retrieval, and databases. In Web scale applications, the method of minwise hashing is a standard technique for efficiently estimating similarities, by hashing each  set (or equivalently binary vector) in a dictionary of size $|\Omega| = 2^{64}$ to about $k$ hashed values ($k=200$ to 500 is common). The standard industry practice is to store each hashed value using 64 bits. The recently developed  b-bit minwise hashing stores only the lowest $b$ bits with small $b$. b-Bit minwise hashing is  successful in applications which  care about pairs of high similarities (e.g., duplicate  detection).

However, many applications  involve computing all pairwise similarities (and most of the pairs are not similar). Furthermore, some applications really care about containment (e.g., the fraction that one object is contained by another) instead of resemblance. Interestingly, the current standard methods for minwise hashing and b-bit minwise hashing
 perform poorly for cases of low resemblance and high containment.

Our contributions in this paper include the statistically optimal estimator for standard minwise hashing and several new estimators for b-bit minwise hashing. For important scenarios (e.g., low resemblance and high containment), improvements of about an order of magnitude can be obtained. The full MLE solution for b-bit minwise hashing involves a contingency table of  $2^b\times 2^b$ cells, which can be prohibitive if $b$ is large. Our analysis suggests that if we only use 3 cells, i.e., the sum of the diagonals and the two sums of the off-diagonals, we can still achieve significant improvements compared to the current practice.

\appendix

\vspace{-0.1in}
\section{Proof of Lemma~\ref{lem_Ptd}}\label{proof_lem_Ptd}\vspace{-0.1in}

Consider two sets $S_1, S_2\in\Omega = \{0, 1, 2, ..., D-1\}$. Apply a random permutation $\pi: \Omega\rightarrow \Omega$ on a $S_1$, $S_2$, and store the two minimums: $z_1 = \min(\pi(S_1))$, $z_2 = \min(\pi(S_2))$.  Assuming $D\rightarrow \infty$, \cite{Proc:Li_Konig_WWW10} provided two basic probability formulas:

{\small\begin{align}\notag
&\mathbf{Pr}\left(z_1 = i,\ z_2 = j, \ i<j \right) \\
= &r_2(r_1-s)\left[1-r_2\right]^{j-i-1}\left[1-(r_1+r_2-s)\right]^i\\\notag
&\mathbf{Pr}\left(z_1 = i,\ z_2 = j, \ i>j \right) \\
= &r_1(r_2-s)\left[1-r_1\right]^{i-j-1}\left[1-(r_1+r_2-s)\right]^j
\end{align}}

We will also need to derive $\mathbf{Pr}\left(z_1 = i,\ z_2 = i\right)$. The exact expression is  given by
{\small\begin{align}\notag
\mathbf{Pr}\left(z_1 = i,\ z_2 = i\right)=&\frac{\binom{D-i-1}{a-1}\binom{D-i-a}{f_1-a}\binom{D-i-f_1}{f_2-a}}
{\binom{D}{a}\binom{D-a}{f_1-a}\binom{D-f_1}{f_2-a}}\\\notag
=&\frac{a(D-i-1)!(D-f_1-f_2+a)!}{D!(D-f_1-f_2+a-i)!}\\\notag
=&\frac{a}{D}\frac{\prod_{t=0}^{i-1} D-f_1-f_2+a-t}{\prod_{t=0}^{i-1} D-1-t }
\end{align}}
For convenience, we introduce the following notation:
\begin{align}\notag
r_1 = \frac{f_1}{D}, \hspace{0.2in} r_2 = \frac{f_2}{D}, \hspace{0.2in} s = \frac{a}{D}.
\end{align}
Also, we assume $D$ is large (which is virtually always satisfied in practice). We can obtain a reasonable approximation (analogous to the Possion approximation of binomial):
\begin{align}\notag
&\mathbf{Pr}\left(z_1 = i,\ z_2 = i\right) = s\left[1-(r_1+r_2-s)\right]^i
\end{align}
To verify this, as expected,
\begin{align}\notag
&\mathbf{Pr}\left(z_1 =z_2\right)=\sum_{i}\mathbf{Pr}\left(z_1 = i,\ z_2 = i\right)\\\notag
=& \sum_{i}s\left[1-(r_1+r_2-s)\right]^i = \frac{s}{r_1+r_2-s} = R
\end{align}

Now we have tools to compute $\mathbf{Pr}\left(u_{1,b} = t, u_{2,b} = d\right)$, where $t,d \in \{0, 1, 2, 3, ..., 2^b-1\}$.

\begin{align}\notag
&\mathbf{Pr}\left(u_{1,b} = t, u_{2,b} = d, \  t<d\right)\\\notag
=&\mathbf{Pr}(z_1 = t, t+2^b, t+2\times 2^b,t+3\times2^b,..., \\\notag
&\hspace{0.23in} z_2 = d, d+2^b, d+2\times 2^b,d+3\times2^b,...)
\end{align}

{\small\begin{align}\notag
&\mathbf{Pr}\left(z_1 = t,\ \ z_2 = d, d+2^b, d+2\times 2^b,d+3\times2^b,..., \right)\\\notag
=&\sum_{j=0}\mathbf{Pr}\left(z_1 = t,\ \ z_2 = d+j\times 2^b,..., \right)\\\notag
=&\sum_{j=0}r_2(r_1-s)\left[1-r_2\right]^{d+j\times 2^b-t-1}\left[1-(r_1+r_2-s)\right]^t\\\notag
=&r_2(r_1-s)\frac{\left[1-r_2\right]^{d-t-1}}{1-\left[1-r_2\right]^{2^b}}\left[1-(r_1+r_2-s)\right]^t
\end{align}

\begin{align}\notag
&\mathbf{Pr}\left(z_1 = t+2^b,\ \ \ z_2 = d+2^b, d+2\times 2^b,d+3\times2^b,..., \right)\\\notag
=&\sum_{j=1}\mathbf{Pr}\left(z_1 = t+2^b,\ \ z_2 = d+j\times 2^b,..., \right)\\\notag
=&\sum_{j=1}r_2(r_1-s)\left[1-r_2\right]^{d+j\times 2^b-t-2^b-1}\left[1-(r_1+r_2-s)\right]^{t+2^b}\\\notag
=&r_2(r_1-s)\frac{\left[1-r_2\right]^{d-t-1}}{1-\left[1-r_2\right]^{2^b}}\left[1-(r_1+r_2-s)\right]^{t+2^b}
\end{align}

\begin{align}\notag
&\mathbf{Pr}\left(u_{1,b} = t, u_{2,b} = d, \  t<d, \ z_1<z_2\right)\\\notag
=&\sum_{i=0}r_2(r_1-s)\frac{\left[1-r_2\right]^{d-t-1}}{1-\left[1-r_2\right]^{2^b}}\left[1-(r_1+r_2-s)\right]^{t+i\times2^b}\\\notag
=&r_2(r_1-s)\frac{\left[1-r_2\right]^{d-t-1}}{1-\left[1-r_2\right]^{2^b}}
\frac{\left[1-(r_1+r_2-s)\right]^{t}}{1-\left[1-(r_1+r_2-s)\right]^{2^b}}
\end{align}}

Next, we study
\begin{align}\notag
&\mathbf{Pr}\left(u_{1,b} = t, u_{2,b} = d, \  t<d, \ z_1>z_2\right)
\end{align}

{\small\begin{align}\notag
&\mathbf{Pr}\left(z_1 = d,..., \ z_2 = t+2^b, t+2\times 2^b,t+3\times2^b,..., \ t<d \right)\\\notag
=&\sum_{j=1}r_2(r_1-s)\left[1-r_2\right]^{t+j\times2^b-d-1}\left[1-(r_1+r_2-s)\right]^{d}\\\notag
=&r_2(r_1-s)\frac{\left[1-r_2\right]^{t+2^b-d-1}}{1-\left[1-r_2\right]^{2^b}}\left[1-(r_1+r_2-s)\right]^{d}
\end{align}

\begin{align}\notag
&\mathbf{Pr}\left(z_1 = d+2^b,..., \ z_2 = t+2\times 2^b,t+3\times2^b,..., \ t<d \right)\\\notag
=&\sum_{j=2}r_2(r_1-s)\left[1-r_2\right]^{t+j\times2^b-d-2^b-1}\left[1-(r_1+r_2-s)\right]^{d+2^b}\\\notag
=&r_2(r_1-s)\frac{\left[1-r_2\right]^{t+2^b-d-1}}{1-\left[1-r_2\right]^{2^b}}\left[1-(r_1+r_2-s)\right]^{d+2^b}
\end{align}

\begin{align}\notag
&\mathbf{Pr}(z_1 = d,d+2^b,d+2\times2^b..., \\\notag
 &\hspace{0.3in} z_2 = t,t+2^b,t+2\times 2^b,..., \ t<d, \ z_1<z_2 )\\\notag
=&\sum_{i=0}r_2(r_1-s)\frac{\left[1-r_2\right]^{t+2^b-d-1}}{1-\left[1-r_2\right]^{2^b}}\left[1-(r_1+r_2-s)\right]^{d+i\times2^b}\\\notag
=&r_2(r_1-s)\frac{\left[1-r_2\right]^{t+2^b-d-1}}{1-\left[1-r_2\right]^{2^b}}\frac{\left[1-(r_1+r_2-s)\right]^{d}}{1-\left[1-(r_1+r_2-s)\right]^{2^b}}
\end{align}}

By symmetry
\begin{align}\notag
&\mathbf{Pr}(z_1 = t,t+2^b,t+2\times 2^b,...,\\\notag
 &\hspace{0.23in} z_2 = d,d+2^b,d+2\times2^b...,   \ t<d, \ z_1>z_2 )\\\notag
=&r_1(r_2-s)\frac{\left[1-r_1\right]^{t+2^b-d-1}}{1-\left[1-r_1\right]^{2^b}}\frac{\left[1-(r_1+r_2-s)\right]^{d}}{1-\left[1-(r_1+r_2-s)\right]^{2^b}}
\end{align}
Combining the results yields
{\small\begin{align}\notag
&\mathbf{Pr}\left(u_{1,b} = t, u_{2,b} = d, \  t<d\right)\\\notag
=&r_2(r_1-s)\frac{\left[1-r_2\right]^{d-t-1}}{1-\left[1-r_2\right]^{2^b}}
\frac{\left[1-(r_1+r_2-s)\right]^{t}}{1-\left[1-(r_1+r_2-s)\right]^{2^b}}\\\notag
+&r_1(r_2-s)\frac{\left[1-r_1\right]^{t+2^b-d-1}}{1-\left[1-r_1\right]^{2^b}}\frac{\left[1-(r_1+r_2-s)\right]^{d}}{1-\left[1-(r_1+r_2-s)\right]^{2^b}}
\end{align}

\begin{align}\notag
&\mathbf{Pr}\left(u_{1,b} = t, u_{2,b} = d, \  t>d\right)\\\notag
=&\mathbf{Pr}\left(u_{2,b} = d, u_{1,b} = t, \  d<t\right)\\\notag
=&r_1(r_2-s)\frac{\left[1-r_1\right]^{t-d-1}}{1-\left[1-r_1\right]^{2^b}}
\frac{\left[1-(r_1+r_2-s)\right]^{d}}{1-\left[1-(r_1+r_2-s)\right]^{2^b}}\\\notag
+&r_2(r_1-s)\frac{\left[1-r_2\right]^{d+2^b-t-1}}{1-\left[1-r_2\right]^{2^b}}
\frac{\left[1-(r_1+r_2-s)\right]^{t}}{1-\left[1-(r_1+r_2-s)\right]^{2^b}}
\end{align}}

Now, we need to compute $\mathbf{Pr}\left(u_{1,b} = t, u_{2,b} = t\right)$:
{\small\begin{align}\notag
&\mathbf{Pr}\left(u_{1,b} = t, u_{2,b} = t\right) \\\notag
=&\sum_{i=0}\sum_{j=0} \mathbf{Pr}\left(z_1 = t+i\times 2^b, z_2 = t + j\times 2^b\right)
\end{align}
\begin{align}\notag
&\mathbf{Pr}\left(u_{1,b} = t, u_{2,b} = t\right) \\\notag
=&\mathbf{Pr}\left(u_{1,b} = t, u_{2,b} = t, z_1 = z_2\right)\\\notag
 +&\mathbf{Pr}\left(u_{1,b} = t, u_{2,b} = t, z_1 \neq z_2\right)
\end{align}

\begin{align}\notag
&\mathbf{Pr}\left(u_{1,b} = t, u_{2,b} = t, z_1 = z_2\right)\\\notag
=&\sum_{i=0}\mathbf{Pr}\left(z_1=z_2=t+i\times 2^b, ...\right)\\\notag
=&\sum_{i=0} s\left[1-(r_1+r_2-s)\right]^{t+i\times2^b}\\\notag
=&\frac{s\left[1-(r_1+r_2-s)\right]^{t}}{1-\left[1-(r_1+r_2-s)\right]^{2^b}}
\end{align}

\begin{align}\notag
&\mathbf{Pr}\left(u_{1,b} = t, u_{2,b} = t, z_1 < z_2\right)\\\notag
=&\sum_{i=0}\sum_{j=i+1} \mathbf{Pr}\left(z_1 = t+i\times 2^b, z_2 = t+j\times 2^b\right)\\\notag
=&\sum_{i=0}\sum_{j=i+1} r_2(r_1-s)\left[1-r_2\right]^{(j-i)\times 2^b-1}\left[1-(r_1+r_2-s)\right]^{t+i\times 2^b}\\\notag
=&\sum_{i=0} r_2(r_1-s)\frac{\left[1-r_2\right]^{2^b-1}}{1-\left[1-r_2\right]^{2^b}}\left[1-(r_1+r_2-s)\right]^{t+i\times 2^b}\\\notag
=& r_2(r_1-s)\frac{\left[1-r_2\right]^{2^b-1}}{1-\left[1-r_2\right]^{2^b}}\frac{\left[1-(r_1+r_2-s)\right]^{t}}{1- \left[1-(r_1+r_2-s)\right]^{2^b}}
\end{align}

\begin{align}\notag
&\mathbf{Pr}\left(u_{1,b} = t, u_{2,b} = t, z_1 >z_2\right)\\\notag
=& r_1(r_2-s)\frac{\left[1-r_1\right]^{2^b-1}}{1-\left[1-r_1\right]^{2^b}}\frac{\left[1-(r_1+r_2-s)\right]^{t}}{1- \left[1-(r_1+r_2-s)\right]^{2^b}}
\end{align}}

Combining the results yields
\begin{align}\notag
&\mathbf{Pr}\left(u_{1,b} = t, u_{2,b} = t\right) \\\notag
=&\frac{s\left[1-(r_1+r_2-s)\right]^{t}}{1-\left[1-(r_1+r_2-s)\right]^{2^b}}\\\notag
+&r_2(r_1-s)\frac{\left[1-r_2\right]^{2^b-1}}{1-\left[1-r_2\right]^{2^b}}\frac{\left[1-(r_1+r_2-s)\right]^{t}}{1- \left[1-(r_1+r_2-s)\right]^{2^b}}\\\notag
+&r_1(r_2-s)\frac{\left[1-r_1\right]^{2^b-1}}{1-\left[1-r_1\right]^{2^b}}\frac{\left[1-(r_1+r_2-s)\right]^{t}}{1- \left[1-(r_1+r_2-s)\right]^{2^b}}
\end{align}

Finally, we re-write the probabilities in terms of $R=P_=$, $P_<$, and $P_>$ whenever possible:
{\small\begin{align}\notag
&\mathbf{Pr}\left(u_{1,b} = t, u_{2,b} = d, \  t<d\right)\\\notag
=&P_< \frac{r_2\left[1-r_2\right]^{d-t-1}}{1-\left[1-r_2\right]^{2^b}}
\frac{\left[1-(r_1+r_2-s)\right]^{t}(r_1+r_2-s)}{1-\left[1-(r_1+r_2-s)\right]^{2^b}}\\\notag
+&P_> \frac{r_1\left[1-r_1\right]^{t+2^b-d-1}}{1-\left[1-r_1\right]^{2^b}}
\frac{\left[1-(r_1+r_2-s)\right]^{d}(r_1+r_2-s)}{1-\left[1-(r_1+r_2-s)\right]^{2^b}}\\\notag
\end{align}
\begin{align}\notag
&\mathbf{Pr}\left(u_{1,b} = t, u_{2,b} = d, \  t>d\right)\\\notag
=&P_> \frac{r_1\left[1-r_1\right]^{t-d-1}}{1-\left[1-r_1\right]^{2^b}}
\frac{\left[1-(r_1+r_2-s)\right]^{d}(r_1+r_2-s)}{1-\left[1-(r_1+r_2-s)\right]^{2^b}}\\\notag
+&P_< \frac{r_2\left[1-r_2\right]^{d+2^b-t-1}}{1-\left[1-r_2\right]^{2^b}}
\frac{\left[1-(r_1+r_2-s)\right]^{t}(r_1+r_2-s)}{1-\left[1-(r_1+r_2-s)\right]^{2^b}}\\\notag
\end{align}
\begin{align}\notag
&\mathbf{Pr}\left(u_{1,b} = t, u_{2,b} = t\right) \\\notag
=&\left(R+P_<\frac{r_2\left[1-r_2\right]^{2^b-1}}{1-\left[1-r_2\right]^{2^b}}+P_>\frac{r_1\left[1-r_1\right]^{2^b-1}}{1-\left[1-r_1\right]^{2^b}}\right)\times\\\notag
+&\frac{\left[1-(r_1+r_2-s)\right]^{t}(r_1+r_2-s)}{1- \left[1-(r_1+r_2-s)\right]^{2^b}}
\end{align}}

\vspace{-0.2in}
{\small

\begin{thebibliography}{10}

\bibitem{Proc:Bendersky_WSDM09}
Michael Bendersky and W.~Bruce Croft.
\newblock Finding text reuse on the web.
\newblock In {\em WSDM}, pages 262--271, Barcelona, Spain, 2009.

\bibitem{Proc:Broder}
Andrei~Z. Broder.
\newblock On the resemblance and containment of documents.
\newblock In {\em the Compression and Complexity of Sequences}, pages 21--29,
  Positano, Italy, 1997.

\bibitem{Proc:Broder_Stoc98}
Andrei~Z. Broder, Moses Charikar, Alan~M. Frieze, and Michael Mitzenmacher.
\newblock Min-wise independent permutations (extended abstract).
\newblock In {\em STOC}, pages 327--336, Dallas, TX, 1998.

\bibitem{Proc:Broder_WWW97}
Andrei~Z. Broder, Steven~C. Glassman, Mark~S. Manasse, and Geoffrey Zweig.
\newblock Syntactic clustering of the web.
\newblock In {\em WWW}, pages 1157 -- 1166, Santa Clara, CA, 1997.

\bibitem{Proc:Buehrer_WSDM08}
Gregory Buehrer and Kumar Chellapilla.
\newblock A scalable pattern mining approach to web graph compression with
  communities.
\newblock In {\em WSDM}, pages 95--106, Stanford, CA, 2008.

\bibitem{Byers:2002:ICD:633025.633031}
John Byers, Jeffrey Considine, Michael Mitzenmacher, and Stanislav Rost.
\newblock Informed {C}ontent {D}elivery across {A}daptive {O}verlay {N}etworks.
\newblock In {\em {S}{I}{G}{C}{O}{M}{M}}, pages 47--60, 2002.

\bibitem{Proc:Cherkasova_KDD09}
Ludmila Cherkasova, Kave Eshghi, Charles B.~Morrey III, Joseph Tucek, and
  Alistair~C. Veitch.
\newblock Applying syntactic similarity algorithms for enterprise information
  management.
\newblock In {\em KDD}, pages 1087--1096, Paris, France, 2009.

\bibitem{Proc:Chierichetti_KDD09}
Flavio Chierichetti, Ravi Kumar, Silvio Lattanzi, Michael Mitzenmacher,
  Alessandro Panconesi, and Prabhakar Raghavan.
\newblock On compressing social networks.
\newblock In {\em KDD}, pages 219--228, Paris, France, 2009.

\bibitem{Proc:Dasu_SIGMOD02}
T.~Dasu, T.~Johnson, S.~Muthukrishnan, and V.~Shkapenyuk.
\newblock Mining database structure; or, how to build a data quality browser.
\newblock In {\em SIGMOD}, pages 240--251, Madison, WI, 2002.

\bibitem{Article:Dourisboure09}
Yon Dourisboure, Filippo Geraci, and Marco Pellegrini.
\newblock Extraction and classification of dense implicit communities in the
  web graph.
\newblock {\em ACM Trans. Web}, 3(2):1--36, 2009.

\bibitem{Article:Fan_JMLR08}
Rong-En Fan, Kai-Wei Chang, Cho-Jui Hsieh, Xiang-Rui Wang, and Chih-Jen Lin.
\newblock Liblinear: A library for large linear classification.
\newblock {\em Journal of Machine Learning Research}, 9:1871--1874, 2008.

\bibitem{Proc:Fetterly_WWW03}
Dennis Fetterly, Mark Manasse, Marc Najork, and Janet~L. Wiener.
\newblock A large-scale study of the evolution of web pages.
\newblock In {\em WWW}, pages 669--678, Budapest, Hungary, 2003.

\bibitem{Article:Forman09}
George Forman, Kave Eshghi, and Jaap Suermondt.
\newblock Efficient detection of large-scale redundancy in enterprise file
  systems.
\newblock {\em SIGOPS Oper. Syst. Rev.}, 43(1):84--91, 2009.

\bibitem{Proc:Gollapudi_WWW09}
Sreenivas Gollapudi and Aneesh Sharma.
\newblock An axiomatic approach for result diversification.
\newblock In {\em WWW}, pages 381--390, Madrid, Spain, 2009.

\bibitem{Gupta02approximaterange}
Systems~Abhishek Gupta, Abhishek Gupta, Divyakant Agrawal, and Amr~El Abbadi.
\newblock Approximate {R}ange {S}election {Q}ueries in {P}eer-to-{P}eer.
\newblock In {\em In {C}{I}{D}{R}}, 2002.

\bibitem{Proc:Hsieh_ICML08}
Cho-Jui Hsieh, Kai-Wei Chang, Chih-Jen Lin, S.~Sathiya Keerthi, and
  S.~Sundararajan.
\newblock A dual coordinate descent method for large-scale linear svm.
\newblock In {\em Proceedings of the 25th international conference on Machine
  learning}, ICML, pages 408--415, 2008.

\bibitem{Proc:Nitin_WSDM08}
Nitin Jindal and Bing Liu.
\newblock Opinion spam and analysis.
\newblock In {\em WSDM}, pages 219--230, Palo Alto, California, USA, 2008.

\bibitem{Proc:Joachims_KDD06}
Thorsten Joachims.
\newblock Training linear svms in linear time.
\newblock In {\em KDD}, pages 217--226, Pittsburgh, PA, 2006.

\bibitem{Article:Kalpakis08}
Konstantinos Kalpakis and Shilang Tang.
\newblock Collaborative data gathering in wireless sensor networks using
  measurement co-occurrence.
\newblock {\em Computer Communications}, 31(10):1979--1992, 2008.

\bibitem{Article:Li_Konig_CACM11}
Ping Li and Arnd~Christian K\"onig.
\newblock Theory and applications b-bit minwise hashing.
\newblock In {\em Commun. ACM}, 2011.

\bibitem{Report:HashLearning11}
Ping Li, Anshumali Shrivastava, Joshua Moore, and Arnd~Christian K\"onig.
\newblock Hashing algorithms for large-scale learning.
\newblock Technical report.

\bibitem{Proc:Li_Konig_WWW10}
Ping Li and Arnd~Christian \text{K\"{o}nig}.
\newblock b-bit minwise hashing.
\newblock In {\em WWW}, pages 671--680, Raleigh, NC, 2010.

\bibitem{Proc:Li_Konig_NIPS10}
Ping Li, Arnd~Christian \text{K\"{o}nig}, and Wenhao Gui.
\newblock b-bit minwise hashing for estimating three-way similarities.
\newblock In {\em NIPS}, Vancouver, BC, 2010.

\bibitem{Proc:Manku_WWW07}
Gurmeet~Singh Manku, Arvind Jain, and Anish~Das Sarma.
\newblock {D}etecting {N}ear-{D}uplicates for {W}eb-{C}rawling.
\newblock In {\em WWW}, Banff, Alberta, Canada, 2007.

\bibitem{Proc:Najork_WSDM09}
Marc Najork, Sreenivas Gollapudi, and Rina Panigrahy.
\newblock Less is more: sampling the neighborhood graph makes salsa better and
  faster.
\newblock In {\em WSDM}, pages 242--251, Barcelona, Spain, 2009.

\bibitem{Ouyang:2002:CDC:645962.674230}
Zan Ouyang, Nasir~D. Memon, Torsten Suel, and Dimitre Trendafilov.
\newblock Cluster-{B}ased {D}elta {C}ompression of a {C}ollection of {F}iles.
\newblock In {\em {W}{I}{S}{E}}, pages 257--268, 2002.

\bibitem{Proc:Pandey_WWW09}
Sandeep Pandey, Andrei Broder, Flavio Chierichetti, Vanja Josifovski, Ravi
  Kumar, and Sergei Vassilvitskii.
\newblock Nearest-neighbor caching for content-match applications.
\newblock In {\em WWW}, pages 441--450, Madrid, Spain, 2009.

\bibitem{Proc:Shalev-Shwartz_ICML07}
Shai Shalev-Shwartz, Yoram Singer, and Nathan Srebro.
\newblock Pegasos: Primal estimated sub-gradient solver for svm.
\newblock In {\em ICML}, pages 807--814, Corvalis, Oregon, 2007.

\bibitem{Article:Urvoy08}
Tanguy Urvoy, Emmanuel Chauveau, Pascal Filoche, and Thomas Lavergne.
\newblock Tracking web spam with html style similarities.
\newblock {\em ACM Trans. Web}, 2(1):1--28, 2008.

\bibitem{Proc:Yu_KDD10}
Hsiang-Fu Yu, Cho-Jui Hsieh, Kai-Wei Chang, and Chih-Jen Lin.
\newblock Large linear classification when data cannot fit in memory.
\newblock In {\em KDD}, pages 833--842, 2010.

\bibitem{Zhang:2010:MFK:1920841.1920944}
Meihui Zhang, Marios Hadjieleftheriou, Beng~Chin Ooi, Cecilia~M. Procopiuc, and
  Divesh Srivastava.
\newblock On {M}ulti-column {F}oreign {K}ey {D}iscovery.
\newblock {\em Proc. VLDB Endow.}, 3:805--814, September 2010.

\end{thebibliography}

}

\end{document}